\documentclass{article}

 \usepackage[main, final]{neurips_2026}
\usepackage[utf8]{inputenc} % allow utf-8 input
\usepackage[T1]{fontenc}    % use 8-bit T1 fonts
\usepackage{hyperref}       % hyperlinks
\usepackage{url}            % simple URL typesetting
\usepackage{booktabs}       % professional-quality tables
\usepackage{nicefrac}       % compact symbols for 1/2, etc.
\usepackage{microtype}      % microtypography

\usepackage{graphicx}
\usepackage{subcaption}
\usepackage{amsmath,amssymb,amsfonts,amsthm,bbm,mathrsfs}
\usepackage{multirow}
\usepackage{soul}

\usepackage{array}

\usepackage[scr=stixtwofancy,cal=cm]{mathalfa} 
\DeclareMathAlphabet{\mathdutchcal}{U}{dutchcal}{m}{n}
\SetMathAlphabet{\mathdutchcal}{bold}{U}{dutchcal}{b}{n}
\DeclareMathAlphabet{\mathdutchbcal}{U}{dutchcal}{b}{n}
\usepackage{bm}
\usepackage[dvipsnames]{xcolor}

\usepackage{enumitem}

\usepackage{algorithm}
\usepackage{algorithmic} 

\allowdisplaybreaks[4]

\theoremstyle{plain}
\newtheorem{theorem}{Theorem}[section]
\newtheorem{proposition}[theorem]{Proposition}
\newtheorem{lemma}[theorem]{Lemma}

\theoremstyle{definition}
\newtheorem{definition}[theorem]{Definition}
\newtheorem{assumption}[theorem]{Assumption}
\theoremstyle{remark}
\newtheorem{remark}[theorem]{Remark}

\title{Dual-Granularity Learning for Regression with Continuous Noisy Labels}

\author{%
  Hui Guo$^{1}$ \qquad
  Boyu Wang$^{1,3}$ \qquad
  Grace Y. Yi$^{1,2,3,*}$ \\[6pt]
  $^{1}$Department of Computer Science,
  University of Western Ontario \\
  $^{2}$Department of Statistical and Actuarial Sciences,
  University of Western Ontario \\
  $^{3}$Vector Institute \\[4pt]
  \texttt{hguo288@uwo.ca} \quad
  \texttt{bwang@csd.uwo.ca} \quad
  \texttt{gyi5@uwo.ca} \\[2pt]
  {\small $^{*}$Corresponding author}
}

\begin{document}

\maketitle

\begin{abstract}
  Learning with noisy labels has been extensively studied for classification, yet its regression counterpart remains relatively underexplored. Unlike classification, regression involves continuous-valued targets where label noise exhibits heterogeneous magnitudes across instances, {which poses conflicting requirements on the model: correcting small-to-moderate deviations requires sensitivity to label values for instance-level refinement, whereas large deviations caused by corrupted or outlying annotations render exact label values unreliable, necessitating reduced reliance on exact labels and instead leveraging ordinal structure among samples.} These requirements call for methods that capture both global structure and sample-specific deviations in a unified manner.
  To this end, we propose GRAIN, a dual-granularity framework for robust noisy regression. At the fine-grained level, we derive an explicit residual learning mechanism from a robust loss formulation, introducing instance-specific auxiliary variables to decouple label noise from predictive signals. At the coarse-grained level, we introduce a binning-based contrastive regularization that enforces an order-preserving feature structure aligned with label proximity, addressing noise beyond the capacity of local residual refinement. The two levels are coupled via a histogram-based bridging mechanism that enables bidirectional information flow: coarse-grained structural guidance establishes a well-ordered feature space that anchors residual learning, while fine-grained residual refinement corrects instance-level deviations that structural regularization alone cannot resolve. Experimental results show that GRAIN consistently improves robustness and generalization under diverse noise conditions. 
  Code is available at \href{https://github.com/hguo1728/GRAIN}{https://github.com/hguo1728/GRAIN}.
\end{abstract}

\section{Introduction}\label{sec.intro}

Regression is a fundamental learning paradigm with broad applications in multiple fields, including finance, healthcare, and economics. In practice, however, regression models are often trained with imperfect supervision, where noisy labels arise inevitably due to measurement errors, annotation uncertainty, or system bias \cite{yi2017statistical}. Such label noise can destabilize training, degrade predictive reliability, and substantially limit the deployment of regression systems in high-stakes applications.

Despite its practical importance, learning with continuous noisy labels remains relatively underexplored compared to classification~\cite{han2018co,xia2020robust,song2019selfie}. 
This gap stems from a fundamental difference in label structure: regression labels are continuous-valued and lack explicit decision boundaries, {leading to  label noise with heterogeneous magnitude across instances. This induces conflicting requirements on how a model should respond to corrupted supervision. 
For small-to-moderate deviations, label values are imperfect but retain sufficient information for local correction, requiring \emph{sensitivity to label values} for precise instance-level refinement. 
For large deviations, which often arise from outlying or heavily corrupted annotations, exact label values become unreliable and may mislead learning, necessitating \emph{reduced reliance on label values} and instead modeling and enforcing the underlying ordered feature-label structure.
These conflicting requirements cannot be effectively resolved by a single mechanism. Existing approaches largely focus on improving overall robustness~\cite{husimple, yao2022c, castells2020superloss} without explicitly addressing these coupled challenges.}

% leading to two coupled challenges under noisy supervision.
% First, label corruption varies substantially in magnitude across instances. This heterogeneous-magnitude noise requires sample-specific correction, making uniform treatment ineffective and motivating \textit{fine-grained} residual refinement.
% Second, noisy labels distort the relational structure among samples. In deep regression, feature representations are expected to reflect label proximity, where closer labels are mapped to more similar representations while distant ones are separated~\cite{boudiaf2020unifying, zhangimproving}. Corrupted supervision disrupts this geometry, leading to disordered feature structure and degraded generalization, which calls for a \textit{coarse-grained} mechanism to restore smooth, order-preserving global feature structure.
% Existing approaches largely focus on improving overall robustness~\cite{husimple, yao2022c, castells2020superloss} without explicitly addressing these coupled challenges. 

\begin{figure}
    \centering
    \includegraphics[width=\linewidth]{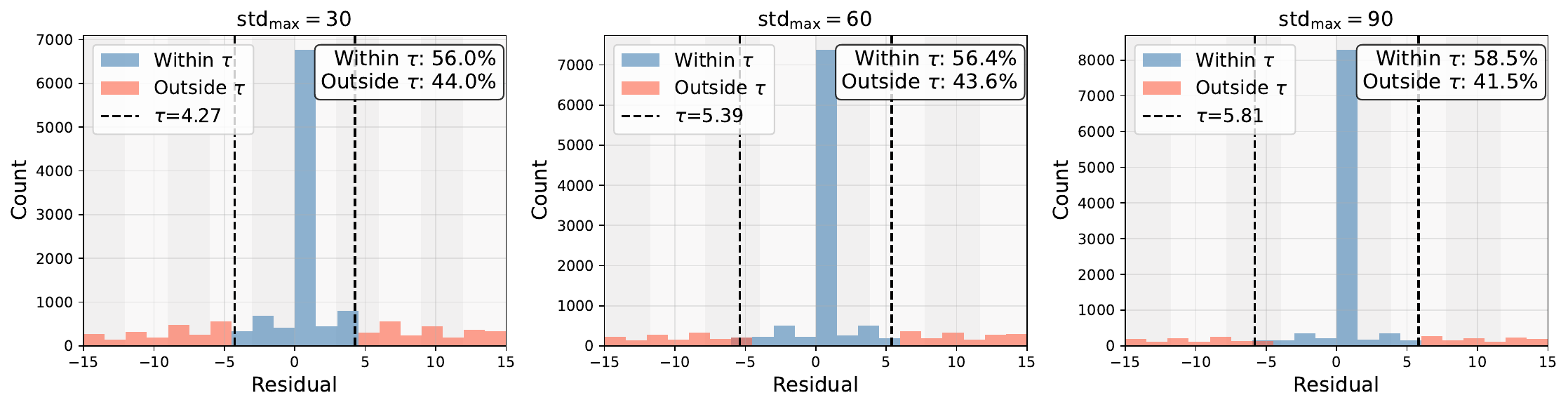}
    \caption{Distribution of true label noise $\{\widetilde{Y}_i - Y_i : i = 1, \ldots, n\}$ at the final training epoch on UTKFace under three noise levels ($\text{std}_{\max}$ = 30, 60, 90). The \textbf{dashed lines} (\textbf{-\,-\,-}) indicate $\pm\tau$.  \textbf{Blue bars} (\textcolor[RGB]{144, 179, 210}{\rule{6pt}{6pt}}) show noise within $(-\tau, \tau)$ (within the residual head's capacity), and \textbf{red bars} (\textcolor[RGB]{255, 161, 144}{\rule{6pt}{6pt}}) show noise outside $(-\tau, \tau)$ (beyond the residual head's capacity). \textbf{The proportion of label noise} within and outside $(-\tau, \tau)$ is annotated in each panel. \textbf{Alternating shaded regions} represent consecutive label bins. The label noise range is clipped to $[-15, 15]$ for visualization purposes.}
    \label{fig:tau_distribution}
\end{figure}

To address these challenges, we propose \textbf{GRAIN}, a dual-granularity framework that explicitly models instance-level residual noise while enforcing a globally ordered feature-label structure within a unified learning process.
At the \textit{fine-grained level}, we derive a piecewise robust loss family for noisy regression and identify a formulation admitting an infimal convolution representation, which introduces auxiliary variables that explicitly capture sample-wise residuals with heterogeneous magnitudes. Based on this formulation, we design a lightweight residual head to learn these variables during training, enabling label noise to be explicitly separated from predictive signals. To prevent the residual head from absorbing predictive signals, we impose an adaptive capacity constraint controlled by a threshold $\tau$, such that it is restricted to modeling only bounded residual components.
As illustrated in Figure~\ref{fig:tau_distribution}, this constraint naturally induces a separation of correction regimes: deviations within $(-\tau,\tau)$ are handled through fine-grained residual refinement, while larger deviations{, where label values become unreliable and induce structural distortion in feature space}, are instead addressed by coarse-grained structural regularization.
Accordingly, at the \textit{coarse-grained level}, we address the remaining structural corruption by introducing a binning-based contrastive objective. Specifically, we perform adaptive binning based on the density of noisy labels and encourage representations of adjacent bins to align while pushing apart distant bins, thereby preserving the ordinal structure of the label space. This design enforces global order in the feature space while remaining robust to large-magnitude noise that cannot be captured by local residual refinement.

\begin{figure}
    \centering
    \includegraphics[width=0.92\linewidth]{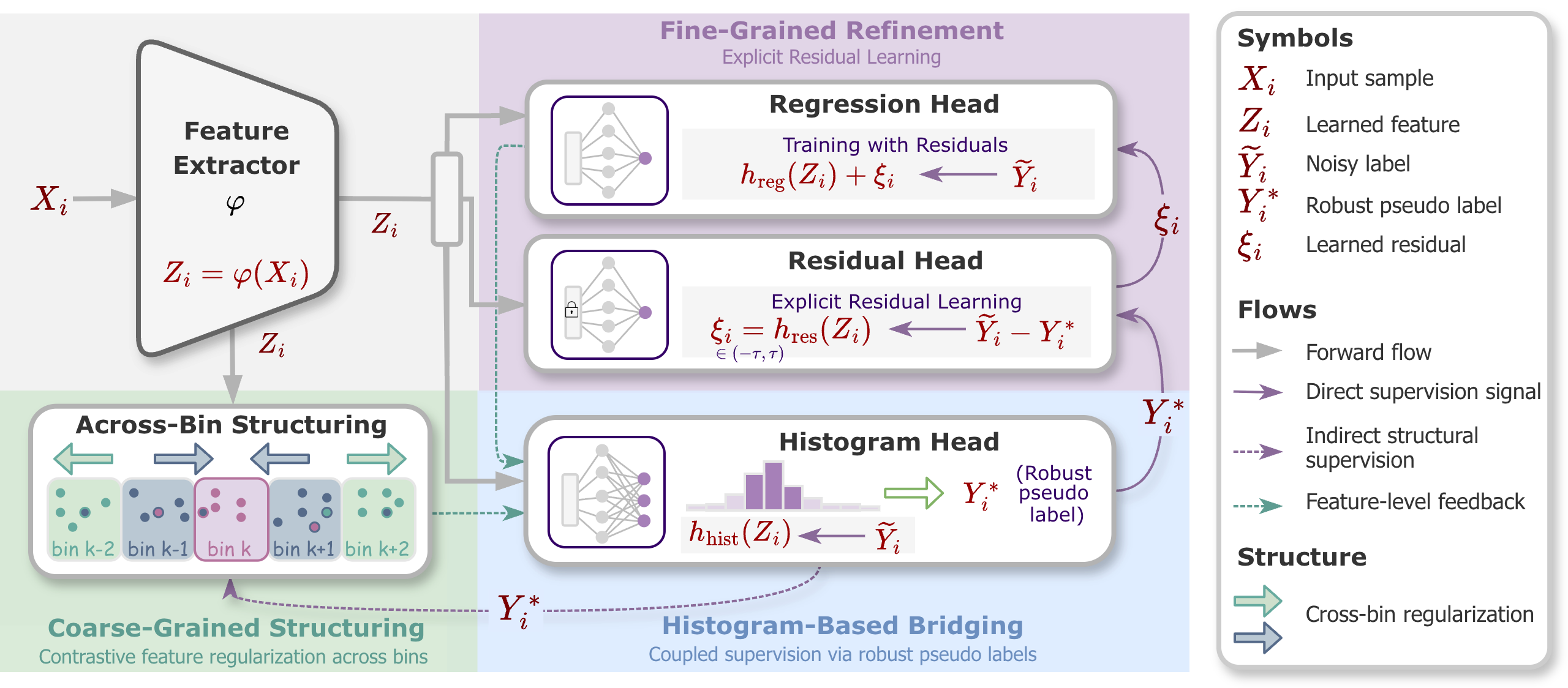}
    \caption{
    Overview of \textbf{GRAIN}. \textbf{Fine-grained refinement} (top right): a regression head $h_\text{reg}$ produces predictions corrected by instance-specific residuals $\xi_i$ learned by a capacity-constrained residual head $h_\text{res} \in (-\tau, \tau)$, decoupling sample-wise noise from predictive signals. \textbf{Coarse-grained structuring} (bottom left): a binning-based contrastive objective enforces an order-preserving feature structure aligned with label proximity. \textbf{Histogram-based bridging} (bottom right): a histogram head $h_\text{hist}$ generates robust pseudo labels $Y_i^*$, guiding bin assignment and providing residual supervision targets, while benefiting in turn from the representations refined by both modules.
    }
    \label{fig:GRAIN_overview}
\end{figure}

As illustrated in Figure \ref{fig:GRAIN_overview}, these two mechanisms operate at different granularities and are inherently complementary. Fine-grained refinement provides precise instance-level correction by capturing localized residual variations, while coarse-grained structuring enforces an order-preserving feature geometry aligned with label proximity.
A histogram head operating over the binned label space serves as a mediator, extracting pseudo labels from the shared feature representations to provide coupled supervision across the two levels: the pseudo labels guide coarse-grained bin assignment and provide supervision for fine-grained residual learning. In turn, the refined and structured feature representations further improve pseudo-label quality, forming a closed-loop between feature learning and pseudo-label estimation. 
Together, this unified design provides a coherent solution to noisy regression by simultaneously addressing local corruption and global structural distortion.

% As shown in Figure \ref{fig:GRAIN_overview}, these two mechanisms act at different granularities and are mutually reinforcing: coarse-grained structural guidance establishes a globally ordered representation space that constrains and stabilizes learning, while fine-grained residual refinement corrects instance-level deviations within this structured space. Together, they provide a coherent solution to noisy regression by simultaneously addressing local corruption and global structural distortion.

%%%%%%%%%%%%%%%%%%%%%%%%%%%%%%%%%%%%%%%%%%%%%%%%%%%%%%%%%%%%%%%%%%%%%%%%%%%%%%%%

\section{Related Work}\label{sec.related_work}

\textbf{Learning with Noisy Labels.}
Learning with noisy labels has been extensively studied in classification, where labels are discrete and noise is often addressed through sample selection \cite{han2018co, xiasample}, self-training \cite{song2019selfie}, {as well as annotation modeling and robust pseudo-labeling \cite{guo2023label,guo2024learning, guo2026addressing}.}
In contrast, learning with continuous noisy labels for regression has received considerably less attention. Existing approaches primarily focus on improving robustness at the loss or sample level, for example via robust loss functions \cite{huber2011robust}, data augmentation \cite{yao2022c}, adaptive reweighting \cite{castells2020superloss}, or sample selection \cite{kim2024sample}.
While effective in reducing sensitivity to noisy labels, these methods treat residuals implicitly as errors rather than deriving explicit residual variables from the loss formulation itself, and typically handle label corruption in a uniform manner, without accounting for its heterogeneous magnitude across instances or its effect on the global structure of the feature space.
Unlike these approaches, our method jointly addresses instance-level residual noise and global structural distortion within a unified dual-granularity framework.

\textbf{Discretization / Binning-based Methods.}
Discretization-based auxiliary classification has been explored in regression to stabilize training and improve representation learning \cite{zhangimproving, stewart2023regression}.
These methods map continuous targets into discrete bins and apply classification-style supervision, which can enhance feature separability but often ignores the intrinsic ordering and continuity of regression labels.
Instead, our approach leverages adaptive binning as a mechanism for structural regularization, enforcing order-preserving relationships across bins to establish global feature structure under noisy supervision rather than redefining the task as classification. Furthermore, a histogram head operating over the same bin structure generates robust pseudo labels that bridge coarse-grained structural guidance and fine-grained residual refinement. {Discretized density guidance has also been studied for source-free regression adaptation \cite{xu2026discretized}; GRAIN instead targets continuous label corruption within a single training distribution.
}

% Instead, our approach leverages adaptive binning as a mechanism for structural regularization rather than redefining the task as classification, enforcing order-preserving relationships across bins to restore global feature structure under noisy supervision. Furthermore, a histogram head operating over the same bin structure generates robust pseudo labels that couple supervision across both levels, jointly integrating coarse-grained structural guidance with fine-grained residual refinement.

%%%%%%%%%%%%%%%%%%%%%%%%%%%%%%%%%%%%%%%%%%%%%%%%%%%%%%%%%%%%%%%%%%%%%%%%%%%%%%%%

\section{Preliminaries}\label{sec.preliminary}

We consider a regression setting with a $p$-dimensional input random variable $\mathrm{X}$ and an associated label $\mathrm{Y}$, where $\mathrm{X}\in\mathcal{X}$ and $\mathrm{Y}\in\mathcal{Y}$, with $\mathcal{X}\subset\mathbb{R}^p$ denoting the input space and $\mathcal{Y}\subset\mathbb{R}$ the label space. In many applications, the true label $\mathrm{Y}$ is not directly observable. Instead, one has access to a noisy label $\widetilde{\mathrm{Y}}\in\mathcal{Y}$. The data are assumed to follow the generative mechanism: 
\begin{align}
    &\mathrm{Y}=f(\mathrm{X};\theta^*)+\varepsilon',\label{eq.true_label}\\
    &\widetilde{\mathrm{Y}}=\mathrm{Y}+\varsigma(\mathrm{X},\mathrm{Y})\varepsilon'',\label{eq.noisy_label}
\end{align}
where {$f(\cdot;\theta^*)$ denotes the unknown ground-truth regression function, indexed by parameter $\theta^*$}; $\varepsilon'$ and $\varepsilon''$ are independent zero-mean noise variables {with finite variance,} independent of $\mathrm{X}$; and  $\varsigma:\mathcal{X}\times\mathcal{Y}\rightarrow\mathbb{R}$ {represents an unknown noise scale function, potentially instance and label-dependent, allowing for heteroskedastic and structured label corruption.} % {\color{red}(add: remark)}

Let $\mathcal{D}=\{\mathrm{X}_i,\widetilde{\mathrm{Y}}_i\}_{i=1}^{N}$ denote a training sample of size $N$, where $\widetilde{\mathrm{Y}}_i$ is the noisy label {associated with} $\mathrm{X}_i$, which may differ from the unobserved true label $\mathrm{Y}_i$ for $i\in[N]$. We assume that $(\mathrm{X}_i,\mathrm{Y}_i,\widetilde{\mathrm{Y}}_i)$ are independent and identically distributed (i.i.d.) according to the {joint} distribution of $(\mathrm{X},\mathrm{Y},\widetilde{\mathrm{Y}})$ characterized by (\ref{eq.true_label})-(\ref{eq.noisy_label}). Our goal is to learn a predictor that approximates the underlying regression function $f(\cdot;\theta^*)$ using only noisy data $\mathcal{D}$, which achieves reliable generalization on unseen inputs.

%%%%%%%%%%%%%%%%%%%%%%%%%%%%%%%%%%%%%%%%%%%%%%%%%%%%%%%%%%%%%%%%%%%%%%%%%%%%%%%%

\section{GRAIN: A Dual-Granularity Framework for Noisy Regression}\label{sec.method}

We present GRAIN, a dual-granularity framework for deep regression under 
noisy labels (Figure~\ref{fig:GRAIN_overview}), designed to effectively address the conflicting requirements induced by heterogeneous label noise.
An adaptive capacity threshold $\tau$ demarcates two regimes: deviations within $(-\tau,\tau)$ are addressed by fine-grained residual refinement, while larger deviations are delegated to coarse-grained structural regularization.
Section~\ref{sec.residual_head} develops a piecewise robust loss family 
with generalization guarantees, and derives an infimal convolution 
reformulation that introduces instance-specific auxiliary variables for 
fine-grained residual modeling within the capacity bound $(-\tau,\tau)$.
Section~\ref{sec.feature_structuring} introduces a binning-based contrastive 
regularization that operates at a coarser granularity, enforcing an 
order-preserving feature geometry across bins of the label space and 
addressing structural corruption beyond the reach of local refinement.
Section~\ref{sec.bridge} describes how a histogram-based bridging mechanism 
mediates supervision across both levels, coupling residual targets and bin 
assignments through shared pseudo labels to form a closed-loop between 
feature learning and noise correction.

\subsection{Fine-Grained Refinement: Loss-Induced Residual Learning}
\label{sec.residual_head}

We begin by focusing on robust prediction given a feature representation, 
with robust feature learning under noisy supervision deferred to 
Section~\ref{sec.feature_structuring}. Let $\mathrm{Z}=\varphi(\mathrm{X})$ 
denote the feature representation of input $\mathrm{X}$, where 
$\varphi:\mathcal{X}\rightarrow\mathcal{Z}\subset\mathbb{R}^d$ embeds the 
input into a $d$-dimensional feature space $\mathcal{Z}$.
We assume that $\varphi$ is sufficiently expressive such that the regression structure in (\ref{eq.true_label}) admits a linear parameterization over $\mathcal{Z}$: $\mathrm{Y}=\langle\mathrm{Z},\theta_{\text{reg}}^*\rangle+\varepsilon'$, where $\theta_{\text{reg}}^*$ represents the resulting regression parameter, and $\langle a,b\rangle$ denotes the dot product of vectors $a$ and $b$.
The training dataset is then $\mathcal{D}^\dagger=\{\mathrm{Z}_i,\widetilde{\mathrm{Y}}_i\}_{i=1}^N$, {with} $\mathrm{Z}_i=\varphi(\mathrm{X}_i)$ for all $i\in[N]$, and let $\widehat{P}^\dagger=\frac{1}{N}\sum_{i=1}^{N}\delta_{\mathrm{Z}_i,\widetilde{\mathrm{Y}}_i}$ denote the  empirical distribution of $(\mathrm{Z},\widetilde{\mathrm{Y}})$ induced from dataset $\mathcal{D}^\dagger$, where $\delta_{\mathrm{Z}_i,\widetilde{\mathrm{Y}}_i}$ represents the Dirac measure at $(\mathrm{Z}_i,\widetilde{\mathrm{Y}}_i)$.
With this setup, we proceed to design a robust loss function over $\widehat{P}^\dagger$ that is resilient to the heterogeneous label noise in $\widetilde{\mathrm{Y}}$, while admitting a residual-based reformulation for explicit noise modeling.

\subsubsection{Robust Loss Family}\label{sec.robust_loss_family}

Label noise in regression is heterogeneous in magnitude: small-to-moderate deviations are addressable at the instance level, while large deviations, which often arise from outlying or heavily corrupted annotations, render exact label values unreliable, such that their influence on training should be limited rather than fully absorbed by the model.
To this end, we introduce a piecewise loss family that applies quadratic sensitivity within a bounded range while controlling the growth of penalties for large deviations:
\begin{equation}\label{eq.piecewise_loss}
    \ell_\tau(\nu)=\left\{
    \begin{matrix}
    \frac{1}{2}\nu^2,\hspace{3.1cm}& \text{if}\ |\nu|\le\tau,\\
    \frac{\tau}{\ell'(\tau)}\left\{\ell(|\nu|)-\ell(\tau)\right\}+\frac{\tau^2}{2},& \text{if}\ |\nu|>\tau,
    \end{matrix}
    \right.
\end{equation}
where $\tau>0$ is a capacity threshold, $\ell(\cdot):[\tau,\infty)\rightarrow 
\mathbb{R}_{\ge0}$ governs large-residual penalization, and the factor $\frac{\tau}{\ell'(\tau)}$ ensures continuous differentiability at $|\nu|=\tau$, with matching value $\frac{\tau^2}{2}$ and derivative $\tau$ from both sides.
The loss $\ell_{\tau}$ reduces to MSE when $|\nu|\le\tau$, preserving 
sensitivity to local prediction errors. For $|\nu|>\tau$, its growth is 
governed by $\ell(\cdot)$, ensuring sub-quadratic penalization of large residuals and preventing large deviations from being overfitted, leaving their treatment to complementary structural components.
Let $\widehat{\theta}_{\text{reg}}^{\tau}$ denote the minimizer of the following empirical risk:
\begin{align}\label{eq.emp_risk}
    \Re^{\tau}_{\text{reg}}(\theta_{\text{reg}})\triangleq\frac{1}{N}\sum_{i=1}^{N}\ell_{\tau}(\langle \mathrm{Z}_i,\theta_{\text{reg}}\rangle -\widetilde{\mathrm{Y}}_i).
\end{align} 
We next identify conditions on $\ell(\cdot)$ under which $\widehat{\theta}_{\text{reg}}^{\tau}$  admits a generalization guarantee.

\begin{assumption}\label{assumption.ell}
    Assume that $\ell(\cdot)$ is convex, twice differentiable, and satisfies 
    $\ell'(t){>} 0$ on $[\tau,\infty)$. Moreover, $\ell'(t)/t$ is non-increasing 
    on $[\tau,\infty)$, or equivalently, $t\ell''(t)-\ell'(t)\le 0$ for all 
    $t\ge \tau$, and $\ell'(t)\le t^\kappa$ for all $t\ge \tau$, where $\kappa\ge 0$ is the constant specified in Condition~\ref{appdx.regularity_condition} 
    \ref{regu.3}\ref{regu.3.b}.
\end{assumption}

Intuitively, the non-increasing $\ell'(t)/t$ condition ensures that the effective 
penalization per unit residual decreases as residual magnitude grows, 
providing robustness to large deviations without discarding them entirely.
Common choices satisfying this assumption include $\ell(t)=t^q$ for $1\le q < 2$ and $t\ge\tau$. Under mild regularity conditions, we establish the following generalization guarantee for $\widehat{\theta}_{\text{reg}}^{\tau}$. Denote 
$\Sigma\triangleq \mathbb{E}(\mathrm{Z}\mathrm{Z}^\top)$, and for any 
$u\in\mathbb{R}^d$, let $\|u\|_{\Sigma}\triangleq\|\Sigma^{1/2}u\|_2$.

\begin{theorem}\label{thm.robust_main}
Assume that the regularity conditions in Appendix \ref{appdx.regularity_condition} hold, and that the function $\ell(\cdot)$ in  \eqref{eq.piecewise_loss} satisfies Assumption \ref{assumption.ell}. Suppose that $d>10$, and $\tau \ge \mathscr{C}_1(\sqrt{d}+\sigma)$, where $\sigma>0$ is defined in Condition \ref{regu.2} of Appendix \ref{appdx.regularity_condition} and $\mathscr{C}_1>0$ is a constant. Furthermore, for any $s>0$, assume that $\max\!\left(\tau, \frac{\tau}{\ell'(\tau)}\right)\le \sigma \sqrt{\frac{N}{d+s}}$, and $N > \mathscr{C}_2(d+s)$ for some constant $\mathscr{C}_2>0$. Then, 
\begin{enumerate}[label=(\roman*)]
    \item \label{thm.robust_main_1} with probability at least $1-5e^{-s}$, $\|\widehat{\theta}_{\text{reg}}^{\tau}-\theta_{\text{reg}}^*\|_{\Sigma}\le \mathscr{C}^\dagger\sqrt{\frac{d+s}{N}}$ for some constant $\mathscr{C}^\dagger>0$;
    \item \label{thm.robust_main_2} with probability at least $1-7e^{-s}$, for any future input $\mathrm{X}_{\text{new}}$ with feature $\mathrm{Z}_{\text{new}}\triangleq\varphi(\mathrm{X}_{\text{new}})$, the associated prediction error satisfies: $\text{err}\triangleq \left|\langle\mathrm{Z}_{\text{new}},\widehat{\theta}_{\text{reg}}^{\tau}\rangle-\langle\mathrm{Z}_{\text{new}},\theta_{\text{reg}}^*\rangle\right|\le \mathscr{C}^{\ddagger}\sqrt{\frac{ds+s^2}{N}}$ for some constant $\mathscr{C}^{\ddagger}>0$.
\end{enumerate}
\end{theorem}
% Assume regularity conditions in Appendix \ref{appdx.regularity_condition}, and assume that function $\ell(\cdot)$ in the piecewise loss (\ref{eq.piecewise_loss}) satisfies the assumptions in Assumption \ref{assumption.ell}. For any $s>0$, suppose that $d>10$, $\max(\tau, \tau/\ell'(\tau))\le\sigma\sqrt{N/(d+s)}$, 
% $\tau\ge\mathscr{C}_1(\sqrt{d}+\sigma)$ for some constant $\mathscr{C}_1>0$, where $\sigma>0$ is defined in Condition \ref{regu.2} of  Appendix \ref{appdx.regularity_condition}. 
% Further assume that $N>\mathscr{C}_2(d+s)$ and $\lambda\le \mathscr{C}_3\sqrt{\frac{d+s}{N}}$ for some constants $\mathscr{C}_2>0$ and $\mathscr{C}_3>0$. Then,  

\begin{remark}
A distributionally robust extension of Theorem~\ref{thm.robust_main}, 
which accounts for distributional uncertainty induced by noisy labels, 
is presented in Theorem~\ref{thm.robust_main_dro} in 
Appendix~\ref{appdx.dro_extension}. Theoretical limitations and future 
directions are discussed in Appendix~\ref{appdx.limitation}.
\end{remark}

{Theorem~\ref{thm.robust_main} establishes non-asymptotic guarantees for
both parameter estimation and prediction under heterogeneous label noise. In particular, the parameter estimation error is bounded at the rate $\mathcal{O}(\sqrt{(d+s)/N})$, while the prediction error admits a bound of order $\mathcal{O}(\sqrt{(ds+s^2)/N})$.}
% Theorem~\ref{thm.robust_main} establishes that when $s=\mathcal{O}(d)$, both  the parameter estimation error and prediction error decay at the near-optimal rate $\mathcal{O}(\sqrt{(d+s)/N})$, demonstrating robust statistical guarantees under heterogeneous label noise. 
Beyond these statistical guarantees, for any $\ell(\cdot)$ satisfying 
Assumption~\ref{assumption.ell}, the corresponding piecewise loss $\ell_\tau$ in \eqref{eq.piecewise_loss} admits an infimal convolution representation, which  provides a principled bridge between the loss design and the fine-grained residual modeling mechanism in our framework, as detailed in Section~\ref{sec.residual}.

\subsubsection{Residual-Based Reformulation}\label{sec.residual}

Building on the piecewise loss $\ell_{\tau}(\cdot)$, we derive an equivalent reformulation 
of the empirical objective $\Re^{\tau}_{\text{reg}}$ in \eqref{eq.emp_risk} 
that explicitly separates instance-specific label deviations from the 
predictive signal, thus providing the theoretical foundation 
for the residual head design developed in Section~\ref{sec.residual_design}.

\begin{proposition}
\label{prop:infimal_convolution}
Let $\ell:[\tau,\infty)\to\mathbb{R}_{\ge 0}$ be convex and differentiable{, with $\ell'(\tau)>0$. Assume that $\ell'(t)/t$ is strictly decreasing on $[\tau,\infty)$.}
Then $\ell_\tau(\cdot)$ in \eqref{eq.piecewise_loss} is convex and 
admits the infimal convolution representation
\begin{equation}
    \ell_\tau(\nu)
    =
    \inf_{\xi\in\mathbb{R}}
    \left\{
    \frac{1}{2}\xi^2+\phi_\tau(\nu-\xi)
    \right\},
\end{equation}
where $\phi_\tau:\mathbb{R}\to\mathbb{R}_{\ge 0}$ is defined via the following 
parametric characterization: for any $t\ge\tau$, let 
$u = t-\frac{\tau\ell'(t)}{\ell'(\tau)}$; then
\begin{equation}
    \phi_\tau(u) 
    = \frac{\tau}{\ell'(\tau)}\bigl\{\ell(t)-\ell(\tau)\bigr\}
    +\frac{\tau^2}{2}
    -\frac{1}{2}\left\{\frac{\tau\ell'(t)}{\ell'(\tau)}\right\}^2.
\end{equation}
{The mapping $t\mapsto u$ is continuous and strictly increasing from $[\tau,\infty)$ onto $[0,\infty)$, so that each $u\geq 0$ corresponds to a unique $t\geq\tau$. For $u<0$, define $\phi_\tau(u)=\phi_\tau(-u)$. Moreover, $\phi_\tau$ is convex, even, and satisfies $\phi_\tau(0)=0$.}
% and $\phi_\tau(-u)=\phi_\tau(u)$ for $u<0$. The mapping $t\mapsto u$ is 
% injective and covers all $u\ge 0$, so that $\phi_\tau$ is well-defined on 
% $\mathbb{R}$. Moreover, $\phi_\tau$ satisfies $\phi_\tau(0)=0$, 
% $\phi_\tau(-u)=\phi_\tau(u)$, and is convex on $\mathbb{R}$.
\end{proposition}

\begin{remark}
When $\ell(t)=t$ for $t\ge\tau$, the piecewise loss $\ell_\tau$ in 
\eqref{eq.piecewise_loss} reduces to the Huber loss with threshold $\tau$. 
In this case, $\phi_\tau(u)=\tau|u|$, yielding 
$\ell_\tau(\nu) = \inf_{\xi\in\mathbb{R}}\left\{\frac{1}{2}\xi^2 
+ \tau|\nu-\xi|\right\}$.
\end{remark}

With Proposition \ref{prop:infimal_convolution}, minimizing the empirical risk $\Re^{\tau}_{\text{reg}}(\theta_{\text{reg}})$ in (\ref{eq.emp_risk}) over $\theta_{\text{reg}}$ becomes equivalent to the following joint minimization over $\theta_{\text{reg}}$ and $\boldsymbol{\xi}$:
\begin{align}\label{eq.loss_prediction_layer}
    {\Re^{\tau}_{\text{reg}}}\left(\theta_{\text{reg}},\boldsymbol{\xi}\right)
    \triangleq&\frac{1}{N}\bigg\{\sum_{i=1}^{N}\underbrace{\frac{1}{2}\xi^2_i}_{\text{residual capacity penalty}}+\underbrace{\phi_\tau\left(\widetilde{\mathrm{Y}}_i-\langle \mathrm{Z}_i,\theta_{\text{reg}}\rangle-\xi_i\right)}_{\text{explicit residual learning}}\bigg\},
\end{align}
where $\boldsymbol{\xi}\triangleq\{\xi_i\}_{i=1}^{N}$, and each auxiliary variable 
$\xi_i\in\mathbb{R}$ captures the instance-specific label deviations, with the penalty $\frac{1}{2}\xi^2_i$ discouraging large residuals and preventing absorption of predictive signals. This decomposition allows the regression model to focus on the underlying predictive mapping while delegating such deviations to $\xi_i$, preventing noise from contaminating global model parameters. 

% The reformulation in \eqref{eq.loss_prediction_layer} provides an explicit 
% decomposition of each prediction error into two components: a structured part 
% governed by the regression model $\theta_{\text{reg}}$, and an instance-specific 
% residual captured by $\xi_i$. This decomposition allows the regression model 
% to focus on learning the underlying predictive mapping, while delegating 
% instance-specific label deviations to the auxiliary variables $\xi_i$, 
% preventing noise from contaminating the global model parameters. This 
% perspective forms the basis for the fine-grained residual modeling mechanism 
% developed in the next subsection.

\subsubsection{Residual Modeling Design}\label{sec.residual_design}

The reformulation in \eqref{eq.loss_prediction_layer} naturally suggests modeling 
$\{\xi_i\}$ as instance-specific residuals, but directly optimizing over 
free variables risks degenerate solutions in which residuals absorb  
prediction signals, leaving the regression model under-trained. 
To prevent this, we parameterize the residual variables through a 
learnable function on the feature representation. 
{For the algorithmic implementation, we extend the fixed feature
representation considered in Sections \ref{sec.robust_loss_family}--\ref{sec.residual} to a learnable feature mapping  $\varphi(\cdot;\theta_{\text{f}}):\mathcal{X}\rightarrow\mathcal{Z}$ parameterized by $\theta_{\text{f}}$.}
Specifically, we 
introduce {a lightweight residual head 
$\mathdutchcal{h}_{\mathrm{res}}(\cdot;\theta_{\mathrm{res}})$ built on top of the feature extractor $\varphi(\cdot;\theta_{\mathrm{f}})$, with $\theta_{\mathrm{res}}$ denoting the parameters of the residual head;} 
and define
\begin{align}\label{eq.residual_output}
    \xi_i(\theta_{\mathrm{res}})
    \triangleq 
    \tau \cdot \tanh\big(\mathdutchcal{h}_{\mathrm{res}}(\mathrm{Z}_i;\theta_{\mathrm{res}})\big),
    \quad \mathrm{Z}_i \triangleq {\varphi(\mathrm{X}_i;\theta_{\text{f}})},
\end{align}
which enforces the capacity constraint $\xi_i \in (-\tau,\tau)$. 
% Here, $\text{tanh}(\cdot)$ denotes the hyperbolic tangent function, though other bounded nonlinear activations (e.g., sigmoid) could also be used. 
Under this parameterization, residual correction is tied to the learned feature representation rather than treated as independent degrees of freedom, ensuring that residual correction is consistent with the learned structure of the data.

{Motivated by the residual decomposition in \eqref{eq.loss_prediction_layer}, we train the residual head using the following pseudo-label-guided objective:}
\begin{align}\label{eq.residual_train}
    \frac{1}{N}\sum_{i=1}^{N}
    \left\{
    \frac{1}{2}\xi^2_i(\theta_{\mathrm{res}})
    +\phi_\tau\big((\widetilde{\mathrm{Y}}_i-\mathrm{Y}^*_i)
    -\xi_i(\theta_{\mathrm{res}})\big)
    \right\},
\end{align}
where $\mathrm{Y}^*_i$ denotes a robust pseudo label whose construction is specified in (\ref{eq.EMA_pseudo_label}) of Appendix \ref{appdx.implement_details}. The residual head targets the deviation 
$\widetilde{\mathrm{Y}}_i - \mathrm{Y}^*_i$, decoupling residual correction 
from the predictive signal. The residual parameters $\theta_{\mathrm{res}}$ 
are updated in an alternating manner with other model parameters, which prevents the residual head from co-adapting with the regression model and ensures that $\xi_i$ captures only instance-specific deviations rather than replacing the role of the regression model.

Finally, rather than fixing $\tau$ as a hyperparameter, we adapt it each epoch as 
a scaled median absolute deviation (MAD) \cite{rousseeuw1993alternatives} of the current prediction 
residuals $\{\widetilde{\mathrm{Y}}_i - {\mathrm{Y}^*_i}\}_{i=1}^N$ (detailed in Appendix \ref{appdx.implement_details}).
This adaptive $\tau$ defines a principled boundary between the two granularities: deviations within $(-\tau,\tau)$ are handled through instance-level residual 
refinement, while larger deviations, which cannot be reliably captured at 
the instance level, are instead addressed by the coarse-grained structural 
mechanism introduced in Section~\ref{sec.feature_structuring}, as illustrated in Figure~\ref{fig:tau_distribution}.

\subsection{Coarse-Grained Structuring: Contrastive  Regularization across Bins}\label{sec.feature_structuring}

For large deviations, where exact label values become unreliable, direct supervision may mislead the model and induce structural distortion in the feature space: representations no longer reflect label proximity, and the global geometry degrades.
While the fine-grained module performs instance-level correction within $(-\tau,\tau)$ under label supervision, it does not directly enforce a globally consistent, order-preserving structure aligned with label proximity.
Accordingly, we introduce a coarse-grained structuring mechanism that operates at the distribution level, enforcing an order-preserving geometry across bins of the label space.

\subsubsection{Discretization of Label Space}\label{sec.cls_head}

We partition the label space $\mathcal{Y}$ into $K$ equal-width bins $\mathcal{Y}=\cup_{k=1}^{K}
\mathcal{Y}_k$ (detailed in Appendix~\ref{appdx.implement_details}). 
For $k\in[K]$ with $[K]\triangleq\{1,\ldots,K\}$, let $\mathcal{I}_k$ index samples whose labels fall into bin $k$, where after a warm-up phase bin assignment is based on the robust pseudo labels $\{\mathrm{Y}^*_i\}_{i=1}^{N}$ (detailed in Appendix~\ref{appdx.implement_details}) rather than the original noisy labels. Define 
$\mathcal{D}_k\triangleq\{(\mathrm{X}_i,\widetilde{\mathrm{Y}}_i):
i\in\mathcal{I}_k\}$. Based on this discretization, we introduce a histogram 
head $\mathdutchcal{h}_{\text{hist}}(\cdot;\theta_{\text{hist}})$ parameterized by $\theta_{\text{hist}}$ on top of the feature extractor $\varphi(\cdot;\theta_{\text{f}})$, producing logits $s_i = \mathdutchcal{h}_{\text{hist}}
(\varphi(\mathrm{X}_i;\theta_{\text{f}});\theta_{\text{hist}})\in\mathbb{R}^K$ 
and bin softmax probabilities $\mathdutchcal{p}_i=\mathrm{softmax}(s_i)\in\Delta^{K-1}$ over the $K$-simplex $\Delta^{K-1}$.

The bin structure serves two purposes. First, following prior 
work~\cite{kim2024sample,stewart2023regression}, 
classification-based objectives encourage greater feature diversity than 
regression-only training; accordingly, we include a cross-entropy loss over 
bin assignments to mitigate representation collapse under noisy supervision:
\begin{align}\label{eq.loss_cls}
\Re_{\text{ce}}(\theta_{\text{hist}},\theta_{\text{f}})
=-\frac{1}{N}\sum_{k\in[K]}\sum_{i\in\mathcal{I}_k}\log\mathdutchcal{p}_i[k],
\end{align}
Second, and more centrally, the bin structure 
provides the foundation for the cross-bin contrastive regularization below, which constitutes the core coarse-grained structuring mechanism.

\subsubsection{Order-Preserving Cross-Bin Contrastive Objective}\label{sec.uot}

We construct contrastive relations at the level of bin-wise feature 
distributions. For each bin $k$, define the empirical feature distribution
$\widehat{P}^{(k)}_{z}(\theta_{\text{f}})=\frac{1}{|\mathcal{I}_k|}
\sum_{i\in\mathcal{I}_k}\delta_{\mathrm{Z}_i(\theta_{\text{f}})}$,
where $\mathrm{Z}_i(\theta_{\text{f}})\triangleq\varphi(\mathrm{X}_i;
\theta_{\text{f}})$, $|\mathcal{I}_k|$ denotes the cardinality of $\mathcal{I}_k$, and {$\delta_v$ denotes the Dirac measure concentrated at $v$}. 
For each $k\in[K]$, we form a triplet $\big(\widehat{P}^{(k)}_{z}(\theta_{\text{f}}), \widehat{P}^{(k_{\text{pos}})}_{z}(\theta_{\text{f}}), \widehat{P}^{(k_{\text{neg}})}_{z}(\theta_{\text{f}})\big)$, where $k_{\text{pos}}$ and $k_{\text{neg}}$ {denote the} indices of nearby and distant bins, respectively, {satisfying} $|k-k_{\text{pos}}|<|k-k_{\text{neg}}|$. 

%We minimize the distributional distance between $\widehat{P}^{(k)}_{z}(\theta_{\text{f}})$ and positive distributions, while maximizing its distance to negative ones. 

Next, we define positive and negative bin neighborhoods based on label proximity. Specifically, for a positive integer $R$ and $k\in[K]$, let 
$\mathcal{N}^{+}_{k}\triangleq\{k-R,\ldots,k-1,k+1,\ldots,k+R\}\cap [K]$ index the nearest neighboring bins of bin $k$, and 
$\mathcal{N}^{-}_{k}\triangleq\{k-2R,\ldots,k-(R+1),k+(R+1),\ldots,k+2R\}\cap[K]$ 
denote the indices of farther bins. 
The contrastive losses are then computed as weighted averages:
\begingroup%\small
\begin{align}
    \Re_{\text{pos}}
    \triangleq&\frac{1}{K}\sum_{k=1}^{K}\sum_{k'\in\mathcal{N}^+_k}\frac{1/|k-k'|}{\sum_{j\in\mathcal{N}^+_k}1/|k-j|}\,\mathscr{d}\big(\widehat{P}^{(k)}_{z},\widehat{P}^{(k')}_{z}\big);\label{eq.loss_feature_pos}\\
    \Re_{\text{neg}}
    \triangleq&-\frac{1}{K}\sum_{k=1}^{K}\sum_{k'\in\mathcal{N}^-_k}\frac{(k-k')^2}{\sum_{j\in\mathcal{N}^-_k}(k-j)^2}\,
    \mathscr{d}\big(\widehat{P}^{(k)}_{z},\widehat{P}^{(k')}_{z}\big).\label{eq.loss_feature_neg}
\end{align}
\endgroup
where $\theta_{\text{f}}$-dependence of $\Re_{\text{pos}}$, $\Re_{\text{neg}}$, $\widehat{P}^{(k)}_{z}$, and $\widehat{P}^{(k')}_{z}$ is omitted temporarily for brevity, and {$\mathscr{d}(P,Q)$ denotes a discrepancy measure between distributions $P$ and $Q$}. Here, the positive weights emphasize closer neighbors, {while} the negative weights penalize more distant bin pairs proportionally to their label separation. 
We set $R=1$ by default. {In our implementation, we employ} an unbalanced optimal transport (UOT) cost (Definition \ref{def.UOT}) under a Gaussian approximation~\cite{janati2021advances} {as the discrepancy measure}, providing further robustness to bin misassignments induced by label noise.

This objective enforces an order-preserving feature geometry: representations from nearby bins are pulled together to preserve label continuity, while distant bins are pushed apart to maintain discriminability. Unlike instance-level residual refinement, this coarse-grained structuring mechanism directly targets the structural corruption induced by large-magnitude noise, namely the deviations that lie beyond the capacity threshold $\tau$ and fall outside the scope of fine-grained correction. 
%Together, the two mechanisms operate at complementary granularities, motivating the unified framework developed in Section~\ref{sec.bridge}.

\subsection{Unified Framework: Bridging across Granularities}\label{sec.bridge}

Let {$\vartheta\triangleq(\theta_{\text{reg}}^\top, \theta_{\text{hist}}^\top, \theta_{\text{res}}^\top, \theta_{\text{f}}^\top)^\top$} denote all trainable parameters. Combining (\ref{eq.loss_prediction_layer}), (\ref{eq.loss_cls}), (\ref{eq.loss_feature_pos}) and (\ref{eq.loss_feature_neg}), and suppressing parameter dependence for notational simplicity, the joint training objective is
\begin{align}\label{eq.loss_all}
    \mathcal{L}(\vartheta)=\lambda_{\text{reg}}\Re_{\text{reg}}^{\tau}+\lambda_{\text{ce}}\Re_{\text{ce}}+\lambda_{\text{pos}}\Re_{\text{pos}}+\lambda_{\text{neg}}\Re_{\text{neg}},
\end{align}
where $\lambda_{\text{reg}}$, $\lambda_{\text{ce}}$, $\lambda_{\text{pos}}$, and $\lambda_{\text{neg}}$ are positive hyperparameters balancing the respective components. {In practice, we fix $\lambda_{\text{ce}}=1.0$ and the ratio $\lambda_{\text{neg}}/\lambda_{\text{pos}}=0.5$, reducing the tuning to two effective hyperparameters: $\lambda_{\text{reg}}$ and $\lambda_{\text{pos}}$.} Training details and Algorithm~\ref{alg:GRAIN} are deferred to 
Appendix~\ref{appdx.implement_details}.

The key component bridging the two granularities is the histogram head, which generates robust pseudo labels $\{\mathrm{Y}^*_i\}$ from the shared feature representations and couples the two modules through a unified supervisory signal. Each pseudo label is initialized from the observed noisy label $\widetilde{\mathrm{Y}}_i$ and updated each epoch via exponential moving average (EMA) of the histogram head's weighted bin-center predictions, {retaining the smallest set of highest-probability bins whose cumulative probability mass reaches at least 68\%} (motivated by one-standard-deviation Gaussian coverage), with confidence computed as $1-K^*/K$, where $K^*$ is the number of retained bins; full details are given in (\ref{eq.68_set})-(\ref{eq.EMA_conf}) of Appendix~\ref{appdx.implement_details}.

% The two mechanisms in GRAIN operate at complementary granularities, as 
% illustrated in Figure~\ref{fig:GRAIN_overview}. The fine-grained module 
% corrects instance-level label deviations through explicit residual modeling, 
% while the coarse-grained module restores global feature geometry through 
% order-preserving contrastive regularization. A key component bridging these 
% two levels is the histogram head, which generates robust pseudo labels 
% $\{\mathrm{Y}^*_i\}_{i=1}^N$ from the shared feature representations and 
% couples the two modules through a unified supervisory signal. Specifically, 
% each pseudo label is derived via a weighted average over the most confident 
% bin predictions (top $68\%$, motivated by one-standard-deviation Gaussian 
% coverage), followed by an exponential moving average across epochs for 
% temporal stability; full details are given in 
% Appendix~\ref{appdx.implement_details}.

The pseudo labels $\{\mathrm{Y}^*_i\}$ from the histogram head couple the two modules through a shared supervisory signal. In the fine-grained module, they provide residual 
targets $\widetilde{\mathrm{Y}}_i-\mathrm{Y}^*_i$ in 
\eqref{eq.residual_train}, decoupling residual estimation from the noisy 
label directly. In the coarse-grained module, they stabilize bin assignments under noisy supervision, improving the reliability of the contrastive objective. Through these roles, the two modules jointly produce structured and refined feature representations: the coarse-grained module enforces global order in the feature space, establishing a well-organized geometric structure aligned with label proximity, while fine-grained refinement performs precise instance-level correction within this structured space. The resulting representations, in turn, improve the quality of histogram head predictions, yielding more reliable pseudo labels in subsequent epochs. This closed-loop interaction between pseudo-label estimation and feature learning allows each component to progressively reinforce the others, enabling coordinated handling of both global structural distortion and instance-level label noise throughout training.

\section{Experiment Details}

\subsection{Experimental Setup}\label{sec.main_exp_setup}

\textbf{Evaluation Metrics.} We evaluate model performance using three standard regression metrics: 
mean absolute error (MAE), 
root mean squared error (RMSE), 
% mean absolute percentage error (MAPE), 
% and Pearson correlation coefficient ($R$), 
and coefficient of determination ($R^2$), 
where lower values of MAE and RMSE indicate better predictive accuracy, and higher values of $R^2$ indicate a higher proportion of variance explained by the model. In the reported tables, we show the desired direction using ``$\downarrow$'' for metrics to minimize and ``$\uparrow$'' for metrics to maximize. The detailed definitions and computation of each metric are deferred to Appendix~\ref{appdx.exp_detail}.

\textbf{Datasets.} We conduct experiments on four datasets spanning three different regression tasks: \textbf{Age estimation} on UTKFace \cite{zhifei2017cvpr} and APPA-REAL \cite{clapes2018apparent}, \textbf{Head-pose estimation} on Biwi Kinect \cite{fanelli2013random}, and \textbf{Bone-age assessment} on RSNA \cite{halabi2019rsna} combined with DHA \cite{pietka2001computer}. For UTKFace and Biwi Kinect, we simulate label noise by adding Gaussian perturbations with varying maximal standard deviations $\text{std}_{\text{max}}$. Here, $\text{std}_{\text{max}}$ controls the magnitude of the injected noise during generation, rather than representing the actual standard deviation of the resulting noisy labels; higher values correspond to stronger noise. In our experiments, we consider three levels, $\text{std}_{\text{max}}\in\{30, 60, 90\}$. In contrast, APPA-REAL and the Bone-Age datasets contain inherent, real-world label noise. Additional details on dataset preprocessing and noise simulation are provided in Appendix~\ref{appdx.exp_detail}.

\textbf{Baselines.} We consider the following baselines for regression tasks: (i) Vanilla, which directly trains with MSE; (ii) SuperLoss \cite{castells2020superloss}; (iii) RDI \cite{husimple}; (iv) AUX \cite{husimple}; and (v) C-Mixup \cite{yao2022c}. We also include several classical methods originally designed for noisy-label classification that can be adapted to regression: (1) CDR \cite{xia2020robust}, (2) CoTeaching \cite{han2018co}, (3) CNLCU-S \cite{xiasample}, (4) CNLCU-H \cite{xiasample}, (5) Selfie \cite{song2019selfie}, and (6) CoSelfie \cite{song2019selfie}. Detailed descriptions of all methods are provided in Appendix~\ref{appdx.exp_detail}. For fair comparison, all baselines are implemented in two ways: (a) using the original regression loss directly, and (b) with a histogram head following the same binning strategy as GRAIN. In the reported tables, results from the latter implementation are indicated by ``BIN''. Baselines that simultaneously train two models are further indicated by the superscript symbol ``$\dagger$''. For both GRAIN and all baselines, we adopt standard (basic) data augmentation during training, including random cropping, horizontal flipping, and rotation. Results obtained with slightly stronger data augmentations, which further add color jitter and mixed image transformations, are also reported and denoted as ``GRAIN-{\small AUG}''.

% SuperLoss \cite{castells2020superloss} \\
% RDI \cite{husimple} \\
% AUX \cite{husimple} \\
% C-Mixup \cite{yao2022c} \\
% CDR \cite{xia2020robust} \\
% SIGUA \cite{han2020sigua} \\
% Selfie \cite{song2019selfie} \\
% CoTeaching $^\dagger$ \cite{han2018co} \\
% CNLCU-S $^\dagger$ \cite{xiasample} \\
% CNLCU-H $^\dagger$ \cite{xiasample} \\
% CoSelfie $^\dagger$ \cite{song2019selfie} \\

% ======================================================================

% ==============================================================================
% ==============================================================================
%                        [GRAIN] UTKFace - Gaussian
% ==============================================================================
% ==============================================================================

\begin{table*}[t]
\centering
\caption{
Mean $\pm$ standard deviation on UTKFace under simulated Gaussian label noise with corruption levels controlled by $\mathrm{std}_{\max}$, and on Bone-Age. The best baseline for each metric is highlighted in {\color{blue}blue}, and results from GRAIN and GRAIN-{\small AUG} are presented in \textbf{boldface}.
}
\label{tab:utkface_gaussian_BoneAge}
\setlength{\tabcolsep}{2pt} 
\resizebox{\textwidth}{!}{
\begin{tabular}{l|ccc|ccc|ccc||ccc} % !{\vrule width 1pt}
\toprule
\multirow{2}{*}{\bf Method} 
& \multicolumn{3}{c|}{\bf UTKFace ($\text{\bf std}_{\text{\bf max}}\text{\bf  = 30}$)} 
& \multicolumn{3}{c|}{\bf UTKFace ($\text{\bf std}_{\text{\bf max}}\text{\bf  = 60}$)} 
& \multicolumn{3}{c||}{\bf UTKFace ($\text{\bf std}_{\text{\bf max}}\text{\bf  = 90}$)} 
& \multicolumn{3}{c}{\bf Bone-Age} \\
\cmidrule(lr){2-4}
\cmidrule(lr){5-7}
\cmidrule(lr){8-10}
\cmidrule(lr){11-13}
 & \bf MAE $\downarrow$ & \bf RMSE $\downarrow$ & $\boldsymbol{R}^2$ $\uparrow$ 
 & \bf MAE $\downarrow$ & \bf RMSE $\downarrow$ & $\boldsymbol{R}^2$ $\uparrow$ 
 & \bf MAE $\downarrow$ & \bf RMSE $\downarrow$ & $\boldsymbol{R}^2$ $\uparrow$ 
 & \bf MAE $\downarrow$ & \bf RMSE $\downarrow$ & $\boldsymbol{R}^2$ $\uparrow$ \\
\midrule

% ==============================================================================
%                       [GRAIN] UTKFace - Gaussian - MSE
% ==============================================================================

Vanilla 
& $8.13_{\pm 0.17}$ & $11.20_{\pm 0.12}$ & $0.70_{\pm 0.01}$ 

& $9.94_{\pm 0.19}$ & $13.07_{\pm 0.22}$ & $0.59_{\pm 0.01}$ 

& $10.25_{\pm 0.29}$ & $13.43_{\pm 0.31}$ & $0.56_{\pm 0.02}$  

& $13.56_{\pm 0.25}$ & $17.44_{\pm 0.36}$ & {\color{blue}$0.91_{\pm 0.00}$}  \\

SuperLoss 
& $7.69_{\pm 0.19}$ & $11.43_{\pm 0.41}$ & $0.68_{\pm 0.02}$ 

& $8.18_{\pm 0.12}$ & $11.95_{\pm 0.32}$ & $0.66_{\pm 0.02}$ 

& \ $8.23_{\pm 0.24}$ & $11.97_{\pm 0.33}$ & $0.65_{\pm 0.02}$  

& $13.59_{\pm 0.11}$ & $17.33_{\pm 0.18}$ & {\color{blue}$0.91_{\pm 0.00}$}  \\

RDI 
& $8.09_{\pm 0.11}$ & $11.31_{\pm 0.21}$ & $0.69_{\pm 0.01}$ 

& $9.71_{\pm 0.29}$ & $12.86_{\pm 0.23}$ & $0.60_{\pm 0.01}$ 

& $10.03_{\pm 0.26}$ & $13.31_{\pm 0.32}$ & $0.57_{\pm 0.02}$ 

& $13.26_{\pm 0.34}$ & $17.23_{\pm 0.49}$ & {\color{blue}$0.91_{\pm 0.01}$} \\

AUX 
& $8.19_{\pm 0.16}$ & $11.40_{\pm 0.31}$ & $0.69_{\pm 0.02}$ 

& $9.74_{\pm 0.24}$ & $12.90_{\pm 0.16}$ & $0.60_{\pm 0.01}$ 

& $10.23_{\pm 0.36}$ & $13.51_{\pm 0.29}$ & $0.56_{\pm 0.02}$ 

& $13.35_{\pm 0.11}$ & $17.21_{\pm 0.13}$  & $0.91_{\pm 0.00}$  \\

C-Mixup 
& $7.56_{\pm 0.16}$ & {\color{blue}$10.40_{\pm 0.24}$} & {\color{blue}$0.74_{\pm 0.01}$} 

& $9.69_{\pm 0.15}$ & $13.07_{\pm 0.32}$ & $0.59_{\pm 0.02}$ 

& $10.11_{\pm 0.24}$ & $13.45_{\pm 0.30}$ & $0.56_{\pm 0.02}$

& $13.51_{\pm 0.17}$ & $17.52_{\pm 0.23}$ & {\color{blue}$0.91_{\pm 0.00}$} \\

% SPR (MSE) & \\

CDR 
& $8.21_{\pm 0.15}$ & $11.49_{\pm 0.07}$ & $0.68_{\pm 0.00}$ 

& $9.88_{\pm 0.15}$ & $13.06_{\pm 0.23}$ & $0.59_{\pm 0.02}$ 

& $10.39_{\pm 0.25}$ & $13.71_{\pm 0.18}$ & $0.55_{\pm 0.01}$ 

& $13.34_{\pm 0.14}$ & $17.34_{\pm 0.16}$ & {\color{blue}$0.91_{\pm 0.00}$} \\

% SIGUA & \\

Selfie 
& $8.09_{\pm 0.06}$ & $12.32_{\pm 0.24}$ & $0.63_{\pm 0.01}$ 

& $8.30_{\pm 0.12}$ & $12.70_{\pm 0.29}$ & $0.61_{\pm 0.02}$ 

& \ $8.38_{\pm 0.29}$ & $12.77_{\pm 0.45}$ & $0.61_{\pm 0.03}$ 

& $15.43_{\pm 0.76}$ & $19.65_{\pm 0.86}$ & $0.88_{\pm 0.01}$\\

CoTeaching $^\dagger$ 
& $7.84_{\pm 0.16}$ & $11.95_{\pm 0.37}$ & $0.66_{\pm 0.02}$ 

& $7.98_{\pm 0.09}$ & $12.18_{\pm 0.28}$ & $0.64_{\pm 0.02}$ 

& \ $8.28_{\pm 0.29}$ & $12.71_{\pm 0.49}$ & $0.61_{\pm 0.03}$  

& $14.39_{\pm 0.41}$ & $18.36_{\pm 0.70}$ & $0.90_{\pm 0.01}$ \\

CNLCU-S $^\dagger$ 
& $8.16_{\pm 0.15}$ & $12.64_{\pm 0.35}$  & $0.61_{\pm 0.02}$ 

& $8.17_{\pm 0.22}$ & $12.66_{\pm 0.45}$ & $0.61_{\pm 0.03}$ 

& \ $8.48_{\pm 0.32}$ & $13.04_{\pm 0.59}$ & $0.59_{\pm 0.04}$ 

& $15.97_{\pm 0.53}$ & $20.54_{\pm 0.71}$ & $0.87_{\pm 0.01}$ \\

CNLCU-H $^\dagger$ 
& $8.30_{\pm 0.18}$ & $12.83_{\pm 0.37}$ & $0.60_{\pm 0.02}$ 

& $8.31_{\pm 0.15}$ & $12.86_{\pm 0.33}$ & $0.60_{\pm 0.02}$ 

& \ $8.49_{\pm 0.35}$ & $12.99_{\pm 0.58}$ & $0.59_{\pm 0.04}$  

& $15.83_{\pm 0.68}$ & $20.12_{\pm 0.94}$ & $0.88_{\pm 0.01}$\\

CoSelfie $^\dagger$ 
& $7.84_{\pm 0.15}$ & $11.87_{\pm 0.38}$ & $0.66_{\pm 0.02}$  

& $8.03_{\pm 0.22}$ & $12.30_{\pm 0.52}$ & $0.63_{\pm 0.03}$ 

& \ $8.21_{\pm 0.22}$ & $12.47_{\pm 0.34}$ & $0.62_{\pm 0.02}$ 

& $15.50_{\pm 0.46}$ & $19.70_{\pm 0.46}$ & $0.88_{\pm 0.01}$ \\

% Huber & \\

\midrule

% ==============================================================================
%                       [GRAIN] UTKFace - Gaussian - CE
% ==============================================================================

Vanilla {\small(BIN)} 
& $7.79_{\pm 0.09}$ & $11.87_{\pm 0.12}$ & $0.66_{\pm 0.01}$ 

& \ $8.44_{\pm 0.17}$ & $13.12_{\pm 0.30}$ & $0.58_{\pm 0.02}$ 

& $8.46_{\pm 0.28}$ & $13.06_{\pm 0.44}$ & $0.59_{\pm 0.03}$  

& $13.31_{\pm 0.22}$ & $17.43_{\pm 0.53}$ & {\color{blue}$0.91_{\pm 0.01}$}  \\

SuperLoss {\small(BIN)} 
& $7.61_{\pm 0.19}$ & $11.35_{\pm 0.27}$ & $0.69_{\pm 0.01}$ 

& {\color{blue}$7.90_{\pm 0.19}$} & {\color{blue}$11.61_{\pm 0.31}$} & {\color{blue}$0.67_{\pm 0.02}$}  

& \ {\color{blue}$7.95_{\pm 0.19}$} & {\color{blue}$11.65_{\pm 0.26}$} & {\color{blue}$0.67_{\pm 0.01}$}  

& $13.55_{\pm 0.33}$ & $17.41_{\pm 0.33}$ & {\color{blue}$0.91_{\pm 0.00}$} \\

RDI {\small(BIN)} 
& $7.84_{\pm 0.12}$ & $11.89_{\pm 0.17}$ & $0.66_{\pm 0.01}$ 

& $8.41_{\pm 0.10}$ & $12.88_{\pm 0.27}$ & $0.60_{\pm 0.02}$ 

& \ $8.46_{\pm 0.21}$ & $13.29_{\pm 0.45}$ & $0.57_{\pm 0.03}$

& $13.18_{\pm 0.15}$ & {\color{blue}$16.84_{\pm 0.33}$}  & $0.91_{\pm 0.00}$  \\

AUX {\small(BIN)} 
& $7.70_{\pm 0.09}$ & $11.68_{\pm 0.12}$ & $0.67_{\pm 0.01}$  

& $8.35_{\pm 0.10}$ & $12.73_{\pm 0.08}$ & $0.61_{\pm 0.01}$ 

& \ $8.45_{\pm 0.18}$ & $13.20_{\pm 0.25}$ & $0.58_{\pm 0.02}$  

& $13.27_{\pm 0.31}$ & $17.16_{\pm 0.68}$  & {\color{blue}$0.91_{\pm 0.01}$}\\

C-Mixup {\small(BIN)} 
& {\color{blue}$7.44_{\pm 0.25}$} & {\color{blue}$10.40_{\pm 0.36}$} & {\color{blue}$0.74_{\pm 0.02}$} 

& $8.29_{\pm 0.14}$ & $12.84_{\pm 0.32}$ & $0.60_{\pm 0.02}$ 

& \ $8.49_{\pm 0.21}$ & $13.24_{\pm 0.29}$ & $0.58_{\pm 0.02}$ 

& {\color{blue}$13.12_{\pm 0.54}$} & $17.45_{\pm 1.04}$  & {\color{blue}$0.91_{\pm 0.01}$}\\

% SPR (CE) & \\

CDR {\small(BIN)}
&  $7.74_{\pm 0.15}$ & $10.82_{\pm 0.25}$ & {$0.72_{\pm 0.01}$} 

& $8.56_{\pm 0.28}$ & $13.48_{\pm 0.67}$ & $0.56_{\pm 0.04}$

& \ $8.58_{\pm 0.27}$ & $13.16_{\pm 0.33}$ & $0.58_{\pm 0.02}$  

& $13.45_{\pm 0.29}$ & $17.52_{\pm 0.84}$ & {\color{blue}$0.91_{\pm 0.01}$} \\

% SIGUA {\small(BIN)} & \\

Selfie {\small(BIN)} 
& $8.20_{\pm 0.36}$ & $12.50_{\pm 0.66}$ & $0.62_{\pm 0.04}$ 

& $8.44_{\pm 0.47}$ & $12.79_{\pm 0.72}$ & $0.60_{\pm 0.05}$ 

& $9.07_{\pm 0.44}$ & $13.83_{\pm 0.48}$ & $0.54_{\pm 0.03}$   

& $16.35_{\pm 1.13}$ & $20.81_{\pm 1.56}$ & $0.87_{\pm 0.02}$ \\

CoTeaching {\small(BIN)} $^\dagger$ 
& $8.22_{\pm 0.33}$ & $12.66_{\pm 0.61}$ & $0.61_{\pm 0.04}$ 

& $8.29_{\pm 0.32}$ & $12.66_{\pm 0.64}$ & $0.61_{\pm 0.04}$ 

& \ $8.75_{\pm 0.47}$ & $13.26_{\pm 0.53}$ & $0.58_{\pm 0.03}$ 

& $14.58_{\pm 0.47}$ & $18.68_{\pm 0.75}$ & $0.89_{\pm 0.01}$\\

CNLCU-S {\small(BIN)} $^\dagger$ 
& $7.89_{\pm 0.21}$ & $12.19_{\pm 0.44}$ & $0.64_{\pm 0.03}$ 

& $8.29_{\pm 0.60}$ & $12.84_{\pm 1.27}$ & $0.60_{\pm 0.08}$ 

& \ $8.34_{\pm 0.39}$ & $13.04_{\pm 0.59}$ & $0.59_{\pm 0.04}$ 

& $15.96_{\pm 0.79}$ & $20.22_{\pm 1.01}$ & $0.88_{\pm 0.01}$ \\

CNLCU-H {\small(BIN)} $^\dagger$ 
& $7.91_{\pm 0.18}$ & $12.24_{\pm 0.37}$ & $0.64_{\pm 0.02}$ 

& $8.45_{\pm 0.52}$ & $13.13_{\pm 0.91}$ & $0.58_{\pm 0.06}$ 

& \ $8.58_{\pm 0.25}$ & $13.29_{\pm 0.38}$ & $0.57_{\pm 0.02}$ 

& $17.39_{\pm 0.97}$ & $22.29_{\pm 1.37}$ & $0.85_{\pm 0.02}$ \\

CoSelfie {\small(BIN)} $^\dagger$ 
& $8.29_{\pm 0.27}$ & $12.61_{\pm 0.31}$ & $0.62_{\pm 0.02}$ 

& $8.46_{\pm 0.86}$ & $12.90_{\pm 1.33}$ & $0.59_{\pm 0.09}$ 

& \ $8.60_{\pm 0.73}$ & $13.06_{\pm 1.01}$ & $0.59_{\pm 0.06}$ 

& $14.44_{\pm 0.49}$ & $18.62_{\pm 0.56}$ & $0.89_{\pm 0.01}$ \\

\midrule
\bf GRAIN 
& $\bm{7.01_{\pm 0.10}}$ & $\bm{10.28_{\pm 0.24}}$ & $\bm{0.74_{\pm 0.01}}$

& $\bm{7.54_{\pm 0.14}}$ & $\bm{11.10_{\pm 0.33}}$ & $\bm{0.70_{\pm 0.02}}$

% & $\bm{7.60_{\pm 0.16}}$ & $\bm{11.17_{\pm 0.20}}$ & $\bm{0.55_{\pm 0.06}}$ & $\bm{0.84_{\pm 0.01}}$ & $\bm{0.70_{\pm 0.01}}$ 
& $\bm{7.65_{\pm 0.12}}$ & $\bm{11.28_{\pm 0.22}}$ & $\bm{0.70_{\pm 0.01}}$ 

& \bm{$12.85_{\pm 0.26}$} & \bm{$16.55_{\pm 0.32}$} & \bm{$0.92_{\pm 0.00}$} 
\\

\bf GRAIN-{\small{AUG}} 
& $\bm{6.81_{\pm 0.13}}$ & $\bm{9.98_{\pm 0.28}}$ & $\bm{0.76_{\pm 0.01}}$

& $\bm{7.30_{\pm 0.14}}$ & $\bm{10.74_{\pm 0.28}}$ & $\bm{0.72_{\pm 0.01}}$

& $\bm{7.35_{\pm 0.22}}$ & $\bm{10.75_{\pm 0.36}}$ & $\bm{0.72_{\pm 0.02}}$ 

& \bm{$12.80_{\pm 0.22}$} & \bm{$16.40_{\pm 0.29}$} & \bm{$0.92_{\pm 0.00}$}\\

\bottomrule
\end{tabular}
}
\end{table*}

% ======================================================================

\textbf{Training Details.} For all datasets except Bone-Age, we use a ResNet-50 backbone \cite{he2016deep}; for the Bone-Age dataset, we employ DenseNet-121 \cite{huang2017densely}. The initial learning rate is taken as $5\times 10^{-4}$ or $10^{-3}$, and a cosine learning rate scheduler is applied. For all methods, each noisy dataset is split into training and validation sets with a $4:1$ ratio, and model selection is performed based on validation MAE. All experiments are repeated with five random seeds (1–5).
%, and the reported results correspond to the average across these runs.

\subsection{Main Results}\label{sec.main_results}

\textbf{Comparison with Baseline Methods.} Table~\ref{tab:utkface_gaussian_BoneAge} summarizes the results on UTKFace under simulated Gaussian label noise with different corruption levels, and on Bone-Age dataset with real-world label noise, evaluated {using three} metrics. Across nearly all noise levels and metrics, GRAIN and GRAIN-{\small AUG} consistently demonstrate the best performance, showing strong robustness to both synthetic and real-world label noise. Additional results on the Biwi Kinect dataset with synthetic Gaussian noise are reported in Tables~\ref{tab:biwi_kinect_yaw_gaussian}-\ref{tab:biwi_kinect_roll_gaussian}, and results on APPA-REAL with real-world label noise are reported in Table \ref{tab:appa}, all appearing in Appendix~\ref{appdx.exps}; they all exhibit the same patterns.

% Together, these results highlight the effectiveness of GRAIN on both synthetic and real-world noisy-label datasets.

% ======================================================================

\begin{figure}[htbp]
    \centering
    \begin{minipage}[b]{0.9\textwidth}
        \vspace{0pt}
        \centering
        \includegraphics[width=\linewidth]{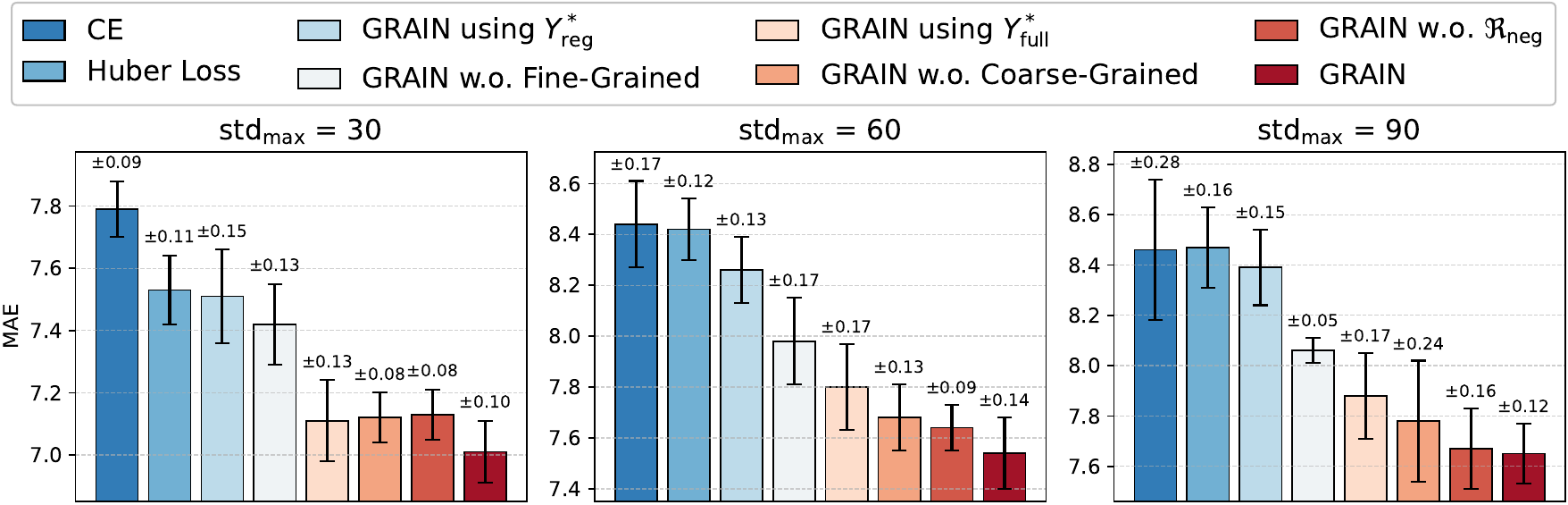}
        \caption{Ablation study results.}
        \label{fig:ablation}
    \end{minipage}
    \hfill
    \begin{minipage}[b]{0.9\textwidth}
        \vspace{12pt}
        \centering
        \includegraphics[width=\linewidth]{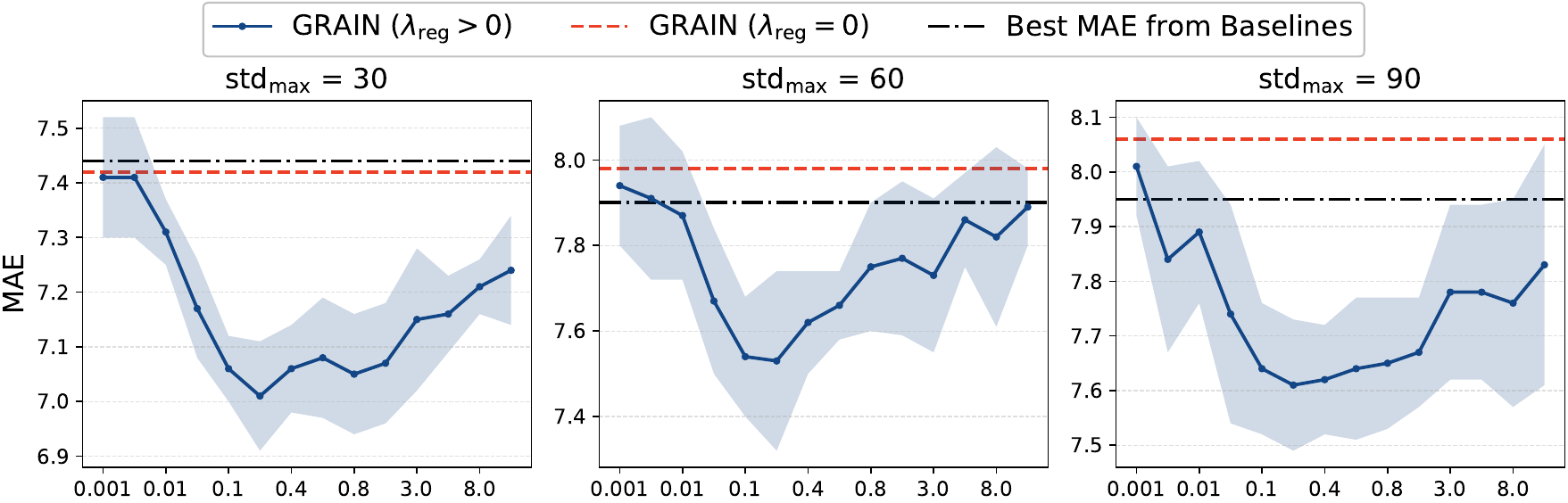}
        \caption{Hyperparameter sensitivity on $\lambda_{\text{reg}}$.}
        \label{fig:hyper_lam_reg_MAE}
    \end{minipage}
\end{figure}

\textbf{Ablation Studies.} We conduct an ablation study to isolate the contributions of key components in GRAIN. Specifically, we consider:
(i) adaptive binning procedure in Appendix \ref{appdx.implement_details} with the CE loss (``CE''); 
(ii) direct training with the Huber loss (``Huber Loss''); 
(iii) training the residual head $\mathdutchcal{h}_{\text{res}}$ using the regression head output in place of $\mathrm{Y}^*_i$ in~(\ref{eq.EMA_pseudo_label}) (``GRAIN using $\mathrm{Y}^*_{\text{reg}}$''); 
(iv) GRAIN without fine-grained residual learning (``GRAIN w.o. Fine-Grained''); (v) computing $\mathrm{Y}^*_i$ in~(\ref{eq.EMA_pseudo_label}) using the full classification softmax output instead of the top 68\% predictions (``GRAIN using $\mathrm{Y}^*_{\text{full}}$''); 
(vi) GRAIN without coarse-grained cross-bin contrastive regularization (``GRAIN w.o. Coarse-Grained''); 
(vii) GRAIN without the negative term $\Re_{\text{neg}}$ (``GRAIN w.o. $\Re_{\text{neg}}$''); 
and (viii) the full GRAIN model (``GRAIN''). Figure~\ref{fig:ablation} reports the results on UTKFace under simulated Gaussian label noise at varying corruption levels, highlighting the individual contributions of each related component.

% \begin{figure}[h]
%     \centering
%     \includegraphics[width=\linewidth]{fig/ablation.pdf}
%     \caption{Ablation study results.}
    
% \end{figure}

% ======================================================================

\textbf{Hyperparameter Sensitivity Analysis.} We further analyze the sensitivity of GRAIN to different loss coefficients. Since the magnitude of the CE loss varies less than that of the regression loss during training, we fix the CE loss weight to $\lambda_{\mathrm{ce}}=1.0$ in all experiments, and treat the regression loss weight $\lambda_{\mathrm{reg}}$ as a hyperparameter. Figure~\ref{fig:hyper_lam_reg_MAE} reports the performance under different values of $\lambda_{\mathrm{reg}}$ on UTKFace with simulated Gaussian label noise at varying corruption levels. The red dashed line indicates the performance at $\lambda_{\mathrm{reg}}=0$, corresponding to the removal of the regression loss; and the black dashed line denotes the best-performing baseline method. The results show that GRAIN achieves consistently strong performance within a reasonable range of $\lambda_{\mathrm{reg}}$. 
Additional sensitivity analyses on $\lambda_{\mathrm{pos}}$, the ratio $\lambda_{\mathrm{neg}}/\lambda_{\mathrm{pos}}$, the number of contrastive pairs, and the number of bins are provided in Figures~\ref{fig:hyper_lam_pos_MAE}, \ref{fig:hyper_uot_pos_neg_ratio_MAE}, \ref{fig:hyper_uot_pair_MAE}, and Figures \ref{fig:hyper_bin_num_quantile}-\ref{fig:hyper_bin_num_K} in Appendix~\ref{appdx.exps}. These results show that GRAIN remains robust across a wide range of hyperparameter choices under varying noise levels.

% \begin{figure}[h]
%     \centering
%     \includegraphics[width=\linewidth]{fig/hyper_lam_reg_MAE.pdf}
%     \caption{Hyperparameter sensitivity analysis on $\lambda_{\text{reg}}$.}
    
% \end{figure}

% \textbf{Additional Analyses.}
% Additional analyses on the residual head behavior (Figure \ref{fig:scatter_residual}),
% the dynamic evolution of the adaptive parameter $\tau$ (Figure \ref{fig:tau_dynamics}),
% and results under diverse discrepancy choices for $\mathscr{d}$ in (\ref{eq.loss_feature_pos}) and (\ref{eq.loss_feature_neg}) (Figure \ref{fig:uot_ot_cosine})
% are provided in Appendix~\ref{appdx.exps}.

\textbf{Additional Analyses.}
{
We further investigate the residual head behavior,
the dynamics of the adaptive threshold $\tau$,
and the choice of discrepancy measure, with results
provided in Appendix~\ref{appdx.exps}.
First, Figure~\ref{fig:scatter_residual} shows that
the learned residuals exhibit negligible correlation
with the true labels (Pearson $R\approx 0.04$),
whereas the regression predictions retain strong
correlations, supporting the intended separation
between residual correction and predictive signals.
Second, Figure~\ref{fig:tau_dynamics} demonstrates
that $\tau$ gradually decreases during training
while maintaining larger values under higher noise
levels, illustrating its adaptability to both noise
severity and training dynamics.
Finally, Figure~\ref{fig:uot_ot_cosine} compares
UOT with balanced OT and cosine similarity as
the discrepancy measure $\mathscr{d}$ in
(\ref{eq.loss_feature_pos}) and
(\ref{eq.loss_feature_neg}).
}

%%%%%%%%%%%%%%%%%%%%%%%%%%%%%%%%%%%%%%%%%%%%%%%%%%%%%%%%%%%%%%%%%%%%%%%%%%%%%%%%

\section{Conclusion}\label{sec.conclusion}

In this work, we introduce GRAIN, a dual-granularity framework for robust 
regression under noisy labels. At the fine-grained level, instance-specific 
auxiliary variables derived from a robust loss formulation decouple label 
noise from predictive signals. At the coarse-grained level, a binning-based 
contrastive regularization enforces an order-preserving feature geometry 
aligned with label proximity. The two levels are coupled via a 
histogram-based bridging mechanism that enables bidirectional information 
flow between structural and residual learning. Theoretical analysis and 
extensive experiments on synthetic and real-world datasets demonstrate that 
GRAIN consistently outperforms baselines across diverse settings. Future work includes extending the theoretical guarantees to weakly 
dependent settings and more general label noise generation models, and 
developing a formal analysis of the contrastive regularization and its 
interaction with residual learning. A detailed discussion of limitations, future directions, and broader impact is provided in 
Appendix~\ref{appdx.limitation}.

% \section*{Acknowledgements}
% Yi is the Canada Research Chair in Data Science (Tier 1). Her research was supported by the Canada Research Chairs Program and the Natural Sciences and Engineering Research Council of Canada (NSERC).

\subsection*{Acknowledgements}
Yi is a Tier 1 Canada Research Chair in Data Science. Her research is supported by the Canada Research Chairs Program and the Natural Sciences and Engineering Research Council of Canada (NSERC), with partial support from the Canadian Statistical Sciences Institute (CANSSI) through its Collaborative Research Team program. Wang's research is supported by the Natural Sciences and Engineering Research Council of Canada (NSERC), Discovery Grants program.

% \iffalse

\newpage

% \section*{References}

% References follow the acknowledgments in the camera-ready paper. Use unnumbered first-level heading for
% the references. Any choice of citation style is acceptable as long as you are
% consistent. It is permissible to reduce the font size to \verb+small+ (9 point)
% when listing the references.
% Note that the Reference section does not count towards the page limit.
% \medskip

{
\small
\bibliographystyle{unsrt}
% \bibliography{references}

}

%%%%%%%%%%%%%%%%%%%%%%%%%%%%%%%%%%%%%%%%%%%%%%%%%%%%%%%%%%%%
\newpage
\appendix

\numberwithin{equation}{section}
\numberwithin{table}{section}
\numberwithin{figure}{section}

\begin{center}
    \bf\Large Appendix: Theoretical Derivations, Implementation Details,\\
    and Additional Experiments
\end{center}

\section{Technical Details}\label{sec.appdx_Technical}

Using \cite{zhou2018new}, \cite{sun2020adaptive}, and \cite{spokoiny2012parametric}, we establish Theorem \ref{thm.robust_main} under suitable regularity conditions. We first present additional notations and preliminary results, then we identify regularity conditions, and finally, we prove Theorem \ref{thm.robust_main}.

\paragraph{Notations.} Let $\mathbb{R}_{\ge 0}$ denote the set of all nonnegative real values, and let $\mathbbm{1}(\cdot)$ represent the indicator function. For any positive integer $K$, we denote $[K]\triangleq\{1,\ldots,K\}$. 
{For any set $\Omega\subseteq\mathbb{R}^m$, where $m$ is a positive
integer, let $\mathcal{P}(\Omega)$ denote the set of all Borel probability measures on $\Omega$, and let $\mathcal{M}_q^+(\Omega)$ denote the set of finite nonnegative Borel measures on $\Omega$ with finite $q$-th moments for $q\geq 1$. We further define
$\mathcal{P}_q(\Omega)=\mathcal{P}(\Omega)\cap\mathcal{M}_q^+(\Omega)$ as the set of probability measures with finite $q$-th moments.}
For two vectors $a,a'\in\mathbb{R}^m$, we use $\langle a,a'\rangle$ to denote their dot product. For vector $a=(a_1,...,a_p)^{\top}$ and $q\in[1,+\infty]$, the $L_q$ norm of $a$ is defined as $\|a\|_q=(\sum_{j=1}^p|a_j|^q)^{1/q}$ if $1\le q<\infty$, and $\|a\|_{\infty}=\max_{j}|a_j|$ if $q=+\infty$. For any norm $\|\cdot\|$ on $\mathbb{R}^m$, we use $\|\cdot\|_*$ to denote its dual norm, defined as $\|a\|_*=\sup\left\{\langle a,a'\rangle:\|a'\|\le 1\right\}$. 
For a proper, closed, 
convex function $f:\mathbb{R}^m\rightarrow\mathbb{R}\cup\{+\infty\}$, its convex 
conjugate (Fenchel conjugate) is defined as 
$f^*(s)=\sup_{a\in\mathbb{R}^m}\left\{\langle s,a\rangle - f(a)\right\}$.
The Lipschitz modulus of a function 
$\mathdutchcal{h}:\Omega\rightarrow\mathbb{R}$ is denoted by $\text{Lip}(\mathdutchcal{h})=\sup_{a\ne a'}\frac{|\mathdutchcal{h}(a)-\mathdutchcal{h}(a')|}{\|a-a'\|}$. 

Let $\mathbb{S}^{m-1}\triangleq\{u\in\mathbb{R}^m:\|u\|_2=1\}$, where $\|u\|_2$ denotes the $L_2$ norm for the vector $u$. For an $m\times m$ matrix ${V}$, we use $\|V\|_2$ to denote its spectral norm, defined as $\|V\|_2=\sup_{u\in\mathbb{S}^{m-1}}\|Vu\|_2$. For ${V}$ having real eigenvalues, let $\lambda_1({V}),\ldots,\lambda_m({V})$ denote its eigenvalues with $\lambda_1({V})\ge\ldots\ge\lambda_m({V})$, and use $\lambda_{\text{min}}({V})$ and $\lambda_{\text{max}}({V})$ to stress the smallest and largest eigenvalues of ${V}$, respectively. We write ${V}\succ 0$ (or ${V}\succeq 0$) if ${V}$ is a positive (or semipositive) definite matrix.

\subsection{Preliminaries}

\begin{definition}[OT, Kantorovitch formulation, \citealp{santambrogio2015optimal}]\label{def.OT}
Let $\Omega$ denote a compact metric space equipped with a non-negative cost function $\mathscr{c}:\Omega\times\Omega\rightarrow\mathbb{R}_{\ge 0}$. For any ${P_1, P_2}\in \mathcal{P}(\Omega)$, the \emph{optimal transport cost} between $P_1$ and $P_2$ is defined as
\begingroup\small
\begin{align}\label{eq.def_OT}
    \text{OT}({P_1, P_2})\triangleq\inf_{\Pi\in\text{Cpl}({P_1, P_2})}\int_{\Omega\times\Omega}\mathscr{c}(u_1,u_2)d\Pi(u_1,u_2),
\end{align}
\endgroup
where $\text{Cpl}(P_1, P_2)$, commonly referred to as \emph{transport plans}, denotes the set of couplings between $P_1$ and $P_2$, i.e., probability measures on $\Omega \times \Omega$ with marginals $P_1$ and $P_2$. Any minimizer {$\Pi^*\in \text{Cpl}(P_1, P_2)$} of {(\ref{eq.def_OT})} is called an \emph{optimal transport plan} between $P_1$ and $P_2$.
\end{definition}

\begin{definition}[UOT, {\citealp{sejourne2023unbalanced}}]\label{def.UOT}
Consider the setup in Definition \ref{def.OT}. 
For any ${P_1, P_2}\in \mathcal{P}(\Omega)$ and $\gamma,\gamma'>0$, the \emph{entropy-regularized unbalanced optimal transport cost} between $P_1$ and $P_2$ is defined as
\begin{align*}
    \text{UOT}^{\gamma'}_{\gamma}({P_1, P_2})\triangleq&\inf_{\Pi\in{\mathcal{M}^+_2(\Omega\times\Omega)}}\int_{\Omega\times\Omega}\mathscr{c}(u_1,u_2)d\Pi(u_1,u_2)\\
    &\hspace{1cm}+\gamma \text{KL}(\Pi_1\|P_1)+\gamma \text{KL}(\Pi_2\|P_2)\\
    &\hspace{1cm}+\gamma'\text{KL}(\Pi\|P_1\otimes P_2),
\end{align*}
where {$\Pi_1$ and $\Pi_2$ denote the first and second marginal distributions of  $\Pi$, respectively}, $\text{KL}(\cdot\|\cdot)$ is the Kullback–Leibler (KL) divergence, and $P_1\otimes P_2$ is the product measure of $P_1$ and $P_2$. When $\gamma'=0$, we write the UOT cost as $\text{UOT}_{\gamma}(P_1, P_2)$.
\end{definition}

\begin{remark}
Unlike balanced OT in Definition~\ref{def.OT}, which enforces exact marginal matching and is therefore sensitive to outliers, UOT relaxes these hard constraints into a soft alignment, penalizing large deviations while permitting controlled mass mismatch. This flexibility allows the transport plan to deviate from the original marginals, effectively downweighting outliers and improving robustness.
\end{remark}

\begin{lemma}[\cite{zhou2018new}]\label{lemma.zhou2018}
For any positive integer $d\ge 1$ and convex loss function $\ell:\mathbb{R}^d\rightarrow\mathbb{R}$, let $\mathtt{D}_{\ell}(\beta_1,\beta_2)\triangleq \ell(\beta_1)-\ell(\beta_2)-\langle \nabla \ell(\beta_2),\beta_1-\beta_2\rangle$ and $\Bar{\mathtt{D}}_{\ell}(\beta_1,\beta_2)\triangleq \mathtt{D}_{\ell}(\beta_1,\beta_2)+\mathtt{D}_{\ell}(\beta_2,\beta_1)$ for $\beta_1,\beta_2\in\mathbb{R}^d$. Then, for any $\beta_{\eta}\triangleq \beta_1+\eta(\beta_2-\beta_1)$ with $\eta\in[0,1]$, $\Bar{\mathtt{D}}_{\ell}(\beta_1,\beta_\eta)\le \eta \Bar{\mathtt{D}}_{\ell}(\beta_1,\beta_2)$.
\end{lemma}

\begin{lemma}\label{lemma.matrix}
Let ${A,B}\in\mathbb{R}^{m_1\times m_2}$, with $m_1$ and $m_2$ denoting two positive integers. Then the following results hold.
\begin{enumerate}[label=(\roman*)]
    
    \item \label{lemma.matrix_2} $\|A\|_2=\sup_{u\in\mathbb{S}^{m_2-1}}\|A {u}\|_2=\sup_{u\in\mathbb{S}^{m_1-1}, v\in\mathbb{S}^{m_2-1}}u^\top Av$; $\|A\|_2^2=\|A^\top A\|_2=\lambda_{\max}(A^\top A)$. {Specifically, for a vector $a\in\mathbb{R}^{m_1}$, we have $\|a\|_2=\sup_{u\in\mathbb{S}^{m_1-1}}u^\top a$.}
    % If $m_1=m_2=m$ and $A$ is symmetric, we further have 

    \item \label{lemma.matrix_1} If $m_1=m_2=m$ and $A$ is symmetric, then $\sup_{u\in\mathbb{S}^{m-1}}{u}^\top{A}{u}=\lambda_{\max}({A})$ and $\inf_{u\in\mathbb{S}^{m-1}}{u}^\top{A}{u}=\lambda_{\min}({A})$. {Additionally,} $\|A\|_2=\max_{1\le i\le m}|\lambda_{i}(A)|=\max\{|\lambda_{\max}(A)|,|\lambda_{\min}(A)|\}$. 
    
    \item \label{lemma.matrix_3} (Weyl's inequality) If $m_1=m_2=m$ and $A$ and $B$ are symmetric, then 
    \begin{itemize}
        \item $\lambda_{i+j-1}(A+B)\le\lambda_i(A)+\lambda_j(B)$ with $1\le i,j,i+j-1\le m$;
        \item $\lambda_i(A)+\lambda_j(B)\le\lambda_{i+j-m}(A+B)$ with $1\le i,j,i+j-m\le m$. 
    \end{itemize}
    In particular, $\lambda_{\min}(A+B)\ge\lambda_{\min}(A)+\lambda_{\min}(B)$, and $\lambda_{\max}(A+B)\le\lambda_{\max}(A)+\lambda_{\max}(B)$. 
\end{enumerate}

\end{lemma}

\begin{lemma}[Modified from Corollaries 1.12 and 1.13 in \cite{spokoiny2012parametric}]\label{lemma.upper_bound}
Let $\zeta$ denote a random vector in $\mathbb{R}^m$. Assume that there exist matrix $V_0\in\mathbb{R}^{m\times m}$ and constant $\mathtt{g}>0$ such that
\begin{align}\label{lemma.upper_bound_cond}
    \sup_{u\in\mathbb{R}^{m}}\log\mathbb{E}\left\{\exp\left(t\frac{u^\top\zeta}{\|V_0u\|_2}\right)\right\}\le t^2/2,\ \ |t|\le \mathtt{g}.
\end{align}
Let $D_0\in\mathbb{R}^{m\times m}$ denote a given positive definite matrix, and define $\mathbbm{B}\triangleq D_0^{-1/2}V_0 D_0^{-1/2}$. Let $\mathtt{m}\triangleq \text{tr}(\mathbbm{B}^2)$ and $\mathtt{v}\triangleq\sqrt{2\text{tr}(\mathbbm{B}^4)}$, and assume $\mathtt{g}\ge\sqrt{2\mathtt{m}}$. Denote $\mathtt{c}_1\triangleq \mathtt{g}^2/(\mathtt{m}+\mathtt{g}^2)$, $\mathtt{c}_2\triangleq \sqrt{\mathtt{g}^2/\mathtt{c}_1^2-\mathtt{m}/\mathtt{c}_1}=\sqrt{\mathtt{m}+\mathtt{g}^2}$, $\mathtt{c}_3\triangleq \mathtt{c}_1 \mathtt{c}_2=\sqrt{\mathtt{g}^2-\mathtt{c}_1 \mathtt{m}}=\mathtt{g}^2/\sqrt{\mathtt{m}+\mathtt{g}^2}$, and $\mathtt{c}_4\triangleq \frac{1}{2}\left\{\mathtt{c}_3\mathtt{c}_2+\log\det(I_m-\mathtt{c}_1\mathbbm{B}^2)\right\}$. Then 
\begin{itemize}
    \item[(a)] for any $\mathtt{x}\le\mathtt{c}_4$,
    \begin{align*}
        \mathbb{P}\left\{\|D_0^{-1}\zeta\|^2_2\ge \mathscr{h}_1(\mathtt{x},\mathbbm{B})\right\}\le 2e^{-\mathtt{x}}+8.4e^{-\mathtt{c}_4},
    \end{align*}
    where 
    \begin{equation*}
        \mathscr{h}_1(\mathtt{x},\mathbbm{B})\triangleq\left\{
        \begin{matrix}
        \mathtt{m}+\sqrt{2\mathtt{x}\mathtt{v}},\ \ & \mathtt{x}\le \mathtt{v}/18,\ \ \ \ \ \ \ \ \ \\
        \mathtt{m}+6\mathtt{x},\ \ \ \ \ \ \  & \mathtt{v}/18<\mathtt{x}\le\mathtt{c}_4.
        \end{matrix}
        \right.
    \end{equation*}

    \item[(b)] for any $\mathtt{x}>\mathtt{c}_4$, 
    \begin{align*}
        \mathbb{P}\left\{\|D_0^{-1}\zeta\|^2_2\ge \mathscr{h}_2(\mathtt{x},\mathbbm{B})\right\}\le 8.4e^{-\mathtt{x}},
    \end{align*}
    where $\mathscr{h}_2(\mathtt{x},\mathbbm{B})\triangleq |\mathtt{c}_2+2(\mathtt{x}-\mathtt{c}_4)/\mathtt{c}_3|^2$. 

\end{itemize}

\end{lemma}

\begin{lemma}[Sub-gaussian properties; \cite{vershynin2018high}, Proposition 2.5.2]\label{lemma.sub_gaussian_property}
For $j=1,\ldots,5$, let $K_j$ denote positive constants that may differ from each other 
by at most an absolute constant factor. Let $\zeta$ be a random variable. Then the following statements are equivalent: 
\begin{enumerate}[label=(\roman*)]
    \item\label{lemma.sub_gaussian_property_1} Tail probability:
        $\mathbb{P}(|\zeta| \ge t) \le 2 \exp(-t^{2}/K_{1}^{2})
        \ \ \text{for all } t \ge 0.$

    \item\label{lemma.sub_gaussian_property_2} Moment bound: 
        $\|\zeta\|_{L_{p}} {\triangleq} \{\mathbb{E}(|\zeta|^{p})\}^{1/p} 
        \le K_{2} \sqrt{p}
        \ \ \text{for all } p \ge 1.$

    \item\label{lemma.sub_gaussian_property_3} Bound for the moment generating function:
        $\mathbb{E}\{\exp(t^{2} \zeta^{2})\} 
        \le \exp(K_{3}^{2} t^{2})
        \ \ \text{for all } t \text{ with } |t| \le \frac{1}{K_{3}}.$

    \item\label{lemma.sub_gaussian_property_4} Bound for a special moment generating function:
        $\mathbb{E}\{\exp(\zeta^{2}/K_{4}^{2})\} \le 2.$

\end{enumerate}

Moreover, if $\mathbb{E}\zeta = 0$ then properties (i)–(iv) are also equivalent to the following one:

\begin{enumerate}[label=(\roman*)]
    \setcounter{enumi}{4}
    \item\label{lemma.sub_gaussian_property_5} Bound for the moment generating function:
        $\mathbb{E}\{\exp(t \zeta)\}
        \le \exp(K_{5}^{2} t^{2})
        \ \ \text{for all } t \in \mathbb{R}.$
\end{enumerate}
\end{lemma}

\begin{definition}[Sub-Gaussian random variables and vectors] \ 
\begin{enumerate}[label=(\roman*)]
    \item\label{def.sub_gaussian_variable} [\cite{vershynin2018high}, Definition 2.5.6] 
    A random variable $\zeta$ satisfying one of the statements from \ref{lemma.sub_gaussian_property_1} to \ref{lemma.sub_gaussian_property_4} in Lemma \ref{lemma.sub_gaussian_property} is called a 
    \emph{sub-gaussian random variable}. 
    The \emph{sub-gaussian norm} of $\zeta$, denoted $\|\zeta\|_{\psi_{2}}$, 
    is defined to be the smallest possible $K_{4}$ in statement \ref{lemma.sub_gaussian_property_4}, i.e., 
    \[
        \|\zeta\|_{\psi_{2}}
        = \inf\left\{ t > 0 : \mathbb{E}\exp(\zeta^{2}/t^{2}) \le 2 \right\}.
    \]
    {Here, the notation $\psi_2$ comes from the Orlicz function $\psi_{\alpha}(x) = \exp(x^\alpha) - 1$ with $\alpha=2$, which is commonly used to characterize sub-Gaussian tail behavior.}

    \item\label{def.sub_gaussian_vector} [\cite{vershynin2018high}, Definition 3.4.1]
    A random vector $\zeta$ in $\mathbb{R}^{m}$ is called \emph{sub-Gaussian}
    if for any $u \in \mathbb{R}^{m}$, $\langle \zeta, u\rangle$ is a sub-Gaussian
    random variable. The \emph{sub-Gaussian norm} of $\zeta$ is defined as
    \[
        \|\zeta\|_{\psi_{2}}
        = \sup_{u \in \mathbb{S}^{m-1}} 
          \|\langle \zeta, u\rangle\|_{\psi_{2}} .
    \]
\end{enumerate}
\end{definition}

\begin{lemma}[\cite{vershynin2018high}, (4.22); \cite{vershynin2010introduction}, Remark 5.40]\label{lemma.matrix_subgaussian_row}
Assume that $A$ is an $m\times d$ matrix whose rows $A_j$ are independent sub-gaussian random vectors in $\mathbb{R}^d$, with second moment matrix $\Sigma_{A}=\mathbb{E}(A_jA_j^\top)$ for $j\in[m]$. Then for every $w\ge 0$, with probability at least $1-2e^{-C_1w^2}$,
\begin{align*}
    \left\|\frac{1}{m}A^\top A-\Sigma_{A}\right\|_2\le \max(\epsilon, \epsilon^2),\ \text{with}\ \epsilon=C_2\sqrt{\frac{d}{m}}+\frac{w}{\sqrt{m}},
\end{align*}
where $C_1$ and $C_2$ are positive constants depending only on the subgaussian norms of the rows. 
\end{lemma}

\begin{lemma}[Conjugacy of infimal convolution]
\label{lem:conj_inf_conv}
Let $f$ and $g$ be proper, lower semicontinuous, convex functions on $\mathbb{R}$. 
Their infimal convolution is defined by
\[
  (f \square g)(x) = \inf_{x' \in \mathbb{R}} \{ f(x') + g(x - x') \}.
\]
Then the following hold:
\begin{enumerate}
  \item[(i)] \textbf{(Dual identity)} For any proper convex $f$ and $g$ such that 
  $f \square g$ is proper, 
  \[
    (f \square g)^* = f^* + g^*.
  \]
  \item[(ii)] \textbf{(Primal recovery)} If in addition $f \square g$ is lower 
  semicontinuous, for instance if 
  $\operatorname{int}(\operatorname{dom} f) \cap \operatorname{dom} g \neq \emptyset$, 
  then by the biconjugate theorem,
  \[
    f \square g = (f^* + g^*)^*.
  \]
  \item[(iii)] \textbf{(Converse)} If $h$ is proper, lower semicontinuous, convex and 
  $h^* = f^* + g^*$, then $h = f \square g$.
\end{enumerate}
\end{lemma}

\begin{lemma}[Smoothness--strong convexity duality]
\label{lem:smooth_strong_conv}
Let $f:\mathbb{R}\rightarrow\mathbb{R}\cup\{+\infty\}$ be a proper, lower 
semicontinuous, convex function. Then $f$ has $L$-Lipschitz gradient if and only 
if $f^*$ is $\tfrac{1}{L}$-strongly convex, i.e.,
\[
  f^*(tw+(1-t)w') \le t f^*(w)+(1-t)f^*(w') 
  - \frac{1}{2L}t(1-t)(w-w')^2,
  \quad \forall\, w,w'\in\mathbb{R},\ t\in[0,1].
\]
Equivalently, the function $w \mapsto f^*(w) - \tfrac{1}{2L}w^2$ is convex.
\end{lemma}

\begin{lemma}[Quadratic infimal convolution criterion]
\label{lem:quad_inf_conv}
Let $q(x)=x^2/2$, and let $\rho$ be a proper, lower semicontinuous, convex, and 
differentiable function whose derivative is $1$-Lipschitz. Then
\[
  \psi(w):=\rho^*(w)-q^*(w)=\rho^*(w)-\frac{1}{2}w^2
\]
is convex. Consequently, defining $\phi:=\psi^*$, we have
\[
  \rho(\nu)=\inf_{\xi\in\mathbb{R}}
  \left\{
  \frac{1}{2}\xi^2+\phi(\nu-\xi)
  \right\}.
\]
Moreover, $\phi$ is uniquely determined in the class of proper lower semicontinuous convex functions.
\end{lemma}

\begin{proof}
Since $\rho$ is proper, lower semicontinuous, convex, and has $1$-Lipschitz 
derivative, Lemma~\ref{lem:smooth_strong_conv} (with $L=1$) implies that $\rho^*$ 
is $1$-strongly convex, i.e., the function $w \mapsto \rho^*(w) - \tfrac{1}{2}w^2$ 
is convex. Since $q^*(w)=\tfrac{1}{2}w^2$, this shows that $\psi=\rho^*-q^*$ is convex.

As $\rho^*$ is proper and lower semicontinuous, and $q^*$ is continuous and finite-valued, 
it follows that $\psi$ is also proper and lower semicontinuous. Define $\phi:=\psi^*$. 
Then $\phi$ is proper, lower semicontinuous, and convex. By the Fenchel--Moreau theorem,
\[
  \phi^* = \psi^{**} = \psi = \rho^* - q^*.
\]
Hence,
\[
  q^* + \phi^* = \rho^*.
\]
Applying Lemma~\ref{lem:conj_inf_conv}(iii) yields
\[
  \rho = q \square \phi,
\]
which gives the desired representation.

Finally, since $\phi^* = \rho^* - q^*$ is uniquely determined by $\rho$, the function $\phi$ 
is uniquely determined (as a proper lower semicontinuous convex function) via the conjugacy relation.
\end{proof}

\begin{lemma}\label{lemma.convex}
Under Assumption~\ref{assumption.ell}, $\ell_{\tau}(\cdot)$ defined in 
(\ref{eq.piecewise_loss}) is convex.
\end{lemma}
\begin{proof}
Since $\ell_\tau(\cdot)$ is an even function, its derivative $\ell_\tau'(\cdot)$ 
is odd. It therefore suffices to show that $\ell_\tau'$ is non-decreasing on 
$[0,\infty)$ and that $\ell_\tau'(0)=0$, as these together imply $\ell_\tau'$ is 
non-decreasing on all of $\mathbb{R}$.

For $\nu \ge 0$,
\[
\ell_\tau'(\nu)
=
\begin{cases}
\nu, & 0\le \nu\le \tau,\\
\dfrac{\tau}{\ell'(\tau)} \ell'(\nu), & \nu>\tau.
\end{cases}
\]
We have $\ell_\tau'(0)=0$, and the derivative is continuous at $\nu=\tau$ since
\[
\frac{\tau}{\ell'(\tau)}\ell'(\tau)=\tau.
\]
On $[0,\tau)$, $\ell_\tau'(\nu)=\nu$ is non-decreasing. On $(\tau,\infty)$, 
since $\ell$ is convex by Assumption~\ref{assumption.ell}, $\ell'$ is 
non-decreasing, and hence $\ell_\tau'(\nu) = \tfrac{\tau}{\ell'(\tau)}\ell'(\nu)$ 
is also non-decreasing. Continuity at $\tau$ then ensures $\ell_\tau'$ is 
non-decreasing on all of $[0,\infty)$, and therefore on $\mathbb{R}$. 
Hence $\ell_\tau(\cdot)$ is convex.
\end{proof}

\subsection{Regularity Conditions}\label{appdx.regularity_condition}

\begin{enumerate}[label=(\roman*)]
    \item \label{regu.1} The zero-mean error variables $\varepsilon'$ and $\varepsilon''$ in models (\ref{eq.true_label}) and (\ref{eq.noisy_label}) are independent and also independent of $\mathrm{X}$. Without loss of generality, we assume $\mathbb{E}(\varepsilon'^{2})=\mathbb{E}(\varepsilon''^{2})=1$. 
    
    \item \label{regu.2} Let $\Bar{\varsigma}^2(\mathrm{X})\triangleq 1+\mathbb{E}_{\varepsilon'}\left\{\varsigma^2(\mathrm{X},\langle \varphi(\mathrm{X}),\theta^*_{\text{reg}}\rangle+\varepsilon'){|\mathrm{X}}\right\}$. Assume that $\alpha_1\triangleq\mathbb{E}\{\Bar{\varsigma}^{2+\iota_1}(\mathrm{X})\}<\infty$ for some $\iota_1>0$ and denote $\sigma^2\triangleq\mathbb{E}\{\Bar{\varsigma}^2(\mathrm{X})\}$.
    
    \item \label{regu.3} For $\mathrm{Z}=\varphi(\mathrm{X})$ defined in Section \ref{sec.method}, let $\Sigma\triangleq \mathbb{E}(\mathrm{Z}\mathrm{Z}^\top)$ and $\nu^*\triangleq\varepsilon'+\varsigma(\mathrm{X},\mathrm{Y})\varepsilon''$. Assume that $\Sigma$ is strictly positive definite, and let $\zeta\triangleq\Sigma^{-1/2}\mathrm{Z}$. We assume the following conditions.

    \begin{enumerate}[label=(\alph*)]
        \item \label{regu.3.a} $\zeta$ is a sub-Gaussian random vector. Denote by $K_j^{(\zeta)}$, $1 \le j \le 4$, the constants associated with $\zeta$ in Lemma~\ref{lemma.sub_gaussian_property}~\ref{lemma.sub_gaussian_property_1}--\ref{lemma.sub_gaussian_property_4}, and let $\alpha_2 \triangleq \min_{1 \le j \le 4} K_j^{(\zeta)}$.

        \item \label{regu.3.b} There exists a constant $\kappa \ge 0$ such that $|\nu^*|^\kappa \zeta$ is a sub-Gaussian random vector. Denote by $K_j^{(\kappa)}$, $1 \le j \le 4$, the constants associated with $|\nu^*|^\kappa \zeta$ in Lemma~\ref{lemma.sub_gaussian_property}~\ref{lemma.sub_gaussian_property_1}--\ref{lemma.sub_gaussian_property_4}, and let $\alpha_3 \triangleq \min_{1 \le j \le 4} K_j^{(\kappa)}$.

    \end{enumerate}
    
\end{enumerate}

\subsection{Distributionally Robust Extension}
\label{appdx.dro_extension}

{
In this subsection, we examine a distributionally robust
optimization (DRO) perspective on learning with noisy labels.}
Instead of minimizing the empirical risk
$\Re^\tau_{\text{reg}}(\theta_{\text{reg}})$ under the empirical
distribution $\widehat{P}^\dagger$ defined in
Section~\ref{sec.residual_head}, we consider the worst-case risk
over a neighborhood of distributions:
\begin{align}\label{eq.robust_loss}
    \Re^{\tau,\rho}_{\text{reg}}(\theta_{\text{reg}})
    =\sup_{Q\in\Gamma_{\rho}(\widehat{P}^\dagger)}
    \mathbb{E}_{(\mathrm{Z},\widetilde{\mathrm{Y}})\sim Q}
    \left\{
    \ell_\tau(\langle \mathrm{Z},\theta_{\text{reg}}\rangle
    -\widetilde{\mathrm{Y}})
    \right\},
\end{align}
where $\ell_\tau(\cdot)$ is defined in
\eqref{eq.piecewise_loss}, and
\begin{align}\label{eq.uncertainty_set}
    \Gamma_{\rho}(\widehat{P}^\dagger)
    =\left\{
    Q\in\mathcal{P}(\mathcal{Z}\times\mathcal{Y}):
    \mathscr{d}(Q,\widehat{P}^\dagger)\le\rho
    \right\},
\end{align}
denotes an \emph{ambiguity set} centered at
$\widehat{P}^\dagger$, where $\mathscr{d}$ is a discrepancy
metric between probability measures and $\rho>0$ controls
the size of the uncertainty set.
This formulation captures distributional perturbations
arising from label noise.

We construct $\Gamma_{\rho}(\widehat{P}^\dagger)$ using
an optimal transport (OT) discrepancy, which is given
in Definition \ref{def.OT}. Specifically, let the cost
function $\mathscr{c}:\Omega\times\Omega\to\mathbb{R}_+$
be defined as $\mathscr{c}(u_1,u_2)=\|u_1-u_2\|$
for a norm $\|\cdot\|$ on
$\Omega=\mathcal{Z}\times\mathcal{Y}$.
Under this choice, the DRO problem in (\ref{eq.robust_loss}) admits the following dual reformulation.

\begin{lemma}[\citealp{shafieezadeh2019regularization}]
\label{lemma.dual}
Assume that
$\ell_\tau(\cdot)$ is convex and $L$-Lipschitz continuous.
Then, $\inf_{\theta_{\text{reg}}}
\Re^{\tau,\rho}_{\text{reg}}(\theta_{\text{reg}})$
is equivalent to the following optimization problem:
\begingroup\small
\begin{align*}
    \inf_{\theta_{\mathrm{reg}}}
    \Bigg[
    \mathbb{E}_{(\mathrm{Z},\widetilde{\mathrm{Y}})
    \sim\widehat{P}^\dagger}
    \left\{
    \ell_\tau\left(
    \langle\mathrm{Z},\theta_{\text{reg}}\rangle
    -{\widetilde{\mathrm{Y}}}
    \right)
    \right\}+\rho\text{Lip}(\ell_\tau)
    \|(\theta_{\text{reg}}^\top,-1)^\top\|_*
    \Bigg],
\end{align*}
\endgroup
where $\text{Lip}(\ell_\tau)$ denotes the Lipschitz
modulus of the loss function $\ell_\tau$, and
$\|\cdot\|_*$ represents the dual norm of $\|\cdot\|$.
\end{lemma}

{
Lemma \ref{lemma.dual} establishes an exact regularized
reformulation of the DRO problem \eqref{eq.robust_loss}
under its stated conditions.
In particular, when $\ell(t)=t$, the piecewise loss
$\ell_\tau$ reduces to the Huber loss with threshold
$\tau$, which is globally Lipschitz continuous with
$\text{Lip}(\ell_\tau)=\tau$.
For more general choices of $\ell$, however,
$\ell_\tau$ need not be globally Lipschitz, and
the equivalence in Lemma \ref{lemma.dual}
does not necessarily apply.
}
For example, if the norm $\|\cdot\|$ in the cost
function $\mathscr{c}$ is set to the $L_q$ norm for
any $1<q<\infty$, then the corresponding dual norm
$\|\cdot\|_*$ is the $L_{q'}$ norm with
$1/q+1/q'=1$. In particular, when $q=2$, both
norms $\|\cdot\|$ and $\|\cdot\|_*$ are identical
to the $L_2$ norm.

Motivated by {the connection between DRO
and regularization}, we consider the following
{ridge-regularized} empirical objective:
\begin{align}\label{eq.emp_risk_dro}
    \Re^{\tau,\lambda}_{\text{reg}}(\theta_{\text{reg}})
    \triangleq
    \frac{1}{N}\sum_{i=1}^{N}
    \ell_{\tau}(
    \langle \mathrm{Z}_i,\theta_{\text{reg}}\rangle
    -\widetilde{\mathrm{Y}}_i)
    +\frac{\lambda}{2}\|\theta_{\text{reg}}\|^2_2,
\end{align}
where $\lambda{\geq}0$ is a
regularization parameter.{
Unlike the exact dual-norm regularization in
Lemma \ref{lemma.dual}, the squared $L_2$ penalty
in \eqref{eq.emp_risk_dro} is adopted as a
DRO-motivated regularization choice.}
Let $\widehat{\theta}_{\text{reg}}^{\tau,\lambda}$
denote a minimizer of \eqref{eq.emp_risk_dro}.{
The following result establishes statistical guarantees
for this ridge-regularized estimator under the general
conditions of Theorem~\ref{thm.robust_main}.
When $\lambda=0$, it recovers
Theorem~\ref{thm.robust_main}.
}

\begin{theorem}\label{thm.robust_main_dro}
Assume that the conditions in
Theorem \ref{thm.robust_main} hold and that
$\lambda \le \mathscr{C}_3
\sqrt{\frac{d+s}{N}}$
for some constant $\mathscr{C}_3>0$. Then,
\begin{enumerate}[label=(\roman*)]
    \item \label{thm.robust_main_dro_1}
    with probability at least $1-5e^{-s}$,
    $\|\widehat{\theta}_{\text{reg}}^{\tau,\lambda}
    -\theta_{\text{reg}}^*\|_{\Sigma}
    \le \mathscr{C}^\dagger
    \sqrt{\frac{d+s}{N}}$
    for some constant $\mathscr{C}^\dagger>0$;

    \item \label{thm.robust_main_dro_2}
    with probability at least $1-7e^{-s}$,
    for any future input $\mathrm{X}_{\text{new}}$
    with feature
    $\mathrm{Z}_{\text{new}}\triangleq
    \varphi(\mathrm{X}_{\text{new}})$,
    the associated prediction error satisfies:
    $\text{err}\triangleq
    \left|
    \langle\mathrm{Z}_{\text{new}},
    \widehat{\theta}_{\text{reg}}^{\tau,\lambda}\rangle
    -\langle\mathrm{Z}_{\text{new}},
    \theta_{\text{reg}}^*\rangle
    \right|
    \le \mathscr{C}^{\ddagger}
    \sqrt{\frac{ds+s^2}{N}}$
    for some constant $\mathscr{C}^{\ddagger}>0$.
\end{enumerate}
\end{theorem}

\subsection{Proof of Theorem \ref{thm.robust_main_dro} \ref{thm.robust_main_dro_1} (Theorem \ref{thm.robust_main} \ref{thm.robust_main_1})}

Since the DRO-induced empirical risk in (\ref{eq.emp_risk_dro}) reduces to the empirical risk in (\ref{eq.emp_risk}) when $\lambda=0$, the corresponding minimizer $\widehat{\theta}^{\tau,\lambda}{\text{reg}}$ reduces to $\widehat{\theta}^{\tau}{\text{reg}}$. Hence, it suffices to prove Theorem \ref{thm.robust_main_dro} \ref{thm.robust_main_dro_1} only.

Recall that in Section~\ref{sec.residual_head}, with a fixed feature extractor $\varphi(\cdot)$, the data are assumed to follow the generative mechanism specified by (\ref{eq.true_label}) and (\ref{eq.noisy_label}), together with the setup at the beginning of Section~\ref{sec.residual_head}:
\begin{align}
    \mathrm{Y}&=\langle\mathrm{Z},\theta_{\text{reg}}^*\rangle+\varepsilon'\label{eq.true_label_appx}\\
    \widetilde{\mathrm{Y}} &= \mathrm{Y} + \varsigma(\mathrm{X},\mathrm{Y})\varepsilon''\label{eq.noisy_label_appx}
\end{align}
We consider the piecewise loss $\ell_{\tau}$ in (\ref{eq.piecewise_loss}) and rewrite the DRO-induced empirical risk in (\ref{eq.emp_risk_dro}) as:
\begin{align}\label{eq.loss_proof}
    \mathcal{L}(\theta_{\text{reg}})
    \triangleq& \frac{1}{N}\sum_{i=1}^{N}\ell_{\tau}(|\nu_i|)+\frac{\lambda}{2}\|\theta_{\text{reg}}\|^2_2\notag\\
    =& \frac{1}{N}\sum_{i=1}^{N}\left[\frac{1}{2}\nu^2_i\mathbbm{1}(|\nu_i|\le \tau)+\frac{\tau}{\ell'(\tau)}\left\{\ell(|\nu_i|)-\ell(\tau)\right\}\mathbbm{1}(|\nu_i|> \tau)\right]+\frac{\lambda}{2}\|\theta_{\text{reg}}\|^2_2,
\end{align}
where $\nu_i\triangleq \widetilde{Y}_i-\langle \mathrm{Z}_i,\theta_{\text{reg}}\rangle$. 
Additionally,
\begin{align}
    \nabla \mathcal{L}(\theta_{\text{reg}})=&-\frac{1}{N}\sum_{i=1}^{N}\text{sgn}(\nu_i)\ell'_\tau(|\nu_i|)\mathrm{Z}_i+\lambda \theta_{\text{reg}}\label{eq.loss_deriv_1};\\
    \nabla^2 \mathcal{L}(\theta_{\text{reg}})=&\frac{1}{N}\sum_{i=1}^{N}\ell''_{\tau}(|\nu_i|)\mathrm{Z}_i\mathrm{Z}^\top+\lambda I_{d\times d},\label{eq.loss_deriv_2}
\end{align}
where $\text{sgn}(\cdot)$ denotes the sign function, defined as $\text{sgn}(x)=1$ if $x>0$, $\text{sgn}(x)=-1$ if $x<0$, and $\text{sgn}(0)=0$; $I_{d\times d}$ representing the $d\times d$ identity matrix; and
\begin{align}
    \ell'_\tau(|\nu_i|)
    &\triangleq |\nu_i|\mathbbm{1}(|\nu_i|\le \tau)+\frac{\tau\ell'(|\nu_i|)}{\ell'(\tau)}\mathbbm{1}(|\nu_i|> \tau)\notag\\
    &\le |\nu_i|\mathbbm{1}(|\nu_i|\le \tau)+\min(|\nu_i|,\frac{\tau|\nu_i|^\kappa}{\ell'(\tau)}); \label{eq.ell_deriv_1}\\
    \ell''_\tau(|\nu_i|)
    &\triangleq \mathbbm{1}(|\nu_i|\le \tau)+\frac{\tau\ell''(|\nu_i|)}{\ell'(\tau)}\mathbbm{1}(|\nu_i|>\tau),\label{eq.ell_deriv_2}
\end{align}
where the inequality in (\ref{eq.ell_deriv_1}) follows from  $\frac{\tau\ell'(|\nu_i|)}{\ell'(\tau)}\le |\nu_i|$ for $|\nu_i|>\tau$, which is implied by the assumption in Theorem \ref{thm.robust_main_dro} that {$\frac{\ell'(t)}{t}$ is nonincreasing for $t\ge \tau$}, together with the assumption {$\ell'(t)\le t^\kappa$ for $t>\tau$} in Theorem \ref{thm.robust_main_dro}.

For $\eta\in(0,1)$, let $\widehat{\theta}^{\tau,\lambda}_{\text{reg}}(\eta)\triangleq \theta^*_{\text{reg}}+\eta(\widehat{\theta}^{\tau,\lambda}_{\text{reg}}-\theta^*_{\text{reg}})$ denote an ``intermediate'' estimator of $\theta^*_{\text{reg}}$. Then, 
\begin{align}\label{eq.tmp_1}
    \big\langle \nabla\mathcal{L}(\widehat{\theta}^{\tau,\lambda}_{\text{reg}}(\eta))-\nabla\mathcal{L}(\theta^*_{\text{reg}}), \widehat{\theta}^{\tau,\lambda}_{\text{reg}}(\eta)-\theta^*_{\text{reg}}\big\rangle \le&\ \eta \big\langle \nabla\mathcal{L}(\widehat{\theta}^{\tau,\lambda}_{\text{reg}})-\nabla\mathcal{L}(\theta^*_{\text{reg}}), \widehat{\theta}^{\tau,\lambda}_{\text{reg}}-\theta^*_{\text{reg}}\big\rangle\notag\\
    =&\ \eta \big\langle -\nabla\mathcal{L}(\theta^*_{\text{reg}}), \widehat{\theta}^{\tau,\lambda}_{\text{reg}}-\theta^*_{\text{reg}}\big\rangle\notag\\
    =&-\big\langle\nabla\mathcal{L}(\theta^*_{\text{reg}}), \widehat{\theta}^{\tau,\lambda}_{\text{reg}}(\eta)-\theta^*_{\text{reg}}\big\rangle,
\end{align}
where the first step is due to Lemma \ref{lemma.zhou2018} and Lemma \ref{lemma.convex}, the second step holds since $\widehat{\theta}_{\text{reg}}^{\tau,\lambda}$ is a minimizer of $\mathcal{L}(\theta_{\text{reg}})$, and the last step comes from the definition of $\widehat{\theta}^{\tau,\lambda}_{\text{reg}}(\eta)$. 

Next, we examine the left-hand side of (\ref{eq.tmp_1}). By the mean value theorem, there exists a convex combination of $\widehat{\theta}^{\tau,\lambda}_{\text{reg}}(\eta)$ and $\theta^*_{\text{reg}}$, denoted by $\widetilde{\theta}^{\tau,\lambda}_{\text{reg}}(\eta)$, such that $\nabla\mathcal{L}(\widehat{\theta}^{\tau,\lambda}_{\text{reg}}(\eta))-\nabla\mathcal{L}(\theta^*_{\text{reg}})=\nabla^2\mathcal{L}(\widetilde{\theta}^{\tau,\lambda}_{\text{reg}}(\eta))\left\{\widehat{\theta}^{\tau,\lambda}_{\text{reg}}(\eta)-\theta^*_{\text{reg}}\right\}$. Therefore, using {$\Sigma\triangleq \mathbb{E}(\mathrm{Z}\mathrm{Z}^\top)$} and Lemma \ref{lemma.matrix} \ref{lemma.matrix_1}, we obtain
\begin{align}\label{eq.tmp_2}
    &\big\langle \nabla\mathcal{L}(\widehat{\theta}^{\tau,\lambda}_{\text{reg}}(\eta))-\nabla\mathcal{L}(\theta^*_{\text{reg}}), \widehat{\theta}^{\tau,\lambda}_{\text{reg}}(\eta)-\theta^*_{\text{reg}}\big\rangle\notag\\
    =&\left\{\widehat{\theta}^{\tau,\lambda}_{\text{reg}}(\eta)-\theta^*_{\text{reg}}\right\}^\top\nabla^2\mathcal{L}(\widetilde{\theta}^{\tau,\lambda}_{\text{reg}}(\eta))\left\{\widehat{\theta}^{\tau,\lambda}_{\text{reg}}(\eta)-\theta^*_{\text{reg}}\right\}\notag\\
    =&\left[\Sigma^{1/2}\left\{\widehat{\theta}^{\tau,\lambda}_{\text{reg}}(\eta)-\theta^*_{\text{reg}}\right\}\right]^\top\left\{\Sigma^{-1/2}\nabla^2\mathcal{L}(\widetilde{\theta}^{\tau,\lambda}_{\text{reg}}(\eta))\Sigma^{-1/2}\right\}\left[\Sigma^{1/2}\left\{\widehat{\theta}^{\tau,\lambda}_{\text{reg}}(\eta)-\theta^*_{\text{reg}}\right\}\right]\notag\\
    \ge& \left\|\Sigma^{1/2}\left\{\widehat{\theta}^{\tau,\lambda}_{\text{reg}}(\eta)-\theta^*_{\text{reg}}\right\}\right\|^2_2\lambda_{\min}\left(\Sigma^{-1/2}\nabla^2\mathcal{L}(\widetilde{\theta}^{\tau,\lambda}_{\text{reg}}(\eta))\Sigma^{-1/2}\right),
\end{align}
which identifies a lower bound of the right-hand side of (\ref{eq.tmp_1}).

On the other hand, using $\Sigma$ and the Cauchy–Schwarz inequality, the right-hand side of (\ref{eq.tmp_1}) is upper bounded as:
\begin{align}\label{eq.tmp_3}
    -\big\langle\nabla\mathcal{L}(\theta^*_{\text{reg}}), \widehat{\theta}^{\tau,\lambda}_{\text{reg}}(\eta)-\theta^*_{\text{reg}}\big\rangle
    =&\left|\left\{\nabla\mathcal{L}(\theta^*_{\text{reg}})\right\}^\top\left\{\widehat{\theta}^{\tau,\lambda}_{\text{reg}}(\eta)-\theta^*_{\text{reg}}\right\}\right|\notag\\
    =&\left|\left\{\Sigma^{-1/2}\nabla\mathcal{L}(\theta^*_{\text{reg}})\right\}^\top\Sigma^{1/2}\left\{\widehat{\theta}^{\tau,\lambda}_{\text{reg}}(\eta)-\theta^*_{\text{reg}}\right\}\right|\notag\\
    \le&\left\|\Sigma^{-1/2}\nabla\mathcal{L}(\theta^*_{\text{reg}})\right\|_2\left\|\Sigma^{1/2}\left\{\widehat{\theta}^{\tau,\lambda}_{\text{reg}}(\eta)-\theta^*_{\text{reg}}\right\}\right\|_2,
\end{align}
{where the first step holds since $-\big\langle\nabla\mathcal{L}(\theta^*_{\text{reg}}), \widehat{\theta}^{\tau,\lambda}_{\text{reg}}(\eta)-\theta^*_{\text{reg}}\big\rangle$ admits a positive lower bound. Specifically, it is lower bounded by (\ref{eq.tmp_1}), which in turn is strictly positive by (\ref{eq.tmp_2}).
}

Comparing (\ref{eq.tmp_2}) and (\ref{eq.tmp_3}) via (\ref{eq.tmp_1}), we obtain that 
\begin{align}\label{eq.tmp_3_1}
    \left\|\Sigma^{1/2}\left\{\widehat{\theta}^{\tau,\lambda}_{\text{reg}}(\eta)-\theta^*_{\text{reg}}\right\}\right\|_2\lambda_{\min}\left(\Sigma^{-1/2}\nabla^2\mathcal{L}(\widetilde{\theta}^{\tau,\lambda}_{\text{reg}}(\eta))\Sigma^{-1/2}\right)\le \left\|\Sigma^{-1/2}\nabla\mathcal{L}(\theta^*_{\text{reg}})\right\|_2.
\end{align}

{\bf\underline{Step 1:} Upper bounding} $\left\|\Sigma^{-1/2}\nabla\mathcal{L}(\theta^*_{\text{reg}})\right\|_2$ {\bf in (\ref{eq.tmp_3_1}).}

By the triangle inequality, $\left\|\Sigma^{-1/2}\nabla\mathcal{L}(\theta^*_{\text{reg}})\right\|_2$ is upper bounded as:
\begin{align}\label{eq.tmp_4}
    \left\|\Sigma^{-1/2}\nabla\mathcal{L}(\theta^*_{\text{reg}})\right\|_2\le \underbrace{\left\|\Sigma^{-1/2}\mathbb{E}\left\{\nabla\mathcal{L}(\theta^*_{\text{reg}})\right\}\right\|_2}_{\text{(I)}}+\underbrace{\left\|\Sigma^{-1/2}\left[\nabla\mathcal{L}(\theta^*_{\text{reg}})-\mathbb{E}\left\{\nabla\mathcal{L}(\theta^*_{\text{reg}})\right\}\right]\right\|_2}_{\text{(II)}}.
\end{align}

{\bf\underline{Step 1.1:} Examining (I).}

For $i\in[n]$, letting $\zeta_i\triangleq \Sigma^{-1/2}\mathrm{Z}_i$ and $\nu^*_i= \widetilde{Y}_i-\langle \mathrm{Z}_i,\theta^*_{\text{reg}}\rangle$. {Then by (\ref{eq.true_label_appx}) and (\ref{eq.noisy_label_appx})}, $\nu^*_i= \widetilde{Y}_i-\langle \mathrm{Z}_i,\theta^*_{\text{reg}}\rangle=\varepsilon'_i+\varsigma(\mathrm{X}_i,\mathrm{Y}_i)\varepsilon''_i$, yielding that $\mathbb{E}(\nu^*_i)=0$ {and $\mathbb{E}(\nu^*_i\zeta_i)=0$}. By (\ref{eq.loss_deriv_1}), we have
\begin{align}
    \text{(I)}=&\left\|-\frac{1}{N}\sum_{i=1}^{N}\mathbb{E}\left\{\text{sgn}(\nu^*_i)\ell'_\tau(|\nu^*_i|)\Sigma^{-1/2}\mathrm{Z}_i\right\}+\lambda \Sigma^{-1/2}\theta^*_{\text{reg}}\right\|_2\notag\\
    \le& \left\|-\frac{1}{N}\sum_{i=1}^{N}\mathbb{E}\left\{\text{sgn}(\nu^*_i)\ell'_\tau(|\nu^*_i|)\Sigma^{-1/2}\mathrm{Z}_i\right\}\right\|_2+\lambda \|\Sigma^{-1/2}\theta^*_{\text{reg}}\|_2\notag\\
    =&\sup_{u\in\mathbb{S}^{d-1}}\left[-\frac{1}{N}\sum_{i=1}^{N}\mathbb{E}\left\{\text{sgn}(\nu^*_i)\ell'_\tau(|\nu^*_i|)u^\top\zeta_i\right\}\right]+\lambda \|\Sigma^{-1/2}\theta^*_{\text{reg}}\|_2, \label{eq.tmp_4_1}
\end{align}
where the last step is due to Lemma \ref{lemma.matrix} \ref{lemma.matrix_2}. 

For any $u\in\mathbb{S}^{d-1}$, we have that by (\ref{eq.ell_deriv_1}),
\begin{align*}
    &\left|\mathbb{E}\left\{\text{sgn}(\nu^*_i)\ell'_\tau(|\nu^*_i|)u^\top\zeta_i\right\}\right|\\
    =&\left|\mathbb{E}\left[\text{sgn}(\nu^*_i)\left\{|\nu^*_i|\mathbbm{1}(|\nu^*_i|\le \tau)u^\top\zeta_i+\frac{\tau\ell'(|\nu^*_i|)}{\ell'(\tau)}\mathbbm{1}(|\nu^*_i|> \tau)u^\top\zeta_i\right\}\right]\right|\notag\\
    =&\left|\mathbb{E}\left[\text{sgn}(\nu^*_i)\left\{|\nu^*_i| u^\top\zeta_i-|\nu^*_i|\mathbbm{1}(|\nu^*_i|> \tau)u^\top\zeta_i+\frac{\tau\ell'(|\nu^*_i|)}{\ell'(\tau)}\mathbbm{1}(|\nu^*_i|> \tau)u^\top\zeta_i\right\}\right]\right|\notag\\
    % =&\left|\mathbb{E}\left\{-\nu^*_i\mathbbm{1}(|\nu^*_i|> \tau)u^\top\zeta_i+\frac{\tau\ell'(\nu^*_i)}{\ell'(\tau)}\mathbbm{1}(|\nu^*_i|> \tau)u^\top\zeta_i\right\}\right|\notag\\
    =&\left|\mathbb{E}\left[-\text{sgn}(\nu^*_i)\left\{|\nu^*_i|-\frac{\tau\ell'(|\nu^*_i|)}{\ell'(\tau)}\right\}\mathbbm{1}(|\nu^*_i|> \tau)u^\top\zeta_i\right]\right|\notag\\
    \le&\mathbb{E}\left[\left\{|\nu^*_i|-\frac{\tau\ell'(|\nu^*_i|)}{\ell'(\tau)}\right\}\mathbbm{1}(|\nu^*_i|> \tau)|u^\top\zeta_i|\right]\notag\\
    =&\mathbb{E}\left\{|\nu^*_i|\mathbbm{1}(|\nu^*_i|> \tau)|u^\top\zeta_i|\right\}-\mathbb{E}\left\{\frac{\tau\ell'(|\nu^*_i|)}{\ell'(\tau)}\mathbbm{1}(|\nu^*_i|> \tau)|u^\top\zeta_i|\right\}\notag\\
    \le&\mathbb{E}\left\{|\nu^*_i|\mathbbm{1}(|\nu^*_i|> \tau)|u^\top\zeta_i|\right\}\notag\\
    \le&\tau^{-1}\mathbb{E}\left\{\nu^{*2}_i|u^\top\zeta_i|\right\}\notag\\
    =&\tau^{-1}\mathbb{E}\left\{\Bar{\varsigma}^2(\mathrm{X}_i)|u^\top\zeta_i|\right\},
\end{align*}
where the second step follows from $\mathbbm{1}(|\nu^*_i|\le \tau)+\mathbbm{1}(|\nu^*_i|> \tau)=1$; the third step uses $\mathbb{E}(\nu^*_i\zeta_i)=0$ together with the decomposition $\nu^*_i=\mathrm{sgn}(\nu^*_i)|\nu^*_i|$; the fourth step holds since $\left||\nu^*_i|-\frac{\tau\ell'(|\nu^*_i|)}{\ell'(\tau)}\right|= |\nu^*_i|-\frac{\tau\ell'(|\nu^*_i|)}{\ell'(\tau)}$ if $|\nu^*_i|\ge \tau$ by (\ref{eq.ell_deriv_1}); the penultimate step uses $\mathbbm{1}(|\nu^*_i|> \tau)\le \frac{|\nu^*_i|}{\tau}$; and the final step follows by integrating out $\varepsilon'_i$ and $\varepsilon''_i$, with $\Bar{\varsigma}^2(\cdot)$ defined in Condition \ref{appdx.regularity_condition} \ref{regu.2}.

Furthermore, using the Cauchy–Schwarz inequality, we obtain that 
\begin{align}\label{eq.tmp_5}
    \left|\mathbb{E}\left\{\text{sgn}(\nu^*_i)\ell'_\tau(|\nu^*_i|)u^\top\zeta_i\right\}\right|
    \le& \tau^{-1}\mathbb{E}\left\{\Bar{\varsigma}(\mathrm{X}_i)|\Bar{\varsigma}(\mathrm{X}_i)u^\top\zeta_i|\right\}\notag\\
    \le& \tau^{-1}\sqrt{\mathbb{E}\{\Bar{\varsigma}^2(\mathrm{X}_i)\}\mathbb{E}\left\{\Bar{\varsigma}^2(\mathrm{X}_i)(u^\top\zeta_i)^2\right\}}\notag\\
    =& \tau^{-1}\sqrt{\mathbb{E}\{\Bar{\varsigma}^2(\mathrm{X}_i)\}\cdot u^\top \Sigma^{-1/2} \mathbb{E}\left\{\Bar{\varsigma}^2(\mathrm{X}_i)\mathrm{Z}_i\mathrm{Z}^\top_i\right\}\Sigma^{-1/2} u}\notag\\
    \le& \frac{\sigma}{\tau} \sqrt{\sup_{u\in\mathbb{S}^{d-1}} \left[u^\top \Sigma^{-1/2} \mathbb{E}\left\{\Bar{\varsigma}^2(\mathrm{X}_i)\mathrm{Z}_i\mathrm{Z}^\top_i\right\}\Sigma^{-1/2} u\right]}\notag\\
    =& \frac{\sigma}{\tau}\sqrt{\lambda_{\max}(\Sigma^{-1/2} \mathbb{E}\left\{\Bar{\varsigma}^2(\mathrm{X}_i)\mathrm{Z}_i\mathrm{Z}^\top_i\right\}\Sigma^{-1/2})}\notag\\
    \triangleq& \frac{\sigma}{\tau} \varpi^{1/2},
\end{align}
where the third step holds by the definition of $\zeta_i$ given at the beginning of Step 1.1, the fourth step uses the definition of $\sigma$ in Condition \ref{appdx.regularity_condition} \ref{regu.2}, the fifth step is due to Lemma \ref{lemma.matrix} \ref{lemma.matrix_1}, and the last step arises by letting {$\varpi\triangleq \lambda_{\max}(\Sigma^{-1/2} \mathbb{E}\left\{\Bar{\varsigma}^2(\mathrm{X}_i)\mathrm{Z}_i\mathrm{Z}^\top_i\right\}\Sigma^{-1/2})$}. 

Combining (\ref{eq.tmp_4_1}) and (\ref{eq.tmp_5}) gives that the first term in (\ref{eq.tmp_4}) is upper bounded as:
\begin{align}\label{eq.tmp_6}
    \text{(I)}\le \frac{\sigma}{\tau} \varpi^{1/2}+\lambda \|\Sigma^{-1/2}\theta^*_{\text{reg}}\|_2.
\end{align}

{\bf\underline{Step 1.2:} Examining (II).}

Next, we provide an upper bound on the second term (II) in (\ref{eq.tmp_4}). Define 
$\zeta^\dagger\triangleq-\sqrt{N}\Sigma^{-1/2}\left[\nabla\mathcal{L}(\theta^*_{\text{reg}})-\mathbb{E}\left\{\nabla\mathcal{L}(\theta^*_{\text{reg}})\right\}\right]$. Then $\text{(II)}=\left\|\frac{1}{\sqrt{N}}\zeta^\dagger\right\|_2$ by (\ref{eq.tmp_4}). By (\ref{eq.loss_deriv_1}),
\begin{align}\label{eq.tmp_7}
    \zeta^\dagger=&\frac{1}{\sqrt{N}}\sum_{i=1}^{N}\text{sgn}(\nu^*_i)\ell'_\tau(|\nu^*_i|)\Sigma^{-1/2}\mathrm{Z}_i-{\lambda}{\sqrt{N}} \Sigma^{-1/2}\theta^*_{\text{reg}}\notag\\
    &+\mathbb{E}\left\{\frac{1}{\sqrt{N}}\sum_{i=1}^{N}\text{sgn}(\nu^*_i)\ell'_\tau(|\nu^*_i|)\Sigma^{-1/2}\mathrm{Z}_i-{\lambda}{\sqrt{N}} \Sigma^{-1/2}\theta^*_{\text{reg}}\right\}\notag\\
    =&\frac{1}{\sqrt{N}}\sum_{i=1}^{N}\left[\text{sgn}(\nu^*_i)\ell'_\tau(|\nu^*_i|)\Sigma^{-1/2}\mathrm{Z}_i-\mathbb{E}\left\{\text{sgn}(\nu^*_i)\ell'_\tau(|\nu^*_i|)\Sigma^{-1/2}\mathrm{Z}_i\right\}\right]\notag\\
    \triangleq&\frac{1}{\sqrt{N}}\sum_{i=1}^{N}\Bar{\zeta}_i,
\end{align}
where $\Bar{\zeta}_i\triangleq \text{sgn}(\nu^*_i)\ell'_\tau(|\nu^*_i|)\Sigma^{-1/2}\mathrm{Z}_i-\mathbb{E}\left\{\text{sgn}(\nu^*_i)\ell'_\tau(|\nu^*_i|)\Sigma^{-1/2}\mathrm{Z}_i\right\}=\text{sgn}(\nu^*_i)\ell'_\tau(|\nu^*_i|)\zeta_i-\mathbb{E}\{\text{sgn}(\nu^*_i)\ell'_\tau(|\nu^*_i|)\zeta_i\}$, with $\zeta_i\triangleq \Sigma^{-1/2}\mathrm{Z}_i$, as defined at the beginning of Step 1.1. % Hence, $\text{(II)}=\left\|\frac{1}{\sqrt{N}}\zeta^\dagger\right\|_2$ by (\ref{eq.tmp_4}). 

To establish an upper bound for $\|\zeta^\dagger\|_2$, we apply Lemma \ref{lemma.upper_bound} by verifying {(\ref{lemma.upper_bound_cond})}. Specifically, for any $u\in\mathbb{S}^{d-1}$ and $|t|\le\mathtt{g}$, where $\mathtt{g}\ge\sqrt{2d}$ is a constant to be specified later,
\begin{align*}
    \mathbb{E}\{\exp(tu^\top\zeta^\dagger)\}
    =&\prod_{i=1}^{N}\mathbb{E}\left\{\exp\left(\frac{t}{\sqrt{N}}u^\top \Bar{\zeta}_i\right)\right\}\\
    \le&\prod_{i=1}^{N}\mathbb{E}\left\{1+\frac{t}{\sqrt{N}}u^\top \Bar{\zeta}_i+\frac{1}{2}\left(\frac{t}{\sqrt{N}}u^\top \Bar{\zeta}_i\right)^2\exp\left(\left|\frac{t}{\sqrt{N}}u^\top \Bar{\zeta}_i\right|\right)\right\}\\
    =&\prod_{i=1}^{N}\left[1+\frac{t^2}{2N}\mathbb{E}\left\{\left(u^\top \Bar{\zeta}_i\right)^2\exp\left(\frac{|t|}{\sqrt{N}}|u^\top \Bar{\zeta}_i|\right)\right\}\right]
\end{align*}
where the first step is due to the independence of $\xi_i$, the second step is due to the inequality $e^x\le 1+x+\frac{x^2}{2}e^{|x|}$ for any $x$, and the third step holds since $\mathbb{E}(\Bar{\zeta}_i)=0$ by the definition of $\Bar{\zeta}_i$ given after (\ref{eq.tmp_7}).

By the definition of $\Bar{\zeta}_i$, (\ref{eq.ell_deriv_1}), and (\ref{eq.tmp_5}), we obtain that for $u\in\mathbb{S}^{d-1}$, 
\begin{align*}
    |u^\top \Bar{\zeta}_i|
    =&\left|\text{sgn}(\nu^*_i)\ell'_\tau(|\nu^*_i|)u^\top\zeta_i-\mathbb{E}\left\{\text{sgn}(\nu^*_i)\ell'_\tau(|\nu^*_i|)u^\top\zeta_i\right\}\right|\\
    \le& |\text{sgn}(\nu^*_i)\ell'_\tau(|\nu^*_i|)u^\top\zeta_i|+\left|\mathbb{E}\left\{\text{sgn}(\nu^*_i)\ell'_\tau(|\nu^*_i|)u^\top\zeta_i\right\}\right|\\
    \le& \tau|u^\top\zeta_i|\mathbbm{1}(|\nu^*_i|\le \tau)+\frac{\tau|\nu_i|^\kappa}{\ell'(\tau)}|u^\top\zeta_i|\mathbbm{1}(|\nu^*_i|> \tau)+\frac{\sigma}{\tau}\varpi^{1/2},
\end{align*}
and 
\begin{align*}
    (u^\top \Bar{\zeta}_i)^2
    =&\left[\text{sgn}(\nu^*_i)\ell'_\tau(|\nu^*_i|)u^\top\zeta_i-\mathbb{E}\left\{\text{sgn}(\nu^*_i)\ell'_\tau(|\nu^*_i|)u^\top\zeta_i)\right\}\right]^2\\
    \le&2\left[\{\ell'_\tau(|\nu^*_i|)u^\top\zeta_i\}^2+\{\mathbb{E}(\ell'_\tau(|\nu^*_i|)u^\top\zeta_i)\}^2\right]\\
    \le&2\left\{\nu^{*2}_i(u^\top\zeta_i)^2+\frac{\sigma^2}{\tau^2}\varpi\right\}.
\end{align*}

Consequently, for $|t|\le\mathtt{g}$ with $\mathtt{g}\ge\sqrt{2d}$ and $u\in\mathbb{S}^{d-1}$, we can further upper bound $\mathbb{E}\{\exp(tu^\top\zeta^\dagger)\}$ as:
\begin{align}\label{eq.tmp_8}
    &\mathbb{E}\{\exp(tu^\top\zeta^\dagger)\}\notag\\
    \le &\prod_{i=1}^{N} \Bigg(1+ \frac{t^2}{N}\exp\left(\frac{\sigma|t|}{\tau\sqrt{N}}\varpi^{1/2}\right)\bigg[\frac{\sigma^2}{\tau^2}\varpi\ \mathbb{E}\left\{\exp\left(\frac{\tau|t|}{\sqrt{N}}|u^\top \zeta_i|\right)\mathbbm{1}(|\nu^*_i|\le\tau)\right\}\notag\\
    &\hspace{0.8cm}+\frac{\sigma^2}{\tau^2}\varpi\ \mathbb{E}\left\{\exp\left(\frac{\tau|t|}{\ell'(\tau)\sqrt{N}}|u^\top \zeta_i||\nu^*_i|^\kappa\right)\mathbbm{1}(|\nu^*_i|>\tau)\right\}+\mathbb{E}\left\{\nu^{*2}_i\left(u^\top{\zeta}_i\right)^2\exp\left(\frac{\tau|t|}{\sqrt{N}}|u^\top \zeta_i|\right)\right\}\bigg]\Bigg)\notag\\
    =&\prod_{i=1}^{N} \Bigg(1+ \frac{t^2}{N}\exp\left(\frac{\sigma|t|}{\tau\sqrt{N}}\varpi^{1/2}\right)\bigg[\frac{\sigma^2}{\tau^2}\varpi\ \mathbb{E}\left\{\exp\left(\frac{\tau|t|}{\sqrt{N}}|u^\top \zeta_i|\right)\right\}\notag\\
    &\hspace{0.8cm}+\frac{\sigma^2}{\tau^2}\varpi\ \mathbb{E}\left\{\exp\left(\frac{\tau|t|}{\ell'(\tau)\sqrt{N}}|u^\top \zeta_i||\nu^*_i|^\kappa\right)\mathbbm{1}(|\nu^*_i|>\tau)\right\}+\mathbb{E}\left\{\Bar{\varsigma}^2(\mathrm{X}_i)\left(u^\top{\zeta}_i\right)^2\exp\left(\frac{\tau|t|}{\sqrt{N}}|u^\top \zeta_i|\right)\right\}\bigg]\Bigg)\notag\\
    \le&\prod_{i=1}^{N} \Bigg[1+ \frac{t^2}{N}\exp\left\{\frac{\sqrt{2}(d+s)}{N}\varpi^{1/2}\right\}\Bigg\{\frac{(d+s)\varpi}{N}\ \underbrace{\left[\mathbb{E}\left\{\exp\left(\sqrt{2}\sigma|u^\top \zeta_i|\right)\right\}+\mathbb{E}\left\{\exp\left({\sqrt{2}\sigma}|u^\top \zeta_i||\nu^*_i|^\kappa\right)\right\}\right]}_{\text{(II)-a}}\notag\\
    &\hspace{0.8cm}+\underbrace{\mathbb{E}\left\{\Bar{\varsigma}^2(\mathrm{X}_i)\left(u^\top{\zeta}_i\right)^2\exp\left(\sqrt{2}\sigma|u^\top \zeta_i|\right)\right\}}_{\text{(II)-b}}\Bigg\}\Bigg],
\end{align}
where the second step holds by integrating out the noise variables $\varepsilon'_i$ and $\varepsilon''_i$, with {$\Bar{\varsigma}^2(\cdot)$ given in Section \ref{appdx.regularity_condition} \ref{regu.2}}, and the last step is true when {$\frac{\tau|t|}{\sqrt{N}}\le\sqrt{2}\sigma$} and {$\frac{\tau|t|}{\ell'(\tau)\sqrt{N}}\le\sqrt{2}\sigma$}, which is ensured by letting {$|t|\le\mathtt{g}=\sqrt{2(d+s)}$} and {$\max(\tau,{\tau}/{\ell'(\tau)})\le\sigma\sqrt{N/(d+s)}$} {as required in Theorem \ref{thm.robust_main_dro}}.

We next identify upper bounds of (II)-a and (II)-b in (\ref{eq.tmp_8}). For (II)-a with any $u\in\mathbb{S}^{d-1}$, we have 
\begin{align}\label{eq.tmp_8_0_1}
    \text{(II)-a}\le&\sup_{v\in\mathbb{S}^{d-1}}\left[\mathbb{E}\left\{\exp\left(\sqrt{2}\sigma|u^\top \zeta_i|\right)\right\}+\mathbb{E}\left\{\exp\left({\sqrt{2}\sigma}|u^\top \zeta_i||\nu^*_i|^\kappa\right)\right\}\right]\notag\\
    \le&\sup_{v\in\mathbb{S}^{d-1}}\mathbb{E}\left[\exp\left\{\sqrt{2}\sigma(v^\top \zeta_i)\right\}\right]+\sup_{v\in\mathbb{S}^{d-1}}\mathbb{E}\left[\exp\left\{{\sqrt{2}\sigma}v^\top \big(\zeta_i|\nu^*_i|^\kappa\big)\right\}\right]
    \le\mathcal{K}_1
\end{align}
for some constant $\mathcal{K}_1>0$, where the second step holds by the symmetry of $\mathbb{S}^{d-1}$ and the subadditivity of supremum, and the last step is due to the sub-Gaussian assumption in Condition \ref{appdx.regularity_condition} \ref{regu.3} and the equivalence properties in Lemma \ref{lemma.sub_gaussian_property}, with $\mathcal{K}_1$ denoting a constant depending on $\sigma$ in Condition \ref{appdx.regularity_condition} \ref{regu.2}, $\alpha_2$ in Condition \ref{appdx.regularity_condition} \ref{regu.3} \ref{regu.3.a}, and $\alpha_3$ in Condition \ref{appdx.regularity_condition} \ref{regu.3} \ref{regu.3.b}.

For (II)-b with any $u\in\mathbb{S}^{d-1}$, we have that
\begin{align}
    \text{(II)-b}\le&\sup_{v\in\mathbb{S}^{d-1}}\mathbb{E}\left\{\Bar{\varsigma}^2(\mathrm{X}_i)\left(v^\top{\zeta}_i\right)^2\exp\left(\sqrt{2}\sigma|v^\top \zeta_i|\right)\right\}\notag\\
    \le&\sup_{v\in\mathbb{S}^{d-1}}\bigg\{\left[{\mathbb{E}\{\Bar{\varsigma}^{2+\iota_1}(\mathrm{X}_i)\}}\right]^{\frac{2}{2+\iota_1}}\left[\mathbb{E}\{(v^\top \zeta_i)^{2(4/\iota_1+2)}\}\right]^{\iota_1/(4+2\iota_1)}\notag\\
    &\ \ \ \ \ \ \ \ \ \ \ \ \ \ \ \left(\mathbb{E}[\exp\{\sqrt{2}\sigma(4/\iota_1+2)|v^\top \zeta_i|\}]\right)^{\iota_1/(4+2\iota_1)}\bigg\}\notag\\
    \le& \left[{\mathbb{E}\{\Bar{\varsigma}^{2+\iota_1}(\mathrm{X}_i)\}}\right]^{\frac{2}{2+\iota_1}}\cdot \sup_{v\in\mathbb{S}^{d-1}}\left[\mathbb{E}\{(v^\top \zeta_i)^{2(4/\iota_1+2)}\}\right]^{\iota_1/(4+2\iota_1)}\notag\\
    &\ \cdot\sup_{v\in\mathbb{S}^{d-1}} \left(\mathbb{E}[\exp\{\sqrt{2}\sigma(4/\iota_1+2)(v^\top \zeta_i)\}]\right)^{\iota_1/(4+2\iota_1)}\notag\\
    \le&\mathcal{K}_2,\label{eq.tmp_8_0_2}
\end{align}
for some constant $\mathcal{K}_2>0$, where the second step is due to holder's inequality, with $\iota_1$ given in Condition \ref{appdx.regularity_condition} \ref{regu.2}, and the last step holds by Condition \ref{appdx.regularity_condition} \ref{regu.2} and the sub-Gaussian assumption in Condition \ref{appdx.regularity_condition} \ref{regu.3}, as well as the equivalence properties in Lemma \ref{lemma.sub_gaussian_property}, with the constant $\mathcal{K}_2$ depending on $\alpha_1$, $\iota_1$, and $\sigma$ in Condition \ref{appdx.regularity_condition} \ref{regu.2} and $\alpha_2$ in Condition \ref{appdx.regularity_condition} \ref{regu.3}.

Applying (\ref{eq.tmp_8_0_1}) and (\ref{eq.tmp_8_0_2}) to (\ref{eq.tmp_8}), we obtain that, for $|t|\le\mathtt{g}=\sqrt{2(d+s)}$ and $u\in\mathbb{S}^{d-1}$,
\begin{align*}
    \mathbb{E}\{\exp(tu^\top\zeta^\dagger)\}
    \le& \prod_{i=1}^{N} \Bigg(1+ \frac{t^2}{N}\exp\left(\frac{\sqrt{2}(d+s)}{N}\varpi^{1/2}\right)\bigg\{\frac{(d+s)\varpi}{N}\ \mathcal{K}_1+\mathcal{K}_2\bigg\}\Bigg)\le \exp(\mathcal{K}^{\dagger} t^2/2),
\end{align*}
where $\mathcal{K}^\dagger\triangleq \max\left\{{2e^{\sqrt{2}\varpi^{1/2}}(\varpi\mathcal{K}_1+\mathcal{K}_2)},1.1\right\}$, and the last step is due to the inequality $1+x\le e^x$, along with the {assumption $N>d+s$ in Theorem \ref{thm.robust_main_dro}}. Hence, $\mathcal{K}^\dagger$ is a constant depending on $\alpha_1$, $\iota_1$, and $\sigma$ in Condition \ref{appdx.regularity_condition} \ref{regu.2}, $\alpha_2$ in Condition \ref{appdx.regularity_condition} \ref{regu.3} \ref{regu.3.a}, $\alpha_3$ in Condition \ref{appdx.regularity_condition} \ref{regu.3} \ref{regu.3.b}, and $\varpi$ defined after (\ref{eq.tmp_5}). By rescaling $\zeta^\dagger$ and $\mathtt{g}$, we further obtain that, for $|t|\le\Bar{\mathtt{g}}\triangleq\sqrt{\mathcal{K}^\dagger}\mathtt{g}=\sqrt{2\mathcal{K}^\dagger(d+s)}$, and $u\in\mathcal{S}^{d-1}$,
\begin{align*}
    \mathbb{E}\left[\exp\left\{tu^\top\left(\zeta^\dagger/\sqrt{\mathcal{K}^\dagger}\right)\right\}\right]\le e^{t^2/2}.
\end{align*}
Consequently, by Lemma \ref{lemma.upper_bound} with $V_0=D_0=I_{d\times d}$, we obtain that for $\sqrt{2d}/18<\mathtt{x}^*\le \mathtt{c}_4$,
\begin{align}\label{eq.tmp_8_1}
    \mathbb{P}\Big\{\big\|\zeta^\dagger\big\|^2_2>\mathcal{K}^\dagger (d+6\mathtt{x}^*)\Big\}=\mathbb{P}\Big(\big\|\zeta^\dagger/\sqrt{\mathcal{K}^\dagger}\big\|^2_2>d+6\mathtt{x}^*\Big)\le 2e^{-\mathtt{x}^*}+8.4e^{-\mathtt{c}_4},
\end{align}
where $\mathtt{c}_4=\frac{1}{2}\left(\Bar{\mathtt{g}}^{2}-d\log\frac{d+\Bar{\mathtt{g}}^{2}}{d}\right)$. By taking the derivative of $\mathtt{c}_4$ with respect to $\Bar{\mathtt{g}}$, we obtain that $\frac{d \mathtt{c}_4}{d \Bar{\mathtt{g}}}=\frac{\Bar{\mathtt{g}}^{3}}{d+\Bar{\mathtt{g}}^{2}}>0$, and hence, $\mathtt{c}_4$ is increasing in $\Bar{\mathtt{g}}$. Therefore, since $\Bar{\mathtt{g}}=\sqrt{2\mathcal{K}^\dagger(d+s)}\ge \sqrt{2.2(d+s)}$, we obtain that
\begin{align*}
    \mathtt{c}_4\ge \frac{1}{2}\left\{2.2(d+s)-d\log \left(3.2+\frac{2.2s}{d}\right)\right\}\ge \frac{1}{2}\left\{2.2(d+s)-d\left(0.2\cdot\frac{s}{d}+1.8\right)\right\}=s+0.2d,
\end{align*}
where the second inequality holds since $\log(3.2+2.2x)\le 0.2x+1.8$ for $x\ge 0$. By taking $\mathtt{x}^*=s+0.2d$ in (\ref{eq.tmp_8_1}), we obtain that, {when $d>10$}, with probability at least $1-3e^{-s}$, $\big\|\zeta^\dagger\big\|^2_2\le\mathcal{K}^\dagger (2.2d+6s)$, and consequently,
\begin{align}\label{eq.tmp_8_2}
    \text{(II)}=\frac{1}{\sqrt{N}}\big\|\zeta^\dagger\big\|_2\le\sqrt{\frac{\mathcal{K}^\dagger (2.2d+6s)}{N}}.
\end{align}

Combining (\ref{eq.tmp_4}), (\ref{eq.tmp_6}), and (\ref{eq.tmp_8_2}), we obtain that, with probability at least $1-3e^{-s}$
\begin{align}\label{eq.tmp_8_3}
    \left\|\Sigma^{-1/2}\nabla\mathcal{L}(\theta^*_{\text{reg}})\right\|_2\le \frac{\sigma}{\tau} \varpi^{1/2}+\lambda \|\Sigma^{-1/2}\theta^*_{\text{reg}}\|_2+\sqrt{\frac{\mathcal{K}^\dagger (2.2d+6s)}{N}}.
\end{align}

{\bf\underline{Step 2:} Lower bounding} $\lambda_{\min}\left(\Sigma^{-1/2}\nabla^2\mathcal{L}(\widetilde{\theta}^{\tau,\lambda}_{\text{reg}}(\eta))\Sigma^{-1/2}\right)$ {\bf in (\ref{eq.tmp_3_1}).}

Let $\mathscr{m}^*$ be a given constant, and we can choose $\eta\in[0,1]$ such that 
\begin{align}\label{eq.tmp_8_4}
    \left\|\Sigma^{1/2}\left\{\widehat{\theta}^{\tau,\lambda}_{\text{reg}}(\eta)-\theta^*_{\text{reg}}\right\}\right\|_2\le\mathscr{m}^*
\end{align}
Indeed, if $\left\|\Sigma^{1/2}\left(\widehat{\theta}^{\tau,\lambda}_{\text{reg}}-\theta^*_{\text{reg}}\right)\right\|_2\le\mathscr{m}^*$, then we take $\eta=1$; otherwise, we can always choose some $\eta<1$ such that $\left\|\Sigma^{1/2}\left\{\widehat{\theta}^{\tau,\lambda}_{\text{reg}}(\eta)-\theta^*_{\text{reg}}\right\}\right\|_2=\mathscr{m}^*$. Since $\widetilde{\theta}^{\tau,\lambda}_{\text{reg}}(\eta)$ is a convex combination of $\widehat{\theta}^{\tau,\lambda}_{\text{reg}}(\eta)$ and $\theta^*_{\text{reg}}$, then replacing $\widehat{\theta}^{\tau,\lambda}_{\text{reg}}(\eta)$ in (\ref{eq.tmp_8_4}) with $\widetilde{\theta}^{\tau,\lambda}_{\text{reg}}(\eta)$ yields $\left\|\Sigma^{1/2}\left\{\widetilde{\theta}^{\tau,\lambda}_{\text{reg}}(\eta)-\theta^*_{\text{reg}}\right\}\right\|_2\le\mathscr{m}^*$. Consequently, 
\begin{align}
    &\lambda_{\min}\left(\Sigma^{-1/2}\nabla^2\mathcal{L}(\widetilde{\theta}^{\tau,\lambda}_{\text{reg}}(\eta))\Sigma^{-1/2}\right)\notag\\
    \ge& \inf_{{\theta_{\text{reg}}}:\left\|\Sigma^{1/2}\left(\theta_{\text{reg}}-\theta^*_{\text{reg}}\right)\right\|_2\le\mathscr{m}^*}\lambda_{\min}\left(\Sigma^{-1/2}\nabla^2\mathcal{L}(\theta_{\text{reg}})\Sigma^{-1/2}\right)\notag\\
    =& \inf_{{\theta_{\text{reg}}}:\left\|\Sigma^{1/2}\left(\theta_{\text{reg}}-\theta^*_{\text{reg}}\right)\right\|_2\le\mathscr{m}^*} \lambda_{\min} \left(\Sigma^{-1/2}\left\{\frac{1}{N}\sum_{i=1}^{N}\mathrm{Z}_i\mathrm{Z^\top_i}\ell''_{\tau}(|\nu_i|)+\lambda I_{d\times d}\right\}\Sigma^{-1/2}\right) \notag\\
    =&\inf_{{\theta_{\text{reg}}}:\left\|\Sigma^{1/2}\left(\theta_{\text{reg}}-\theta^*_{\text{reg}}\right)\right\|_2\le\mathscr{m}^*} \lambda_{\min}\left(\frac{1}{N}\sum_{i=1}^{N}\zeta_i\zeta_i^\top\mathbbm{1}(|\nu_i|\le \tau)+\frac{1}{N}\sum_{i=1}^{N}\zeta_i\zeta_i^\top\frac{\tau\ell''(|\nu_i|)}{\ell'(\tau)}\mathbbm{1}(|\nu_i|>\tau)+\lambda\Sigma^{-1}\right) \notag\\
    \ge& \inf_{{\theta_{\text{reg}}}:\left\|\Sigma^{1/2}\left(\theta_{\text{reg}}-\theta^*_{\text{reg}}\right)\right\|_2\le\mathscr{m}^*} \lambda_{\min}\left(\frac{1}{N}\sum_{i=1}^{N}\zeta_i\zeta_i^\top\mathbbm{1}(|\nu_i|\le \tau)\right)+\lambda_{\min}(\lambda\Sigma^{-1})\notag\\
    =& \inf_{{\theta_{\text{reg}}}:\left\|\Sigma^{1/2}\left(\theta_{\text{reg}}-\theta^*_{\text{reg}}\right)\right\|_2\le\mathscr{m}^*}\inf_{u\in\mathbb{S}^{d-1}}\left\{\frac{1}{N}\sum_{i=1}^{N}(u^\top\zeta_i)^2\mathbbm{1}(|\nu_i|\le \tau)\right\}+\lambda_{\min}(\lambda\Sigma^{-1}),\label{eq.tmp_8_5}
\end{align}
where the second step is due to (\ref{eq.loss_deriv_2}), the third step comes from (\ref{eq.ell_deriv_2}) and the definition $\zeta_i=\Sigma^{-1/2}\mathrm{Z}_i$ at the beginning of Step 1.1, the fourth step is due to Lemma \ref{lemma.matrix} \ref{lemma.matrix_3} and the assumpion that {$\ell(\cdot)$ is convex in Theorem \ref{thm.robust_main_dro}}, and the last step holds by Lemma \ref{lemma.matrix} \ref{lemma.matrix_1}. For any ${\theta_{\text{reg}}}$ satisfying $\left\|\Sigma^{1/2}\left(\theta_{\text{reg}}-\theta^*_{\text{reg}}\right)\right\|_2\le\mathscr{m}^*$, we have that
\begin{align}
    |\nu_i|=&|\widetilde{\mathrm{Y}}_i-\langle \mathrm{Z}_i,\theta_{\text{reg}}\rangle|
    =|\langle \mathrm{Z}_i,\theta^*_{\text{reg}}\rangle+\nu^*_i-\langle \mathrm{Z}_i,\theta_{\text{reg}}\rangle|\notag\\
    \le& |\mathrm{Z}_i^\top(\theta^*_{\text{reg}}-\theta_{\text{reg}})|+|\nu^*_i|=|\zeta_i^\top\Sigma^{-1/2}(\theta^*_{\text{reg}}-\theta_{\text{reg}})|+|\nu^*_i|\notag\\
    \le&\|\zeta_i\|_2\|\Sigma^{-1/2}(\theta^*_{\text{reg}}-\theta_{\text{reg}})\|_2+|\nu^*_i|\le \mathscr{m}^*\|\zeta_i\|_2+|\nu^*_i|,\label{eq.tmp_8_6}
\end{align}
where $\nu_i$ is defined after (\ref{eq.loss_proof}), and $\nu^*_i=\varepsilon'+\varsigma(\mathrm{X},\mathrm{Y})\varepsilon''$, as defined at the beginning of Step 1.1. Therefore, by (\ref{eq.tmp_8_5}) and (\ref{eq.tmp_8_6}), we further obtain that 
\begin{align}\label{eq.tmp_9}
    &\lambda_{\min}\left(\Sigma^{-1/2}\nabla^2\mathcal{L}(\widetilde{\theta}^{\tau,\lambda}_{\text{reg}}(\eta))\Sigma^{-1/2}\right)\notag\\
    \ge& \inf_{u\in\mathbb{S}^{d-1}} \left\{\frac{1}{N}\sum_{i=1}^{N}(u^\top\zeta_i)^2\mathbbm{1}(\mathscr{m}^*\|\zeta_i\|_2+|\nu^*_i|\le \tau)\right\}+\lambda_{\min}(\lambda\Sigma^{-1})\notag\\
    \ge& \inf_{u\in\mathbb{S}^{d-1}} \left[\frac{1}{N}\sum_{i=1}^{N}(u^\top\zeta_i)^2\mathbbm{1}(\mathscr{m}^*\|\zeta_i\|_2+|\nu^*_i|\le \tau)-\frac{1}{N}\sum_{i=1}^{N}\mathbb{E}\{(u^\top\zeta_i)^2\mathbbm{1}(\mathscr{m}^*\|\zeta_i\|_2+|\nu^*_i|\le \tau)\}\right]\notag\\
    &+\inf_{u\in\mathbb{S}^{d-1}}\left[\frac{1}{N}\sum_{i=1}^{N}\mathbb{E}\{(u^\top\zeta_i)^2\mathbbm{1}(\mathscr{m}^*\|\zeta_i\|_2+|\nu^*_i|\le \tau)\}\right]+\lambda_{\min}(\lambda\Sigma^{-1})\notag\\
    =&\lambda_{\min}\left(\frac{1}{N}\sum_{i=1}^{N}\zeta_i\zeta^\top_i\mathbbm{1}(\mathscr{m}^*\|\zeta_i\|_2+|\nu^*_i|\le \tau)-\frac{1}{N}\sum_{i=1}^{N}\mathbb{E}\{\zeta_i\zeta^\top_i\mathbbm{1}(\mathscr{m}^*\|\zeta_i\|_2+|\nu^*_i|\le \tau)\}\right)\notag\\
    &+\inf_{u\in\mathbb{S}^{d-1}}\left[\frac{1}{N}\sum_{i=1}^{N}\mathbb{E}\{(u^\top\zeta_i)^2\mathbbm{1}(\mathscr{m}^*\|\zeta_i\|_2+|\nu^*_i|\le \tau)\}\right]+\lambda_{\min}(\lambda\Sigma^{-1})\notag\\
    \ge&\inf_{u\in\mathbb{S}^{d-1}}\underbrace{\left[\frac{1}{N}\sum_{i=1}^{N}\mathbb{E}\{(u^\top\zeta_i)^2\mathbbm{1}(\mathscr{m}^*\|\zeta_i\|_2+|\nu^*_i|\le \tau)\}\right]}_{\text{(III)-a}}+\lambda_{\min}(\lambda\Sigma^{-1})\notag\\
    &-\underbrace{\left\|\frac{1}{N}\sum_{i=1}^{N}\zeta_i\zeta^\top_i\mathbbm{1}(\mathscr{m}^*\|\zeta_i\|_2+|\nu^*_i|\le \tau)-\frac{1}{N}\sum_{i=1}^{N}\mathbb{E}\{\zeta_i\zeta^\top_i\mathbbm{1}(\mathscr{m}^*\|\zeta_i\|_2+|\nu^*_i|\le \tau)\}\right\|_2}_{\text{(III)-b}},
\end{align}
where the third step is due to Lemma \ref{lemma.matrix} \ref{lemma.matrix_1}, and last step is due to Lemma \ref{lemma.matrix} \ref{lemma.matrix_2}.

Next, we examine (III)-a. Specifically, for any $u\in\mathbb{S}^{d-1}$,
\begin{align*}
    \text{(III)-a}=& \frac{1}{N}\sum_{i=1}^{N}\mathbb{E}\{(u^\top\zeta_i)^2\mathbbm{1}(\mathscr{m}^*\|\zeta_i\|_2+|\nu^*_i|\le \tau)\}\\
    =&\frac{1}{N}\sum_{i=1}^{N}\mathbb{E}\{(u^\top\zeta_i)^2\} - \frac{1}{N}\sum_{i=1}^{N}\mathbb{E}\{(u^\top\zeta_i)^2\mathbbm{1}(\mathscr{m}^*\|\zeta_i\|_2+|\nu^*_i|> \tau)\}\\
    =&\frac{1}{N}\sum_{i=1}^{N}u^\top\Sigma^{-1/2}\mathbb{E}\{Z_i Z_i^\top\}\Sigma^{-1/2} u - \frac{1}{N}\sum_{i=1}^{N}\mathbb{E}\{(u^\top\zeta_i)^2\mathbbm{1}(\mathscr{m}^*\|\zeta_i\|_2+|\nu^*_i|> \tau)\}\\
    =&1-\frac{1}{N}\sum_{i=1}^{N}\mathbb{E}\{(u^\top\zeta_i)^2\mathbbm{1}(\mathscr{m}^*\|\zeta_i\|_2+|\nu^*_i|> \tau)\}\\
    \ge&1-\frac{1}{\tau N}\sum_{i=1}^{N}\mathbb{E}\{(u^\top\zeta_i)^2(\mathscr{m}^*\|\zeta_i\|_2+|\nu^*_i|)\}\\
    =&1-\frac{1}{\tau N}\sum_{i=1}^{N}\left[\mathscr{m}^*\mathbb{E}\{(u^\top\zeta_i)^2\|\zeta_i\|_2\}+\mathbb{E}\{(u^\top\zeta_i)^2|\nu^*_i|\}\right]\\
    \ge&1-\frac{1}{\tau N}\sum_{i=1}^{N}\left[\mathscr{m}^*\sqrt{\mathbb{E}\{(u^\top\zeta_i)^4\}}\sqrt{\mathbb{E}(\zeta_i^\top \zeta_i)}+\sqrt{\mathbb{E}\{(u^\top\zeta_i)^4\}}\sqrt{\mathbb{E}(\nu^{*2}_i)}\right]\\
    \ge& 1-\frac{1}{\tau N}\sum_{i=1}^{N}4\alpha^2_1\left[\mathscr{m}^*\sqrt{\mathbb{E}(\zeta_i^\top \zeta_i)}+\sigma\right],\\
\end{align*}
where the fourth step is due to the definition of $\Sigma$ given after (\ref{eq.tmp_2}) and the fact that $u\in\mathbb{S}^{d-1}$, the fifth step is due to the fact that $\mathbbm{1}(a>b)\le\frac{a}{b}$ for any $a>0$ and $b>0$, the penultimate step is due to the Cauchy–Schwarz inequality, and the last step comes from Condition \ref{appdx.regularity_condition} \ref{regu.3} and the Lemma \ref{lemma.sub_gaussian_property} \ref{lemma.sub_gaussian_property_2}. Since
\begin{align*}
    \mathbb{E}\{\zeta_i^\top \zeta_i\}
    =&\mathbb{E}(Z_i^\top\Sigma^{-1}Z_i)=\mathbb{E}\{\text{tr}(Z_i^\top\Sigma^{-1}Z_i)\}=\mathbb{E}\{\text{tr}(\Sigma^{-1}Z_iZ_i^\top)\}\notag\\
    =&\text{tr}\{\mathbb{E}(\Sigma^{-1}Z_iZ_i^\top)\}=\text{tr}(I_{d\times d})=d,
\end{align*}
we further obatain that 
\begin{align}\label{eq.tmp_10}
    \text{(III)-a}\ge1-4\alpha^2_1\tau^{-1}(\mathscr{m}^*\sqrt{d}+\sigma).
\end{align}

Now we establish a lower bound for (III)-b in (\ref{eq.tmp_9}) using Lemma \ref{lemma.matrix_subgaussian_row}. Specifically, write
\begin{align*}
    \frac{1}{N}\sum_{i=1}^{N}\zeta_i\zeta^\top_i\mathbbm{1}(\mathscr{m}^*\|\zeta_i\|_2+|\nu^*_i|\le \tau)\triangleq \frac{1}{N} A^\top A,
\end{align*}
where $A=(A_1,\ldots,A_N)^\top$ with $A_i=\zeta_i \mathbbm{1}(\mathscr{m}^*\|\zeta_i\|_2+|\nu^*_i|\le \tau)$ for $i\in[N]$. By Condition \ref{appdx.regularity_condition} \ref{regu.3} and Lemma \ref{lemma.matrix_subgaussian_row}, we obtain that with probability at least $1-2e^{-s}$,
\begin{align}\label{eq.tmp_11}
    \text{(III)-b}\le \max(\epsilon,\epsilon^2),\ \text{with}\ \epsilon=\mathcal{K}_4\sqrt{\frac{d+s}{N}}
\end{align}
for some constant $\mathcal{K}_4>0$ that depends only on $\alpha_2$ in Condition \ref{appdx.regularity_condition} \ref{regu.3}.

Combining (\ref{eq.tmp_9}), (\ref{eq.tmp_10}), and (\ref{eq.tmp_11}), we obtain that with probability at least $1-2e^{-s}$,
\begin{align}\label{eq.tmp_12}
    \lambda_{\min}\left(\Sigma^{-1/2}\nabla^2\mathcal{L}(\widetilde{\theta}^{\tau,\lambda}_{\text{reg}}(\eta))\Sigma^{-1/2}\right)\ge 1-4\alpha^2_1\tau^{-1}(\mathscr{m}^*\sqrt{d}+\sigma)-\max(\epsilon,\epsilon^2)\ge\frac{1}{4},
\end{align}
where the last inequality holds if {$\tau\ge 8\alpha^2_1(\mathscr{m}^*\sqrt{d}+\sigma)$ and $N\ge 16\mathcal{K}^2_4(d+s)$}, which is guaranteed by {taking $\tau\ge\mathscr{C}_1(\sqrt{d}+\sigma)$ and $N\ge\mathscr{C}_2(d+s)$ as required by Theorem \ref{thm.robust_main_dro}}.

Combining (\ref{eq.tmp_3_1}), (\ref{eq.tmp_8_3}), and (\ref{eq.tmp_12}), we obtain that with probability at least $1-5e^{-s}$,
\begin{align*}
    \left\|\Sigma^{1/2}\left\{\widehat{\theta}^{\tau,\lambda}_{\text{reg}}(\eta)-\theta^*_{\text{reg}}\right\}\right\|_2\le 4\left\{\frac{\sigma}{\tau} \varpi^{1/2}+\lambda \|\Sigma^{-1/2}\theta^*_{\text{reg}}\|_2+\sqrt{\frac{\mathcal{K}^\dagger (2.2d+6s)}{N}}\right\}<\mathscr{m}^*,
\end{align*}
if we take, for example, $\mathscr{m}^*=4.1\left\{\frac{\sigma}{\tau} \varpi^{1/2}+\lambda \|\Sigma^{-1/2}\theta^*_{\text{reg}}\|_2+\sqrt{\frac{\mathcal{K}^\dagger (2.2d+6s)}{N}}\right\}$. Consequently, $\widehat{\theta}^{\tau,\lambda}_{\text{reg}}(\eta)$ is an interior point of the ball with center $\theta^*_{\text{reg}}$ and radius $\mathscr{m}^*$, and thus, by the construction of $\widehat{\theta}^{\tau,\lambda}_{\text{reg}}(\eta)$ at the beginning of Step 2, we can take $\eta=1$. Therefore, under the assumptions of Theorem \ref{thm.robust_main_dro}, with probability at least $1-5e^{-s}$, 
\begin{align*}
    \left\|\widehat{\theta}^{\tau,\lambda}_{\text{reg}}-\theta^*_{\text{reg}}\right\|_{\Sigma}\triangleq\left\|\Sigma^{1/2}\left\{\widehat{\theta}^{\tau,\lambda}_{\text{reg}}(\eta)-\theta^*_{\text{reg}}\right\}\right\|_2\le \mathscr{C}^\dagger\sqrt{\frac{d+s}{N}}
\end{align*}
for some constant $\mathscr{C}^\dagger>0$. Hence, the proof of Theorem \ref{thm.robust_main_dro} \ref{thm.robust_main_dro_1} is completed.

\subsection{Proof of Theorem \ref{thm.robust_main} \ref{thm.robust_main_2} (Theorem \ref{thm.robust_main_dro} \ref{thm.robust_main_dro_2})}

The prediction error, defined by $\text{err}\triangleq\left|\langle \mathrm{Z}_{\text{new}},\widehat{\theta}^{\tau,\lambda}_{\text{reg}}\rangle-\langle\mathrm{Z}_{\text{new}},\theta^*_{\text{reg}}\rangle\right|$, can be upper bounded as below:
\begin{align*}
    \text{err}
    =&\left|(\Sigma^{-1/2}\mathrm{Z}_{\text{new}})^\top\Sigma^{1/2}(\widehat{\theta}^{\tau,\lambda}_{\text{reg}}-\theta^*_{\text{reg}})\right|\\
    =&\left|\langle\Sigma^{-1/2}\mathrm{Z}_{\text{new}},\frac{\Sigma^{1/2}(\widehat{\theta}^{\tau,\lambda}_{\text{reg}}-\theta^*_{\text{reg}})}{\|\widehat{\theta}^{\tau,\lambda}_{\text{reg}}-\theta^*_{\text{reg}}\|_{\Sigma}}\rangle\right|\cdot\|\widehat{\theta}^{\tau,\lambda}_{\text{reg}}-\theta^*_{\text{reg}}\|_{\Sigma}\\
    \le&\sup_{u\in\mathbb{S}^{d-1}}|\langle \Sigma^{-1/2}\mathrm{Z}_{\text{new}},u\rangle|\cdot \|\widehat{\theta}^{\tau,\lambda}_{\text{reg}}-\theta^*_{\text{reg}}\|_{\Sigma}.
\end{align*}
By Condition \ref{appdx.regularity_condition} \ref{regu.3} and Lemma \ref{lemma.sub_gaussian_property} \ref{lemma.sub_gaussian_property_2}, for any $s>0$, with probability at least $1-2e^{-s}$,
\begin{align*}
    \sup_{u\in\mathbb{S}^{d-1}}|\langle \Sigma^{-1/2}\mathrm{Z}_{\text{new}},u\rangle|\le \mathcal{K}_5 \sqrt{s}
\end{align*}
for some constant $\mathcal{K}_5>0$, which depends only on $\alpha_2$ in Condition \ref{appdx.regularity_condition} \ref{regu.3}. Consequently, combining with Theorem \ref{thm.robust_main_dro} \ref{thm.robust_main_dro_1}, we obtain that, with probability at least $1-7e^{-s}$,
\begin{align*}
    \text{err}\le \mathscr{C}^{\ddagger}\sqrt{\frac{ds+s^2}{N}}
\end{align*}
for some constant $\mathscr{C}^{\ddagger}>0$, which completes the proof of Theorem \ref{thm.robust_main_dro} \ref{thm.robust_main_dro_2}.

\subsection{Proof of Proposition \ref{prop:infimal_convolution}}

\begin{proof}
We verify that $\ell_\tau$ satisfies the conditions in
Lemma~\ref{lem:quad_inf_conv}. By Lemma~\ref{lemma.convex}, $\ell_\tau$ is
convex. Then it suffices to verify that $\ell_\tau'$ is $1$-Lipschitz.
We consider the following three cases.
\begin{itemize}
    \item[(i)] For $0\le s\le t\le \tau$, this is immediate since
    $\ell_\tau'(t)-\ell_\tau'(s)=t-s$.

    \item[(ii)] For $\tau\le s\le t$, since $\ell'(t)/t$ is
    non-increasing and $s\le t$, we have
    \[
    \frac{\ell'(t)}{t}\le \frac{\ell'(s)}{s},
    \]
    which rearranges to
    \[
    \ell'(t)-\ell'(s)\le \frac{t-s}{s}\ell'(s).
    \]
    Since $\ell'(t)/t$ is non-increasing and $s\ge\tau$, we also have
    $\ell'(s)/s\le\ell'(\tau)/\tau$, so
    $\ell'(s)\le s\ell'(\tau)/\tau$. Therefore,
    \[
    \ell'(t)-\ell'(s)
    \le\frac{t-s}{s}\cdot\frac{s\,\ell'(\tau)}{\tau}
    =\frac{(t-s)\ell'(\tau)}{\tau}.
    \]
    Multiplying both sides by $\tau/\ell'(\tau)$ gives
    \[
    \frac{\tau}{\ell'(\tau)}
    \{\ell'(t)-\ell'(s)\}\le t-s,
    \]
    i.e., $\ell_\tau'(t)-\ell_\tau'(s)\le t-s$.

    \item[(iii)] For the cross-region case $0\le s\le\tau\le t$,
    \[
    \ell_\tau'(t)-\ell_\tau'(s)
    =\bigl\{\ell_\tau'(t)-\ell_\tau'(\tau)\bigr\}
    +\{\ell_\tau'(\tau)-\ell_\tau'(s)\}
    \le(t-\tau)+(\tau-s)=t-s.
    \]
    where the inequality holds by applying the upper bounds
    in cases (i) and (ii).
\end{itemize}

Since $\ell_\tau$ is even, the same Lipschitz bound extends to the
whole real line by symmetry. Thus $\ell_\tau$ is convex and has
$1$-Lipschitz derivative. Therefore, by
Lemma~\ref{lem:quad_inf_conv}, we obtain
\begin{align}\label{eq.inf_conv}
    \ell_\tau(\nu)
    =
    \inf_{\xi\in\mathbb{R}}
    \left\{
    \frac{1}{2}\xi^2+\phi_\tau(\nu-\xi)
    \right\},
\end{align}
where $\phi_\tau = \psi^*_\tau$, with
$\psi_{\tau}(w)\triangleq \ell_\tau^*(w)-\frac{1}{2}w^2$.
This proves the infimal convolution representation.
It remains to derive the parametric form of $\phi_\tau$.

Fix $t\ge\tau$ and set
\begin{align}\label{eq.def_w_u}
    w=\ell_\tau'(t)=\frac{\tau}{\ell'(\tau)}\ell'(t),
    \qquad
    u=t-w=t-\frac{\tau\ell'(t)}{\ell'(\tau)}.
\end{align}
Here we restrict to $t\ge\tau$, since for
{$0\le t\le\tau$},
$\ell_\tau(t)=t^2/2$ and hence $\ell_\tau'(t)=t$,
which implies $u=t-\ell_\tau'(t)=0$.
Therefore, all nontrivial values $u>0$ arise from the regime
$t\ge\tau$.

By convexity and differentiability of $\ell_\tau$, we have
$w\in\partial\ell_\tau(t)$, so the Fenchel--Young equality gives
\begin{align}\label{eq.fenchel_eq_w_t}
    \ell_\tau^*(w)=tw-\ell_\tau(t).
\end{align}

Next, since
$\phi_\tau^*(w)=\ell_\tau^*(w)-\tfrac{1}{2}w^2$, we obtain
\begin{align}
    \partial\phi_\tau^*(w)
    =
    \partial\ell_\tau^*(w)-w.
\end{align}
{Since $w\in\partial\ell_\tau(t)$ by
(\ref{eq.def_w_u}), we have
$t\in\partial\ell_\tau^*(w)$ by conjugacy,
and hence
\[
u=t-w\in\partial\ell_\tau^*(w)-w
=\partial\phi_\tau^*(w).
\]
Another application of the Fenchel--Young
equality yields
\begin{align}\label{eq.fenchel_eq_u_w}
    \phi_\tau(u)+\phi_\tau^*(w)=uw.
\end{align}}

% Since $w\in\partial\ell_\tau(t)$ by (\ref{eq.def_w_u}),
% we have $t\in\partial\ell_\tau^*(w)$ by conjugacy,
% {\color{blue}
% and hence
% \[
% u=t-w\in\partial\ell_\tau^*(w)-w
% =\partial\phi_\tau^*(w),
% \]
% where the inclusion follows from the definition of $u$
% in \eqref{eq.def_w_u}.
% }
% Thus $u\in\partial\phi_\tau^*(w)$, and another application
% of the Fenchel--Young equality yields
% \begin{align}\label{eq.fenchel_eq_u_w}
%     \phi_\tau(u)+\phi_\tau^*(w)=uw.
% \end{align}

Therefore,
\begin{align}\label{eq.phi_tau_u}
    \phi_\tau(u)
    &= uw-\phi_\tau^*(w)\\
    &= uw-\ell_\tau^*(w)+\frac{1}{2}w^2\\
    &= uw-\{tw-\ell_\tau(t)\}+\frac{1}{2}w^2\\
    &= \ell_\tau(t)-\frac{1}{2}w^2,
\end{align}
where the first step follows from \eqref{eq.fenchel_eq_u_w},
the second step comes from the definition of $\phi_\tau^*$,
and the third step is due to \eqref{eq.fenchel_eq_w_t}.

Substituting
$\ell_\tau(t)=\tfrac{\tau}{\ell'(\tau)}
\{\ell(t)-\ell(\tau)\}+\tfrac{\tau^2}{2}$
and $w=\tfrac{\tau\ell'(t)}{\ell'(\tau)}$
into \eqref{eq.phi_tau_u} for $t\ge\tau$ yields
\[
\phi_\tau(u)
=
\frac{\tau}{\ell'(\tau)}
\bigl\{\ell(t)-\ell(\tau)\bigr\}
+\frac{\tau^2}{2}
-\frac{1}{2}
\left\{\frac{\tau\ell'(t)}{\ell'(\tau)}\right\}^2.
\]

{
It remains to show that the mapping $t\mapsto u(t)$ is
injective and covers all $u\ge 0$. Define
\[
u(t)=t-\frac{\tau}{\ell'(\tau)}\ell'(t),
\qquad t\ge\tau.
\]
Let $r(t)\triangleq\ell'(t)/t$ and
$r_\tau\triangleq r(\tau)=\ell'(\tau)/\tau>0$.
Then
\[
u(t)=t\left\{1-\frac{r(t)}{r_\tau}\right\}.
\]
By assumption, $r(t)$ is strictly decreasing on
$[\tau,\infty)$, and hence $u(\tau)=0$ and
$u(t)>0$ for all $t>\tau$.

For any $\tau\le s<t$, we have
$r(t)<r(s)\le r_\tau$. Therefore,
\[
\begin{aligned}
\ell'(t)-\ell'(s)
=tr(t)-sr(s)<(t-s)r(s)\le(t-s)r_\tau.
\end{aligned}
\]
Consequently,
\[
\begin{aligned}
u(t)-u(s)
=(t-s)-\frac{\ell'(t)-\ell'(s)}{r_\tau}>0.
\end{aligned}
\]
Thus $u(t)$ is strictly increasing on
$[\tau,\infty)$ and hence injective.

To establish surjectivity, fix any $t_0>\tau$.
Since $r(t)$ is strictly decreasing,
$r(t_0)<r_\tau$. For all $t\ge t_0$,
\[
\begin{aligned}
u(t)
=t\left\{1-\frac{r(t)}{r_\tau}\right\}\ge
t\left\{1-\frac{r(t_0)}{r_\tau}\right\}.
\end{aligned}
\]
Since $1-r(t_0)/r_\tau>0$, it follows that
$u(t)\to\infty$ as $t\to\infty$.
Together with the continuity of $u(t)$ and
$u(\tau)=0$, this shows that $u(t)$ maps
$[\tau,\infty)$ onto $[0,\infty)$.

Therefore, $t\mapsto u(t)$ is a bijection from
$[\tau,\infty)$ onto $[0,\infty)$, ensuring that
the parametric characterization of $\phi_\tau(u)$
is well-defined for every $u\ge 0$.
}

Finally, since $\ell_\tau$ is even, for any $w$ we have
\[
\ell_\tau^*(-w)
=
\sup_{x}\{-wx-\ell_\tau(x)\}
=
\sup_{x}\{w(-x)-\ell_\tau(-x)\}
=
\sup_{x}\{wx-\ell_\tau(x)\}
=
\ell_\tau^*(w).
\]
Thus, $\ell_\tau^*$ is even. Hence
$\phi_\tau^*(w)=\ell_\tau^*(w)-w^2/2$ is even,
and therefore $\phi_\tau=(\phi_\tau^*)^*$ is also even,
giving $\phi_\tau(-u)=\phi_\tau(u)$.
At $t=\tau$, we have $u=0$ and thus
$\phi_\tau(0)=0$. This completes the proof.
\end{proof}

%%%%%%%%%%%%%%%%%%%%%%%%%%%%%%%%%%%%%%%%%%%%%%%%%%%%%%%%%%%%%%%%%%%%%%%%%%%%%%%

% \section{Extended Related Work}

% {\color{red}(START HERE)}

%%%%%%%%%%%%%%%%%%%%%%%%%%%%%%%%%%%%%%%%%%%%%%%%%%%%%%%%%%%%%%%%%%%%%%%%%%%%%%%

\section{Experiment Details}\label{appdx.exp_details}

\subsection{Experiment Setup}\label{appdx.exp_detail}

\paragraph{Metric Computation.} 
Let $y_i$ and $\hat{y}_i$ denote the ground-truth and predicted labels for the $i$-th sample, and let $N$ be the total number of test samples. We employ the following three commonly used evaluation metrics to assess the performance for regression tasks: 
$\text{MAE} = \frac{1}{N}\sum_{i=1}^N |\hat{y}_i - y_i|$; 
$\text{RMSE} = \sqrt{\frac{1}{N}\sum_{i=1}^N (\hat{y}_i - y_i)^2}$; 
% $\text{MAPE} = \frac{1}{N}\sum_{i=1}^N \frac{|\hat{y}_i - y_i|}{|y_i|}$; 
% $R = \frac{\sum_{i=1}^N (\hat{y}_i - \bar{\hat{y}})(y_i - \bar{y})}
% {\sqrt{\sum_{i=1}^N (\hat{y}_i - \bar{\hat{y}})^2}\sqrt{\sum_{i=1}^N (y_i - \bar{y})^2}}$; 
and $R^2 = 1 - \frac{\sum_{i=1}^N (\hat{y}_i - y_i)^2}{\sum_{i=1}^N (y_i - \bar{y})^2}$.
Here, $\bar{y}=\frac{1}{N}\sum_{i=1}^{N}y_i$ and $\bar{\hat{y}}=\frac{1}{N}\sum_{i=1}^{N}\hat{y}_i$ denote the sample means of the ground-truth and predicted labels, respectively. MAE and RMSE measure prediction accuracy from different perspectives, and $R^2$ characterizes the goodness-of-fit between predictions and ground-truth values.

\paragraph{Dataset information.} We evaluate our method on four widely used real-world regression benchmarks spanning facial analysis, human pose estimation, and medical imaging. 
\begin{itemize}
    \item \textbf{UTKFace} \cite{zhifei2017cvpr} is a large-scale facial dataset containing over $N=20,000$ images annotated with age, gender, and ethnicity. In our experiments, face images are resized to $p=3\times224\times224$ and used as inputs $\mathrm{X}$, age annotations serve as the regression targets $\mathrm{Y}$, and synthetic Gaussian noise is added to generate noisy labels $\widetilde{\mathrm{Y}}$. A ResNet-50 backbone is used as the feature extractor, producing $d=2048$-dimensional feature representations.
    
    \item \textbf{Biwi Kinect} \cite{fanelli2013random} is a head pose estimation dataset consisting of RGB-D images captured with a Microsoft Kinect sensor, annotated with continuous head pose angles including yaw, pitch, and roll. The dataset contains approximately $N=15,000$ images from 20 subjects (6 females and 14 males), where four subjects were recorded twice to increase pose variability and robustness. In our experiments, RGB images are resized to $p=3\times224\times224$ and used as inputs $\mathrm{X}$, the head pose angles serve as the regression targets $\mathrm{Y}$, and synthetic Gaussian noise is added to generate noisy labels $\widetilde{\mathrm{Y}}$. A ResNet-50 backbone is used as the feature extractor, producing $d=2048$-dimensional feature representations.
    
    \item \textbf{APPA-REAL} \cite{clapes2018apparent} contains 7,591 facial images annotated with both real age and apparent age, {where apparent age refers to the age perceived by human annotators rather than the chronological age of the subject}. The dataset is split into 4,113 training, 1,500 validation, and 1,978 test images. {In our experiments, face images are resized to $p=3\times224\times224$ and used as inputs $\mathrm{X}$, apparent age annotations are treated as noisy labels $\widetilde{\mathrm{Y}}$, and real age labels are used as the ground-truth targets $\mathrm{Y}$ for evaluation. A ResNet-50 backbone is used as the feature extractor, producing $d=2048$-dimensional feature representations.}

    \item For \textbf{Bone-Age} assessment, we combine two datasets: \textbf{RSNA} \cite{halabi2019rsna} and \textbf{DHA} (Digital Hand Atlas) \cite{pietka2001computer}. The RSNA dataset contains 12,611 training and 1,425 validation hand X-ray images with manually annotated bone ages, and is used for model training. The DHA dataset includes 1,400 left hand and wrist radiographs of children, accompanied by demographic information and radiologists’ assessments based on visual comparison with an atlas and computer-aided diagnosis (CAD). {In our experiments, hand X-ray images, resized to $p=224\times224$, are used as inputs $\mathrm{X}$, bone age annotations from RSNA are treated as noisy labels $\widetilde{\mathrm{Y}}$ for model training, and radiologist-assessed bone ages from DHA are used as ground-truth targets $\mathrm{Y}$ for evaluation.} A DenseNet-121 backbone is used as the feature extractor, producing $d=1024$-dimensional feature representations.
    
\end{itemize}

\paragraph{Noise Simulation.} For UTKFace and Biwi Kinect, we generate synthetic Gaussian label noise using the following procedures:
\begin{enumerate}
    
    \item Let $\mathcal{D}_{\mathrm{y}}\triangleq\{{\mathrm{Y}}_i\}_{i=1}^{N}$ denote the set of clean labels, and let $\mathrm{y}_{\min}$ and $\mathrm{y}_{\max}$ be the minimum and maximum label values in the dataset.

    \item Uniformly partition the interval $[y_{\min}, y_{\max}]$ into $M$ disjoint sub-intervals. In all experiments, we set $M = 4000$, and let $\text{std}_{\max} \in \{30, 60, 90\}$ to control the maximum noise magnitude.

    \item For each sample, uniformly draw an integer variance $\sigma^2$ from $[\text{std}_{\max}]$ and then sample a noise offset from $\mathcal{N}(0, \sigma^2)$. The resulting offset determines an interval index. Let $[y_l, y_u]$ denote the corresponding sub-interval, and define $\Omega \triangleq \mathcal{D}_y \cap [y_l, y_u]$. Noisy label $\widetilde{Y}_i$ is taken as a value drawn from the uniform distribution over $\Omega$.

\end{enumerate}

This noise generation strategy exhibits several desirable properties.
First, by sampling noise magnitudes at the instance level, it introduces heteroscedastic and {sample-varying} perturbations, which more closely resemble real-world annotation noise than i.i.d. Gaussian corruption.
Second, sampling noisy labels from nearby intervals preserves the ordered and metrized structure of the continuous label space, avoiding unrealistic label flips that violate semantic proximity.
Finally, the use of $\text{std}_{\max}$ allows precise control over noise severity while yielding nontrivial and nonuniform noise distributions. Table~\ref{tab:noise_mae} summarizes the average MAE of the generated noisy labels over five random seeds under different values of $\text{std}_{\max}$.

\begin{table}[h]
\centering
\caption{Average MAE of generated noisy labels under different noise levels ($\text{std}_{\max}$).}
\label{tab:noise_mae}
\setlength{\tabcolsep}{8pt}
\resizebox{0.60\textwidth}{!}{
\begin{tabular}{lccc}
\toprule
\textbf{Dataset} & $\textbf{std}_{\max}=30$ & $\textbf{std}_{\max}=60$ & $\textbf{std}_{\max}=90$ \\
\midrule
UTKFace (Age)           & 7.04  & 10.52 & 11.44 \\
% \midrule
Biwi Kinect (Yaw)       & 11.11 & 19.24 & 24.07 \\
Biwi Kinect (Pitch)     & 10.98 & 18.24 & 22.16 \\
Biwi Kinect (Roll)      & 9.64  & 13.61 & 15.53 \\
\bottomrule
\end{tabular}}
\end{table}

\paragraph{Baseline Methods.}
We compare GRAIN with a diverse set of regression and noisy-label learning baselines, including both methods originally proposed for regression and classical approaches from noisy-label classification adapted to regression settings.

\begin{itemize}
    
    \item {\textbf{Existing Regression Methods.}}
    
    \begin{itemize}
        \item \textbf{Vanilla (MSE).}
        This baseline directly trains a regression network using the mean squared error (MSE) loss on noisy labels.
    
        \item \textbf{SuperLoss}~\cite{castells2020superloss}.  
        SuperLoss is a self-paced robust loss function that dynamically reweights training samples according to their loss values, automatically downweighting hard samples with large losses during optimization.
        
        \item \textbf{RDI and AUX}~\cite{husimple}.  
        RDI and AUX are robust regression methods designed to mitigate overfitting to mislabeled data. RDI regularizes learning by constraining the distance between the network parameters and their initialization, while AUX introduces a trainable auxiliary variable for each training example to absorb label noise.
        
        \item \textbf{C-Mixup}~\cite{yao2022c}.  
        C-Mixup extends the Mixup strategy to regression by adjusting the sampling probability based on label similarity, encouraging smoother regression functions under noisy supervision.
    
    \end{itemize}

    \item \textbf{Classification-based noisy-label methods.}  
    We additionally include several classical noisy-label learning methods originally developed for classification, which can be adapted to regression settings:
    \begin{itemize}
        \item \textbf{CDR}~\cite{xia2020robust} divides all parameters into the critical and non-critical ones, and then performs different update rules for different types of parameters.
        
        \item \textbf{CoTeaching}~\cite{han2018co} jointly trains two networks and discards samples with large losses, based on the observation that clean samples tend to have smaller losses in early training stages.
        
        \item \textbf{CNLCU-S and CNLCU-H}~\cite{xiasample} identify clean samples via sample selection guided by uncertainty estimation, with CNLCU-H further incorporating historical information.
        
        \item \textbf{Selfie}~\cite{song2019selfie} adopts a self-training strategy by iteratively refining pseudo labels using model predictions.
        
        \item \textbf{CoSelfie}~\cite{song2019selfie} extends Selfie by combining self-training with co-training to alleviate confirmation bias.
    \end{itemize}

\end{itemize}

Additionally, for all the methods above, we also implement a cross-entropy loss based version by discretizing the continuous target space into ordered bins and introducing a histogram head, following the same binning strategy used in GRAIN. In all experimental results, these adapted variants are denoted by ``BIN''. Unless otherwise specified, all baselines share the same network backbone, training protocol, and model selection strategy as GRAIN to ensure a fair comparison.

\subsection{Additional Implementation Details}\label{appdx.implement_details}

\begin{algorithm}[tb]
\caption{GRAIN Training Procedure}
\label{alg:GRAIN}
\begin{algorithmic}[1]
\STATE \textbf{Input:} Training data $\mathcal{D}=\{(\mathrm{X}_i, \widetilde{\mathrm{Y}}_i)\}_{i=1}^N$, loss weights $\lambda_{\text{reg}}, \lambda_{\text{ce}}, \lambda_{\text{pos}}, \lambda_{\text{neg}}$, warm-up epochs $t_{\text{warm}}$, total epochs $T$
\STATE \textbf{Initialize:} Feature extractor $\theta_{\text{f}}$, regression head $\theta_{\text{reg}}$, histogram head $\theta_{\text{hist}}$, residual head $\theta_{\text{res}}$; {pseudo labels $\{\mathrm{Y}^{*(0)}_i \leftarrow \widetilde{\mathrm{Y}}_i\}_{i=1}^N$; confidences $\{\mathscr{c}^{*(0)}_i \leftarrow 1\}_{i=1}^N$}
\STATE \textbf{Precompute:} Adaptive bin boundaries and representative values $\{\bar{\mathrm{Y}}_k\}_{k=1}^{K}$ via (\ref{eq.bin_width})

\STATE \texttt{\color{blue}// Warm Up}
\FOR{epoch $t = 1$ to $t_{\text{warm}}$}
    \STATE Train $(\theta_{\text{f}}, \theta_{\text{reg}}, \theta_{\text{hist}})$ by minimizing $\lambda_{\text{reg}}\mathcal{L}_{\text{reg}} + \lambda_{\text{ce}}\mathcal{L}_{\text{ce}}$
    \STATE {Update pseudo labels $\{\mathrm{Y}^{*(t)}_i\}_{i=1}^N$ and confidences $\{\mathscr{c}^{*(t)}_i\}_{i=1}^N$ via (\ref{eq.EMA_pseudo_label}) and (\ref{eq.EMA_conf}) if $t \geq \lfloor t_{\text{warm}}/2 \rfloor$}
    \STATE Update $\tau^{(t)}$ via (\ref{eq.tau_EMA}) if $t \geq \lfloor t_{\text{warm}}/2 \rfloor$
\ENDFOR

\FOR{epoch $t = t_{\text{warm}}+1$ to $T$}
    \STATE \texttt{\color{blue}// Pseudo Label and Threshold Update}
    \STATE Update pseudo labels $\{\mathrm{Y}^{*(t)}_i\}_{i=1}^N$ and confidences $\{\mathscr{c}^{*(t)}_i\}_{i=1}^N$ via (\ref{eq.EMA_pseudo_label}) and (\ref{eq.EMA_conf})
    \STATE Update residual capacity threshold $\tau^{(t)}$ via (\ref{eq.tau_EMA})
    \STATE \texttt{\color{blue}// Residual Head Training}
    \STATE Freeze $(\theta_{\text{f}}, \theta_{\text{reg}}, \theta_{\text{hist}})$; update $\theta_{\text{res}}$ by minimizing confidence-weighted residual loss (\ref{eq.residual_train})
    \STATE \texttt{\color{blue}// Main Model Training}
    \STATE Compute confidence-weighted regression loss (\ref{eq.loss_prediction_layer}) and cross-entropy loss (\ref{eq.loss_cls})
    \STATE Construct contrastive triplets using $\{\mathrm{Y}^{*(t)}_i\}$; compute $\mathcal{R}_{\text{pos}}$ and $\mathcal{R}_{\text{neg}}$ via (\ref{eq.loss_feature_pos})--(\ref{eq.loss_feature_neg})
    \STATE Freeze $\theta_{\text{res}}$; update $(\theta_{\text{f}}, \theta_{\text{reg}}, \theta_{\text{hist}})$ by minimizing joint loss (\ref{eq.loss_all})
\ENDFOR
\STATE \textbf{Output:} Trained model parameters $\vartheta = (\theta_{\text{f}}, \theta_{\text{reg}}, \theta_{\text{hist}}, \theta_{\text{res}})$
\end{algorithmic}
\end{algorithm}

\paragraph{Binning procedure (Line 3 of Algorithm \ref{alg:GRAIN}).} Before model training, we adaptively discretize the noisy labels according to their empirical distribution, and keep the number of bins fixed throughout the entire training process. Specifically, {using Python package \texttt{scipy.stats}}, we first perform density estimation on the noisy label set $\{\widetilde{Y}_i\}_{i=1}^N$ using a Gaussian kernel density estimator (KDE). To prevent excessively populated bins from dominating the training dynamics, we determine the bin width based on a local quantile range around the peak of the estimated density. Let $\widehat{F}(\cdot)$ denote the cumulative distribution function (CDF) induced by the KDE, and define the corresponding $100q$th quantile function as $\widehat{Q}(q)\triangleq \inf \left\{y\in\mathcal{Y}:\widehat{F}(y)\ge q\right\}$, where $0<q<1$. Let $y_{\text{peak}}$ denote the location of the density peak and $q_{\text{peak}}\triangleq \widehat{F}(y_{\text{peak}})$ be its associated quantile. We then define the bin width as 
\begin{align}\label{eq.bin_width}
    \mathtt{width}\triangleq \widehat{Q}(q_{\text{peak}}+5\%)-\widehat{Q}(q_{\text{peak}}-5\%),
\end{align}
corresponding to a symmetric 10\% quantile interval around the peak $\mathrm{y}_{\text{peak}}$. This adaptive discretization reflects the underlying label density while mitigating excessive concentration in high-density regions. Given the bin width (\ref{eq.bin_width}), the number of bins is automatically determined as $K\triangleq \lfloor(\mathrm{y}_{\max}-\mathrm{y}_{\min})/\mathtt{width}\rfloor$, where $\mathrm{y}_{\max}\triangleq\max_{1\le i\le N}\widetilde{\mathrm{Y}}_i$ and $\mathrm{y}_{\min}\triangleq\min_{1\le i\le N}\widetilde{\mathrm{Y}}_i$. In the resulting $K$ bins, we let $\Bar{\mathrm{y}}_k$ denote a representative value of bin $k$, which can be chosen as the left endpoint, midpoint, or right endpoint of the corresponding bin interval.

\paragraph{Construction of Robust Pseudo Labels from Histogram Head (Lines {7 and 12} of Algorithm \ref{alg:GRAIN}).} 

We construct robust pseudo labels from the histogram head. Specifically, in the $t$-th epoch, let $\mathdutchcal{p}^{(t)}_i\in\Delta^{K-1}$ denote the softmax output of the histogram head for input $\mathrm{X}_i$. We define $\mathcal{T}^{(t)}_i\subset[K]$ as the minimal set of highest-probability bins whose cumulative probability mass exceeds $68\%$:
\begin{align}\label{eq.68_set}
    \mathcal{T}^{(t)}_i = \arg\min_{\mathcal{S}\subseteq[K]}|\mathcal{S}| \quad \text{s.t.} \quad \sum_{k\in\mathcal{S}} \mathdutchcal{p}^{(t)}_i[k] \geq 0.68,
\end{align}
where bins are added in decreasing order of $\mathdutchcal{p}^{(t)}_i[k]$. We then define the normalized weights as $\mathdutchcal{p}^{(t)*}_i[k]=\frac{\mathdutchcal{p}^{(t)}_i[k]}{\sum_{k'\in\mathcal{T}^{(t)}_i}\mathdutchcal{p}^{(t)}_i[k']}$ for $k\in\mathcal{T}^{(t)}_i$, and $\mathdutchcal{p}^{(t)*}_i[k]=0$ for $k\notin\mathcal{T}^{(t)}_i$. The resulting weighted pseudo label is
\begin{align}\label{eq.pseudo_label_0}
    \breve{\mathrm{Y}}^{(t)*}_i\triangleq\sum_{k=1}^{K}\mathdutchcal{p}^{(t)*}_i[k] \cdot \Bar{\mathrm{y}}_{k},
\end{align}
where $\Bar{\mathrm{y}}_{k}$ is the representative value of bin $k$ from the binning procedure. Alongside the pseudo label, we define a confidence score that reflects the sharpness of the histogram head's prediction:
\begin{align}\label{eq.conf_0}
    \breve{\mathscr{c}}^{(t)}_i\triangleq 1 - \frac{|\mathcal{T}^{(t)}_i|}{K}\in[0,1],
\end{align}
where $|\mathcal{T}^{(t)}_i|$ is the number of selected bins. Intuitively, when the histogram head concentrates its probability mass on fewer bins, the pseudo label is more reliable and $\breve{\mathscr{c}}^{(t)}_i$ is closer to $1$; conversely, when the probability mass is spread across many bins, the pseudo label is less trustworthy and $\breve{\mathscr{c}}^{(t)}_i$ is closer to $0$.

To further stabilize training, we maintain Exponential Moving Averages (EMA) of both the pseudo label and confidence score across epochs. {The pseudo labels and confidence scores are initialized as $\mathrm{Y}^{*(0)}_i = \widetilde{\mathrm{Y}}_i$ and $\mathscr{c}^{*(0)}_i = 1$, respectively, and remain unchanged for $t < \lfloor t_{\text{warm}}/2 \rfloor$. Starting from epoch $t=\lfloor t_{\text{warm}}/2 \rfloor$, both are updated together at each epoch, including during the remaining warm-up epochs, using the following EMA rules:}
\begin{align}
    \mathrm{Y}^{(t)*}_i &\leftarrow \alpha\, \mathrm{Y}^{(t-1)*}_i + (1-\alpha)\, \breve{\mathrm{Y}}^{(t)*}_i,\label{eq.EMA_pseudo_label}\\
    \mathscr{c}^{(t)*}_i &\leftarrow \alpha\, \mathscr{c}^{(t-1)*}_i + (1-\alpha)\, \breve{\mathscr{c}}^{(t)}_i, \label{eq.EMA_conf}
\end{align}
where $\alpha\in(0,1)$ is the EMA decay factor (we use $\alpha=0.9$ in our experiments).
{During training, the per-instance regression and residual losses are weighted by their corresponding pseudo-label confidence $\mathscr{c}_i^*$ before averaging across training instances.}
This construction achieves robustness by concentrating supervision on the most confident bins via the 68\% cumulative threshold, with the confidence score naturally down-weighting pseudo labels when the histogram head spreads probability mass across many bins, and the EMA updates smoothing out per-epoch fluctuations for more stable training.

\paragraph{Residual Capacity $\tau$ Update (Lines {8 and 13} of Algorithm \ref{alg:GRAIN}).}
{Starting at epoch $t=\lfloor t_{\text{warm}}/2\rfloor$,} we update the residual head capacity $\tau$ based on the residual Median Absolute Deviation (MAD)~\cite{rousseeuw1993alternatives}. Let $r^{(t)}_i = |\widetilde{\mathrm{Y}}_i - \mathrm{Y}^{(t)*}_i|$ denote the absolute residual between the noisy label and the current pseudo label for sample $i$. We compute the robust scale estimate
\begin{align*}
    \breve{\tau}^{(t)} \triangleq 1.345 \cdot \text{median}\left(\left\{\Big|r^{(t)}_i - \text{median}\left(\{r^{(t)}_j\}_{j=1}^{N}\right)\Big|\right\}_{i=1}^{N}\right),
\end{align*}
which is a robust and outlier-resistant estimator of the noise scale. The EMA update then follows
\begin{align}\label{eq.tau_EMA}
    \tau^{(t)} \leftarrow \alpha\, \tau^{(t-1)} + (1-\alpha)\, \breve{\tau}^{(t)},
\end{align}
with the same decay factor $\alpha$ as in (\ref{eq.EMA_pseudo_label})--(\ref{eq.EMA_conf}). This update begins at epoch $t = \lfloor t_{\text{warm}} / 2 \rfloor$, {after the pseudo labels and confidence scores have been updated,} at which point $\tau$ is initialized to $\breve{\tau}^{(t)}$.

\paragraph{Hyperparameter Selection.} The implementation of GRAIN involves several hyperparameters. 

\begin{itemize}
    \item[(1)] {\bf Loss weights for the regression and histogram heads.} 
    Observing that the magnitude of the cross-entropy loss varies less than that of the regression loss during training, we fix the classification loss weight $\lambda_{\text{ce}}=1.0$, and gradually ramp up the regression loss weight $\lambda_{\text{reg}}$ from 0 to its maximum value over the first 20 epochs, with the maximum value taken from $\{0.2, 0.5\}$. 
    
    \item[(2)] {\bf Residual head training.} 
    As in the regression head, the parameter $\tau$ is automatically determined during training and does not require manual tuning. 
    
    \item[(3)] {\bf Across-bin contrastive regularization loss.} 
    For the bin-based contrastive regularization, we select $\lambda_{\text{pos}}$ from $\{0.01, 0.05\}$. Instead of tuning $\lambda_{\text{neg}}$ independently, we fix the ratio $\lambda_{\text{neg}}/\lambda_{\text{pos}}=0.5$ throughout all experiments.

\end{itemize}

\paragraph{Computational Infrastructure.}
All experiments were conducted on a single NVIDIA L40S GPU with 48GB 
memory. The average training time per run of GRAIN (across five random 
seeds) is approximately 2.5h on UTKFace, 2.3h on Biwi Kinect, 3.5h 
on Bone-Age, and 0.8h on APPA-REAL. The additional components 
introduced by GRAIN (the residual head, histogram head, and bin-level 
contrastive objective) incur negligible overhead relative to the 
backbone training cost.

\subsection{Extended Results}\label{appdx.exps}

\begin{figure}[H]
    \centering
    \includegraphics[width=0.9\linewidth]{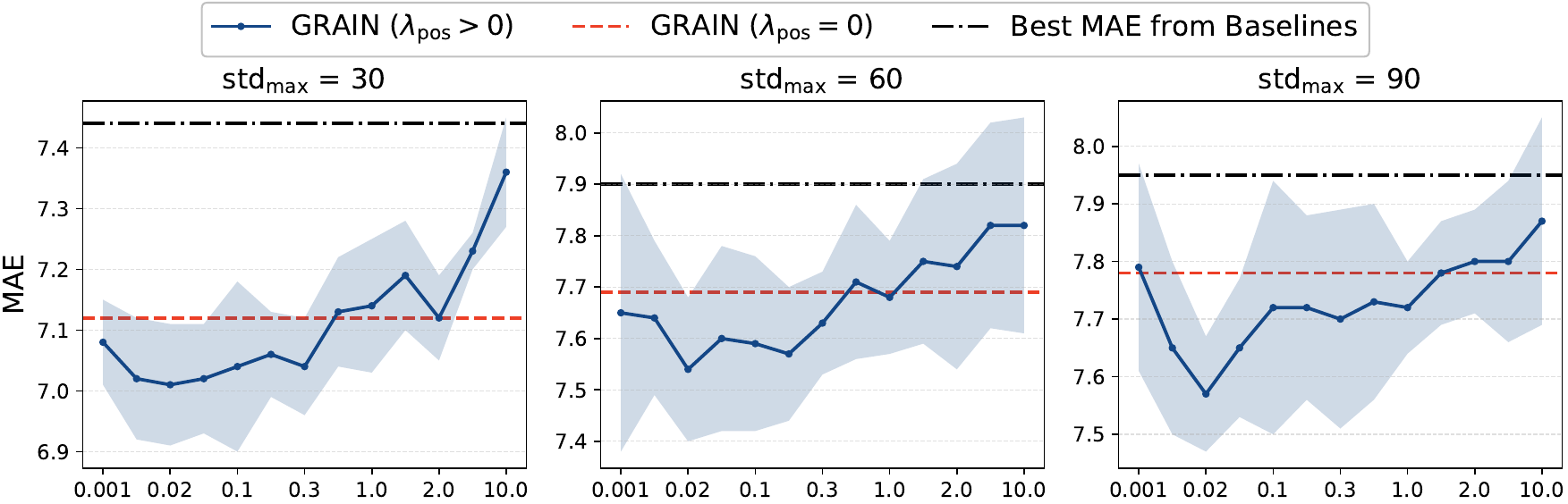}
    \caption{Hyperparameter sensitivity analysis for $\lambda_{\text{pos}}$ on UTKFace.}
    \label{fig:hyper_lam_pos_MAE}
\end{figure}

\begin{figure}[H]
    \centering
    \includegraphics[width=0.9\linewidth]{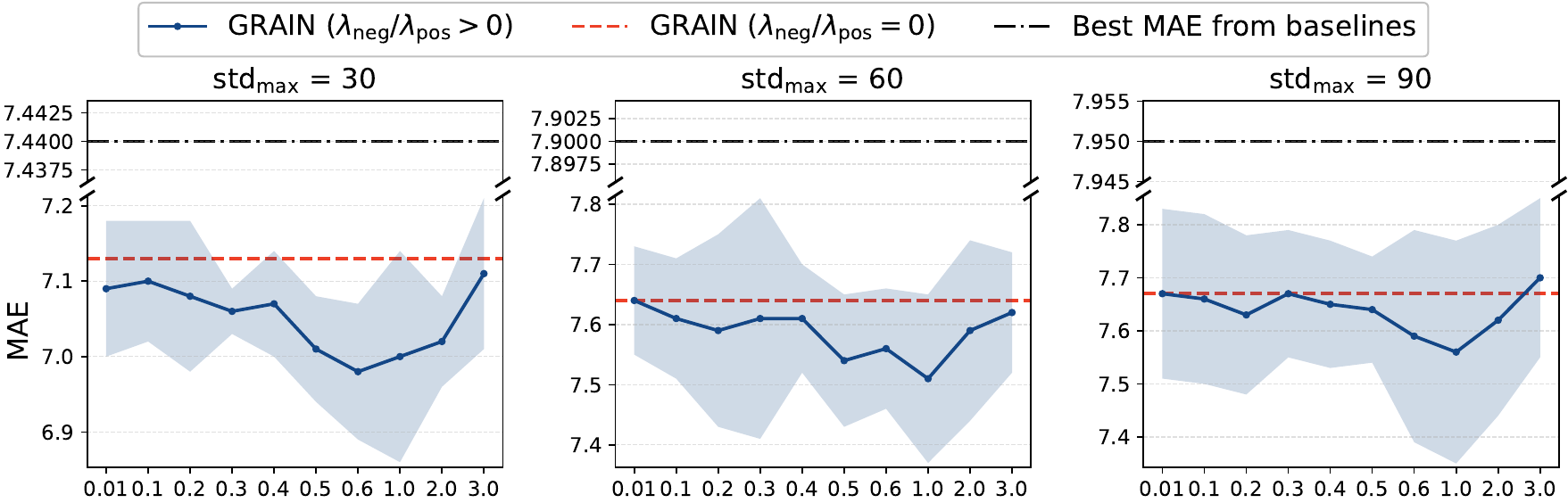}
    \caption{Hyperparameter sensitivity analysis for  $\lambda_{\text{neg}}/\lambda_{\text{pos}}$ on UTKFace.}
    \label{fig:hyper_uot_pos_neg_ratio_MAE}
\end{figure}

\begin{figure}[H]
    \centering
    \includegraphics[width=0.9\linewidth]{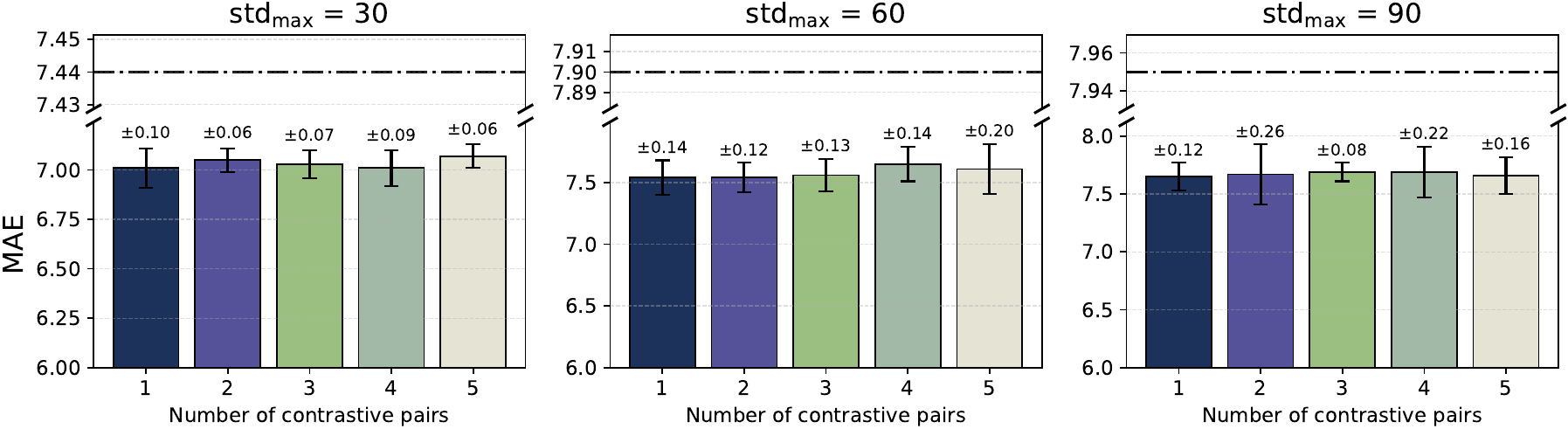}
    \caption{Hyperparameter sensitivity analysis on the number of contrastive pairs on UTKFace. Black dashed lines denote the best-performing baseline method.}
    \label{fig:hyper_uot_pair_MAE}
\end{figure}

\begin{figure}[H]
    \centering
    \includegraphics[width=0.96\linewidth]{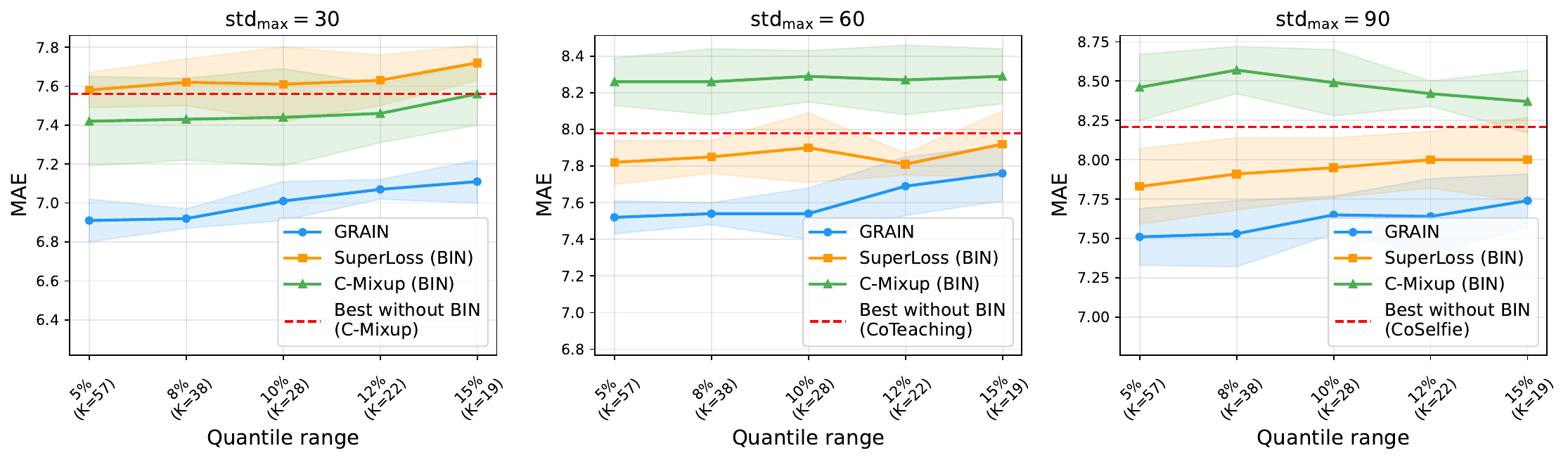}
    \caption{Sensitivity analysis of the binning strategy on UTKFace. The x-axis shows {the symmetric quantile interval around the density peak} $y_{\text{peak}}$ used to determine the bin width (Equation (\ref{eq.bin_width})), with the corresponding number of bins $K$ indicated in parentheses. A larger quantile interval yields a wider bin width and thus fewer bins.
    {GRAIN} (\textcolor[RGB]{33, 150, 243}{\rule{6pt}{6pt}}) is compared against two strong BIN-enhanced baselines, {C-Mixup (BIN)} (\textcolor[RGB]{76, 175, 80}{\rule{6pt}{6pt}}) and {SuperLoss (BIN)} (\textcolor[RGB]{255, 152, 0}{\rule{6pt}{6pt}}). The red dashed line ({\color{red}\textbf{-\,-\,-}}) indicates the best-performing baseline without binning. Shaded regions indicate $\pm$1 standard deviation across 5 random seeds.}
    \label{fig:hyper_bin_num_quantile}
\end{figure}

\begin{figure}[H]
    \centering
    \includegraphics[width=0.94\linewidth]{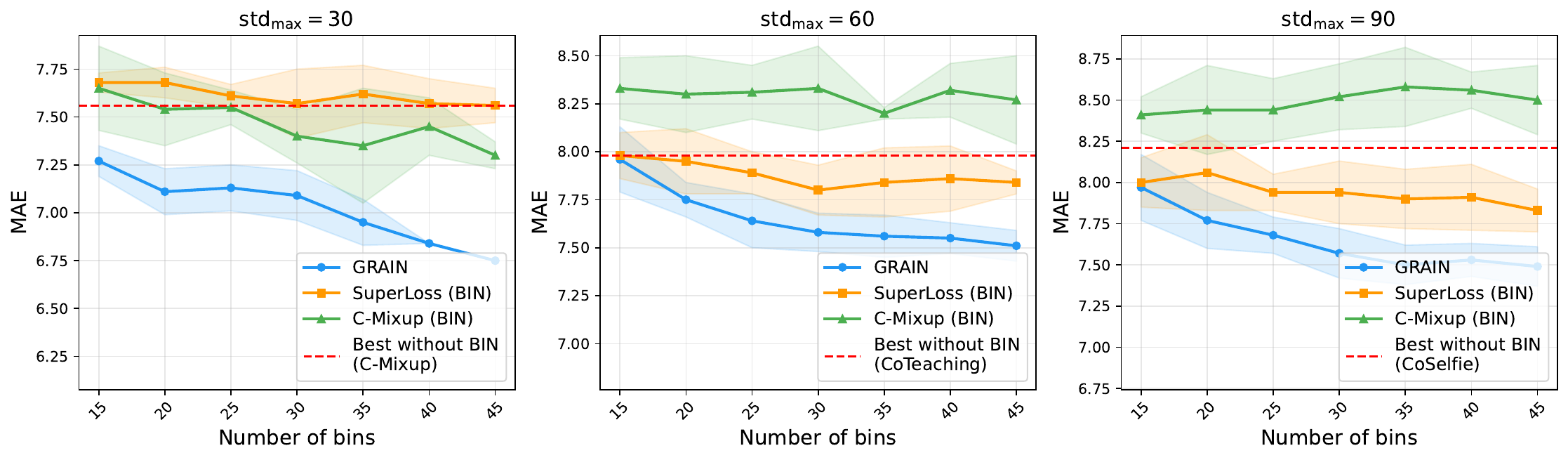}
    \caption{Sensitivity analysis of the binning strategy on UTKFace. The x-axis shows {the number of bins}.
    {GRAIN} (\textcolor[RGB]{33, 150, 243}{\rule{6pt}{6pt}}) is compared against two strong BIN-enhanced baselines, {C-Mixup (BIN)} (\textcolor[RGB]{76, 175, 80}{\rule{6pt}{6pt}}) and {SuperLoss (BIN)} (\textcolor[RGB]{255, 152, 0}{\rule{6pt}{6pt}}). The red dashed line ({\color{red}\textbf{-\,-\,-}}) indicates the best-performing baseline without binning. Shaded regions indicate $\pm$1 standard deviation across 5 random seeds.}
    \label{fig:hyper_bin_num_K}
\end{figure}

\paragraph{Hyperparameter Sensitivity Analysis.} 
Figures~\ref{fig:hyper_lam_pos_MAE} and~\ref{fig:hyper_uot_pos_neg_ratio_MAE} present the hyperparameter sensitivity analysis for $\lambda_{\mathrm{pos}}$ and the ratio $\lambda_{\mathrm{neg}}/\lambda_{\mathrm{pos}}$, respectively, evaluated on the UTKFace dataset under varying levels of simulated Gaussian label corruption. {In Figure~\ref{fig:hyper_lam_pos_MAE}, the red dashed line indicates the performance without positive contrastive regularization ($\lambda_{\mathrm{pos}}=0$), while in Figure~\ref{fig:hyper_uot_pos_neg_ratio_MAE}, it indicates the performance without negative contrastive regularization ($\lambda_{\mathrm{neg}}=0$). The black dashed lines denote the best-performing baseline method.}
Overall, GRAIN remains robust to the choice of contrastive regularization coefficients within a reasonable range across different noise levels. Figure~\ref{fig:hyper_uot_pair_MAE} shows that performance remains stable across different numbers of contrastive pairs. Figures~\ref{fig:hyper_bin_num_quantile} and~\ref{fig:hyper_bin_num_K} show that performance remains stable across different numbers of bins.

\begin{figure}[H]
    \centering
    \includegraphics[width=0.5\linewidth]{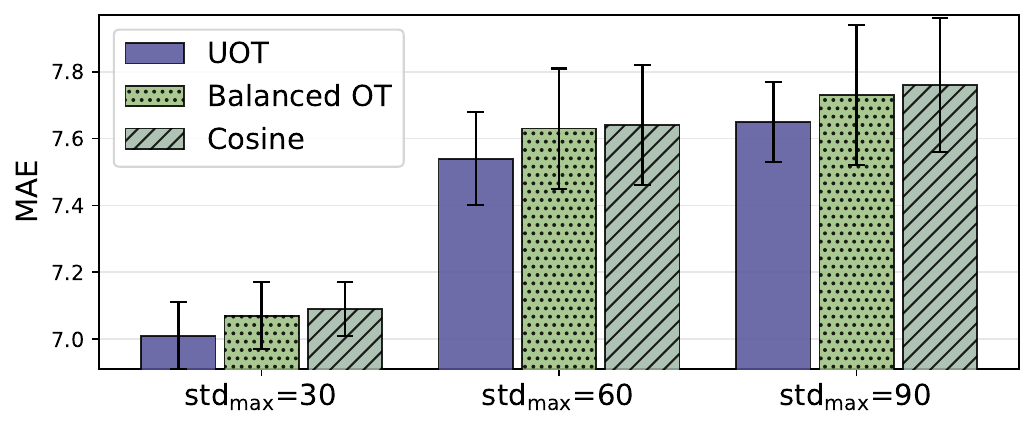}
    \caption{MAE comparison on UTKFace when replacing UOT with balanced OT or cosine similarity in the across-bin contrastive feature regularization. Error bars indicate $\pm 1$ standard deviation across 5 random seeds.}
    \label{fig:uot_ot_cosine}
\end{figure}

\paragraph{UOT, OT, and Cosine Similarity.} Figure~\ref{fig:uot_ot_cosine} compares the performance of GRAIN when using UOT, balanced OT, and cosine similarity as the discrepancy measure in the across-bin contrastive feature regularization.

\begin{figure}[h]
    \centering
    \includegraphics[width=0.9\linewidth]{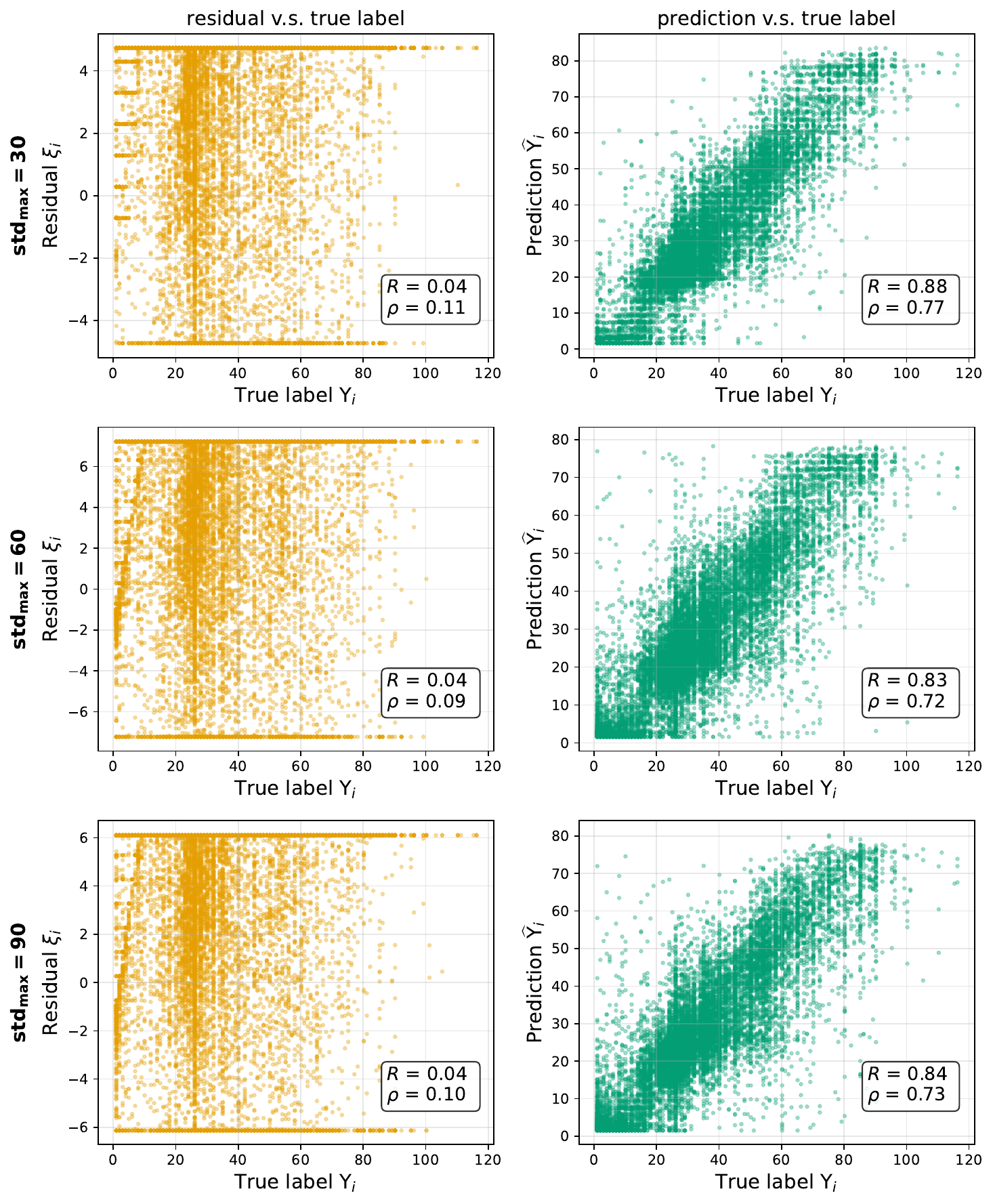}
    \caption{Scatter plots of residual $\xi_i$ vs. true label $\mathrm{Y}_i$ (left column) and prediction $\widehat{\mathrm{Y}}_i$ vs. true label $\mathrm{Y}_i$ (right column) on UTKFace, evaluated at the last training epoch using seed 1. Both Pearson $R$ and Spearman $\rho$ are reported in each panel. }
    \label{fig:scatter_residual}
\end{figure}

\paragraph{Residual Analysis.} The scatter plot in Figure~\ref{fig:scatter_residual} illustrates that the residual head captures label noise rather than predictive signals.

\begin{figure}[H]
    \centering
    \includegraphics[width=0.5\linewidth]
    {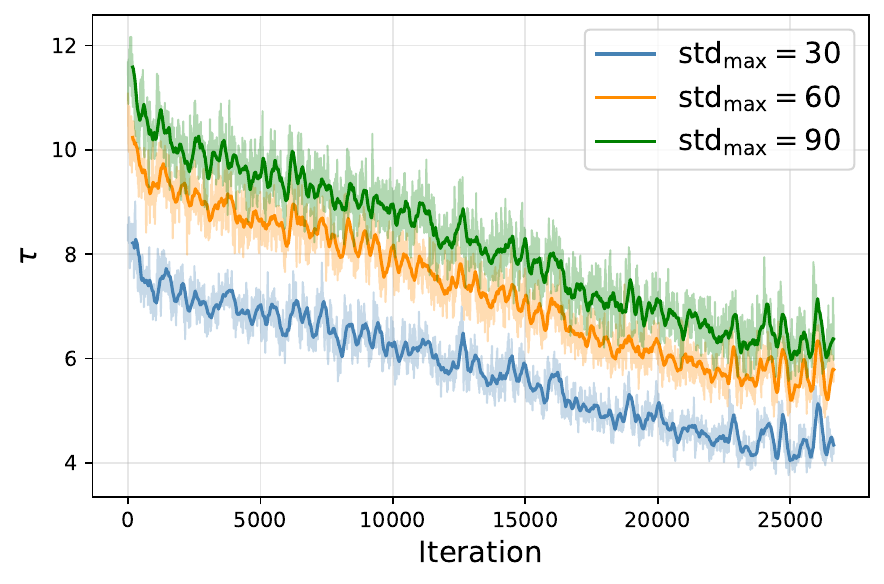}
    \caption{Dynamic evolution of the adaptive
    parameter $\tau$ throughout training on UTKFace.
    The light curves show the raw $\tau$ values,
    while the dark curves show their smoothed
    trajectories via moving average.}
    \label{fig:tau_dynamics}
\end{figure}

\paragraph{$\tau$-Dynamics.}
Figure~\ref{fig:tau_dynamics} shows the dynamic evolution
of the adaptive parameter $\tau$ throughout training

\paragraph{Results on APPA-REAL.} Results on APPA-REAL with real-world label noise are reported in Table~\ref{tab:appa}, demonstrating the effectiveness of GRAIN in handling regression label noise across different tasks.

\begin{table*}[h]
\centering
\caption{Mean $\pm$ {standard deviation} on APPA-REAL dataset with real-world label noise. The best baseline per metric is highlighted in {\color{blue}blue}, and the results produced from our proposed GRAIN and GRAIN-{\small AUG} {methods} are presented in \textbf{boldface}.}
\label{tab:appa}
\setlength{\tabcolsep}{2pt} 
\resizebox{0.45\textwidth}{!}{
\begin{tabular}{l|ccc}
\toprule
\multirow{2}{*}{\bf Method} & \multicolumn{3}{c}{APPA-REAL} \\
\cmidrule(lr){2-4}
 & \bf MAE $\downarrow$ & \bf RMSE $\downarrow$ & $\boldsymbol{R}^2$ $\uparrow$  \\
\midrule

% ==============================================================================
%                      [GRAIN] APPA & BoneAge  - Real - MSE
% ==============================================================================

Vanilla 
& $10.75_{\pm 0.14}$ & $14.83_{\pm 0.16}$ & $0.40_{\pm 0.01}$ 
% & $13.56_{\pm 0.25}$ & $17.44_{\pm 0.36}$ & $0.24_{\pm 0.02}$ & $0.95_{\pm 0.00}$ & {\color{blue}$0.91_{\pm 0.00}$}  
\\

SuperLoss 
& $12.67_{\pm 0.14}$ & $16.77_{\pm 0.18}$ & $0.24_{\pm 0.02}$ 
% & $13.59_{\pm 0.11}$ & $17.33_{\pm 0.18}$ & $0.23_{\pm 0.01}$ & $0.95_{\pm 0.00}$ & {\color{blue}$0.91_{\pm 0.00}$}  
\\

RDI 
& $10.71_{\pm 0.07}$ & $14.80_{\pm 0.13}$ & {\color{blue}$0.41_{\pm 0.01}$} 
% & $13.26_{\pm 0.34}$ & $17.23_{\pm 0.49}$ & $0.23_{\pm 0.04}$ & $0.95_{\pm 0.00}$ & {\color{blue}$0.91_{\pm 0.01}$} 
\\

AUX 
& $10.72_{\pm 0.16}$ & $14.81_{\pm 0.18}$ & {\color{blue}$0.41_{\pm 0.01}$} 
% & $13.35_{\pm 0.11}$ & $17.21_{\pm 0.13}$ & {\color{blue}$0.22_{\pm 0.02}$} & {\color{blue}$0.96_{\pm 0.00}$} & $0.91_{\pm 0.00}$  
\\

C-Mixup 
& $10.66_{\pm 0.08}$ & $14.73_{\pm 0.14}$ & {\color{blue}$0.41_{\pm 0.01}$}  
% & $13.51_{\pm 0.17}$ & $17.52_{\pm 0.23}$ & $0.24_{\pm 0.02}$ & $0.95_{\pm 0.00}$ & {\color{blue}$0.91_{\pm 0.00}$} 
\\

% SPR (MSE) & \\

CDR 
& $10.73_{\pm 0.16}$ & $14.81_{\pm 0.22}$ & {\color{blue}$0.41_{\pm 0.02}$} 
% & $13.34_{\pm 0.14}$ & $17.34_{\pm 0.16}$ & $0.23_{\pm 0.02}$ & $0.95_{\pm 0.00}$ & {\color{blue}$0.91_{\pm 0.00}$} 
\\

% SIGUA & $12.80_{\pm 0.08}$ & $17.03_{\pm 0.13}$ & $1.34_{\pm 0.02}$ & $0.47_{\pm 0.01}$ & $0.21_{\pm 0.01}$ \\

Selfie 
& $12.78_{\pm 0.09}$ & $16.95_{\pm 0.12}$ & $0.22_{\pm 0.01}$ 
% & $15.43_{\pm 0.76}$ & $19.65_{\pm 0.86}$ & $0.27_{\pm 0.03}$ & $0.94_{\pm 0.00}$ & $0.88_{\pm 0.01}$
\\

CoTeaching $^\dagger$ 
& $12.67_{\pm 0.11}$ & $16.85_{\pm 0.16}$ & $0.23_{\pm 0.01}$ 
% & $14.39_{\pm 0.41}$ & $18.36_{\pm 0.70}$ & $0.26_{\pm 0.02}$ & $0.95_{\pm 0.00}$ & $0.90_{\pm 0.01}$ 
\\

CNLCU-S $^\dagger$ 
& $12.76_{\pm 0.09}$ & $16.95_{\pm 0.11}$ & $0.22_{\pm 0.01}$ 
% & $15.97_{\pm 0.53}$ & $20.54_{\pm 0.71}$ & $0.30_{\pm 0.02}$ & $0.94_{\pm 0.00}$ & $0.87_{\pm 0.01}$ 
\\

CNLCU-H $^\dagger$ 
& $12.73_{\pm 0.10}$ & $16.96_{\pm 0.25}$ & $0.22_{\pm 0.02}$ 
% & $15.83_{\pm 0.68}$ & $20.12_{\pm 0.94}$ & $0.33_{\pm 0.02}$ & $0.94_{\pm 0.00}$ & $0.88_{\pm 0.01}$
\\

CoSelfie $^\dagger$ 
& $12.78_{\pm 0.09}$ & $16.95_{\pm 0.12}$ & $0.22_{\pm 0.01}$ 
% & $15.50_{\pm 0.46}$ & $19.70_{\pm 0.46}$ & $0.26_{\pm 0.03}$ & $0.94_{\pm 0.00}$ & $0.88_{\pm 0.01}$ 
\\

\midrule
% ==============================================================================
%                      [GRAIN] APPA & BoneAge  - Real - CE
% ==============================================================================

Vanilla {\small(BIN)} 
& $10.77_{\pm 0.29}$ & $15.69_{\pm 0.48}$ & $0.33_{\pm 0.04}$ 
% & $13.31_{\pm 0.22}$ & $17.43_{\pm 0.53}$ & $0.23_{\pm 0.02}$ & $0.95_{\pm 0.00}$ & {\color{blue}$0.91_{\pm 0.01}$}  
\\

SuperLoss {\small(BIN)} 
& $11.41_{\pm 0.15}$ & $16.03_{\pm 0.18}$ & $0.30_{\pm 0.02}$ 
% & $13.55_{\pm 0.33}$ & $17.41_{\pm 0.33}$ & $0.25_{\pm 0.03}$ & $0.95_{\pm 0.00}$ & {\color{blue}$0.91_{\pm 0.00}$} 
\\

RDI {\small(BIN)} 
& $10.46_{\pm 0.09}$ & $14.78_{\pm 0.06}$ & {\color{blue}$0.41_{\pm 0.01}$} 
% & $13.18_{\pm 0.15}$ & {\color{blue}$16.84_{\pm 0.33}$} & $0.24_{\pm 0.01}$ & $0.96_{\pm 0.00}$ & $0.91_{\pm 0.00}$  
\\

AUX {\small(BIN)} 
& {\color{blue}$10.41_{\pm 0.13}$} & {\color{blue}$14.76_{\pm 0.23}$} & {\color{blue}$0.41_{\pm 0.02}$} 
% & $13.27_{\pm 0.31}$ & $17.16_{\pm 0.68}$ & $0.23_{\pm 0.01}$ & {\color{blue}$0.96_{\pm 0.00}$} & {\color{blue}$0.91_{\pm 0.01}$}
\\

C-Mixup {\small(BIN)} 
& $10.43_{\pm 0.08}$ & $14.98_{\pm 0.19}$ & $0.39_{\pm 0.02}$ 
% & {\color{blue}$13.12_{\pm 0.54}$} & $17.45_{\pm 1.04}$ & {\color{blue}$0.22_{\pm 0.03}$} & $0.95_{\pm 0.01}$ & {\color{blue}$0.91_{\pm 0.01}$}
\\

% SPR (CE) & \\

CDR {\small(BIN)} 
& $10.69_{\pm 0.10}$ & $15.62_{\pm 0.28}$  & $0.34_{\pm 0.02}$ 
% & $13.45_{\pm 0.29}$ & $17.52_{\pm 0.84}$ & $0.24_{\pm 0.02}$ & $0.95_{\pm 0.00}$ & {\color{blue}$0.91_{\pm 0.01}$}
\\

% SIGUA {\small(BIN)} & $12.80_{\pm 0.08}$ & $17.03_{\pm 0.13}$ & $1.34_{\pm 0.02}$ & $0.47_{\pm 0.01}$ & $0.21_{\pm 0.01}$  \\

Selfie {\small(BIN)} 
& $11.69_{\pm 0.22}$ & $16.79_{\pm 0.22}$ & $0.24_{\pm 0.02}$ 
% & $16.35_{\pm 1.13}$ & $20.81_{\pm 1.56}$ & $0.29_{\pm 0.02}$ & $0.94_{\pm 0.01}$ & $0.87_{\pm 0.02}$ 
\\

CoTeaching {\small(BIN)} $^\dagger$ 
& $11.84_{\pm 0.22}$ & $16.92_{\pm 0.31}$ & $0.22_{\pm 0.03}$ 
% & $14.58_{\pm 0.47}$ & $18.68_{\pm 0.75}$ & $0.27_{\pm 0.02}$ & $0.95_{\pm 0.01}$ & $0.89_{\pm 0.01}$
\\

CNLCU-S {\small(BIN)} $^\dagger$ 
& $12.21_{\pm 0.48}$ & $17.52_{\pm 0.86}$ & $0.17_{\pm 0.08}$ 
% & $15.96_{\pm 0.79}$ & $20.22_{\pm 1.01}$ & $0.30_{\pm 0.04}$ & $0.94_{\pm 0.01}$ & $0.88_{\pm 0.01}$ 
\\

CNLCU-H {\small(BIN)} $^\dagger$ 
& $12.09_{\pm 0.27}$ & $17.34_{\pm 0.46}$ & $0.18_{\pm 0.04}$ 
% & $17.39_{\pm 0.97}$ & $22.29_{\pm 1.37}$ & $0.31_{\pm 0.02}$ & $0.93_{\pm 0.01}$ & $0.85_{\pm 0.02}$ 
\\

CoSelfie {\small(BIN)} $^\dagger$ 
& $11.69_{\pm 0.22}$ & $16.79_{\pm 0.22}$ & $0.24_{\pm 0.02}$
% & $14.44_{\pm 0.49}$ & $18.62_{\pm 0.56}$ & $0.27_{\pm 0.03}$ & $0.95_{\pm 0.00}$ & $0.89_{\pm 0.01}$ 
\\

\midrule
\bf GRAIN (ours) 
& $\bm{10.22_{\pm 0.15}}$ & $\bm{14.66_{\pm 0.21}}$  & $\bm{0.42_{\pm 0.02}}$ 
% & $\bm{12.96_{\pm 0.25}}$ & $\bm{16.76_{\pm 0.30}}$ & $\bm{0.22_{\pm 0.01}}$ & $\bm{0.96_{\pm 0.00}}$ & $\bm{0.91_{\pm 0.00}}$
\\

\bf GRAIN-{\small{AUG}} (ours) 
& $\bm{10.00_{\pm 0.12}}$ & $\bm{14.48_{\pm 0.16}}$  & $\bm{0.43_{\pm 0.01}}$
% & $\bm{12.85_{\pm 0.28}}$ & $\bm{16.69_{\pm 0.69}}$ & $\bm{0.22_{\pm 0.02}}$ & $\bm{0.96_{\pm 0.00}}$ & $\bm{0.91_{\pm 0.01}}$
\\
\bottomrule
\end{tabular}
}
\end{table*}

\paragraph{Results on Biwi Kinect.} Tables~\ref{tab:biwi_kinect_yaw_gaussian}, \ref{tab:biwi_kinect_pitch_gaussian}, and \ref{tab:biwi_kinect_roll_gaussian} present the Biwi Kinect results for yaw, pitch, and roll estimation under synthetic Gaussian label noise across varying corruption levels, evaluated with the {three} metrics described in Appendix \ref{appdx.exp_detail}. GRAIN and GRAIN-{\small AUG} consistently outperform all baselines across nearly all noise levels and metrics, demonstrating strong robustness to synthetic label noise. % {\color{red}(the results for GRAIN will be ready later today.)}

% ==============================================================================
% ==============================================================================
%                    [GRAIN] Biwi Kinect (Yaw) - Gaussian
% ==============================================================================
% ==============================================================================

\begin{table*}[h]
\centering
\caption{Mean $\pm$ {standard deviation} on Biwi Kinect (Yaw) under simulated Gaussian label noise with corruption levels controlled by $\text{std}_{\max}$. The best baseline per metric is highlighted in {\color{blue}blue}, and the results produced from our proposed GRAIN and GRAIN-{\small AUG} {methods} are presented in \textbf{boldface}.}
\label{tab:biwi_kinect_yaw_gaussian}
\resizebox{\textwidth}{!}{
\begin{tabular}{l|ccc|ccc|ccc}
\toprule
\multirow{2}{*}{\bf Method} & \multicolumn{3}{c|}{$\text{\bf std}_{\text{\bf max}}\text{\bf  = 30}$} & \multicolumn{3}{c|}{$\text{\bf std}_{\text{\bf max}}\text{\bf  = 60}$} & \multicolumn{3}{c}{$\text{\bf std}_{\text{\bf max}}\text{\bf  = 90}$}\\
\cmidrule(lr){2-4}
\cmidrule(lr){5-7}
\cmidrule(lr){8-10}
 & \bf MAE $\downarrow$ & \bf RMSE $\downarrow$ & $\boldsymbol{R}^2$ $\uparrow$ & \bf MAE $\downarrow$ & \bf RMSE $\downarrow$ & $\boldsymbol{R}^2$ $\uparrow$ & \bf MAE $\downarrow$ & \bf RMSE $\downarrow$ & $\boldsymbol{R}^2$ $\uparrow$   \\
\midrule

% ==============================================================================
%                   [GRAIN] Biwi Kinect (Yaw) - Gaussian - MSE
% ==============================================================================

Vanilla 
& $5.33_{\pm 0.85}$ & $7.66_{\pm 1.65}$ & $0.91_{\pm 0.03}$ 

& $7.54_{\pm 1.09}$ & $10.61_{\pm 1.30}$ & $0.84_{\pm 0.03}$ 

& $9.79_{\pm 1.23}$ & $13.93_{\pm 1.87}$ & $0.73_{\pm 0.05}$ \\

SuperLoss 
& $4.79_{\pm 0.93}$ & $7.26_{\pm 2.28}$ & $0.92_{\pm 0.04}$ 

& $6.09_{\pm 0.66}$ & {\color{blue} $8.46_{\pm 1.19}$} & {\color{blue} $0.90_{\pm 0.03}$} 

& $6.70_{\pm 1.43}$ & $9.61_{\pm 2.56}$ & $0.86_{\pm 0.06}$ 
\\

RDI 
& $5.02_{\pm 0.87}$ & $7.19_{\pm 2.15}$ & $0.92_{\pm 0.04}$ 

& $7.24_{\pm 1.08}$ & $9.97_{\pm 1.75}$ & $0.86_{\pm 0.03}$

& $9.67_{\pm 1.17}$ & $13.63_{\pm 1.53}$ & $0.74_{\pm 0.03}$
\\

AUX 
& $5.87_{\pm 1.07}$ & $8.56_{\pm 2.24}$ & $0.89_{\pm 0.05}$ 

& $7.22_{\pm 1.13}$ & $9.99_{\pm 1.75}$  & $0.86_{\pm 0.03}$ 

& $9.60_{\pm 0.67}$ & $13.29_{\pm 0.96}$ & $0.75_{\pm 0.04}$ 
\\

C-Mixup 
& {\color{blue} $4.75_{\pm 0.70}$} & {\color{blue} $6.94_{\pm 1.38}$} & {\color{blue} $0.93_{\pm 0.03}$}

& $6.49_{\pm 1.04}$ & $9.84_{\pm 2.13}$ & $0.86_{\pm 0.05}$ 

& $8.71_{\pm 0.97}$ & $11.75_{\pm 0.93}$ & $0.80_{\pm 0.05}$ \\

CDR 
& $5.26_{\pm 0.87}$ & $7.78_{\pm 1.89}$ & $0.91_{\pm 0.03}$ 

& $7.66_{\pm 1.29}$ & $10.65_{\pm 1.75}$ & $0.84_{\pm 0.03}$ 

& $9.46_{\pm 0.50}$ & $13.29_{\pm 0.74}$  & $0.73_{\pm 0.04}$ 
\\

Selfie 
& $5.22_{\pm 1.03}$ & $7.99_{\pm 2.45}$ & $0.90_{\pm 0.05}$

& $6.32_{\pm 0.94}$ & $9.08_{\pm 2.10}$ & $0.88_{\pm 0.05}$ 

& $7.44_{\pm 1.26}$ & $10.92_{\pm 2.84}$ & $0.81_{\pm 0.09}$ \\

CoTeaching $^\dagger$ 
& $4.93_{\pm 0.75}$ & $7.71_{\pm 1.95}$ & $0.91_{\pm 0.04}$ 

& {\color{blue} $5.84_{\pm 1.28}$} & {\color{blue} $8.64_{\pm 2.58}$} & $0.89_{\pm 0.05}$ 

& $6.68_{\pm 0.47}$ & $9.67_{\pm 1.66}$ & $0.86_{\pm 0.05}$ 
\\

CNLCU-S $^\dagger$
& $4.97_{\pm 0.86}$ & $7.38_{\pm 1.81}$ & $0.92_{\pm 0.03}$

& $6.11_{\pm 1.00}$ & $8.75_{\pm 1.83}$ & $0.89_{\pm 0.04}$ 

& $6.42_{\pm 0.62}$ & {\color{blue} $9.26_{\pm 1.68}$} & {\color{blue} $0.87_{\pm 0.05}$} 
\\

CNLCU-H $^\dagger$ 
& $4.88_{\pm 0.86}$ & $7.27_{\pm 1.84}$ & $0.92_{\pm 0.03}$ 

& $6.09_{\pm 1.27}$ & $8.86_{\pm 2.27}$ & $0.88_{\pm 0.05}$ 

& {\color{blue} $6.41_{\pm 0.78}$} & $9.42_{\pm 1.96}$ & {\color{blue} $0.87_{\pm 0.05}$}
\\

CoSelfie $^\dagger$ 
& $5.08_{\pm 1.00}$ & $7.81_{\pm 2.63}$ & $0.91_{\pm 0.05}$

& $5.94_{\pm 1.17}$ & $8.76_{\pm 2.16}$ & $0.88_{\pm 0.05}$

& $6.74_{\pm 1.16}$ & $10.08_{\pm 2.23}$ & $0.84_{\pm 0.09}$ 
\\

% Huber & \\
\midrule

% ==============================================================================
%                   [GRAIN] Biwi Kinect (Yaw) - Gaussian - CE
% ==============================================================================

Vanilla {\small(BIN)} 
& $6.18_{\pm 1.44}$ & $9.03_{\pm 2.88}$ & $0.88_{\pm 0.06}$ 

& $7.33_{\pm 1.76}$ & $10.14_{\pm 2.74}$ & $0.85_{\pm 0.07}$ 

& $7.75_{\pm 1.28}$ & $11.08_{\pm 2.46}$ & $0.82_{\pm 0.07}$ \\

SuperLoss {\small(BIN)}
& $5.55_{\pm 1.36}$ & $8.09_{\pm 2.16}$ & $0.90_{\pm 0.04}$ 

& $6.62_{\pm 1.21}$ & $9.48_{\pm 2.52}$ & $0.87_{\pm 0.06}$ 

& $7.17_{\pm 0.94}$ & $10.09_{\pm 1.92}$ & $0.85_{\pm 0.05}$ 
\\

RDI {\small(BIN)} 
& $5.76_{\pm 1.09}$ & $8.29_{\pm 2.07}$ & $0.90_{\pm 0.04}$

& $6.91_{\pm 1.45}$ & $9.45_{\pm 2.40}$ & $0.87_{\pm 0.05}$ 

& $7.66_{\pm 1.04}$ & $10.81_{\pm 1.89}$ & $0.82_{\pm 0.06}$ 
\\

AUX {\small(BIN)} 
& $5.11_{\pm 0.92}$ & $7.59_{\pm 2.04}$ & $0.91_{\pm 0.04}$ 

& $6.91_{\pm 1.45}$ & $9.45_{\pm 2.40}$ & $0.87_{\pm 0.05}$ 

& $7.30_{\pm 1.17}$ & $10.18_{\pm 2.12}$ & $0.85_{\pm 0.05}$ 
\\

C-Mixup {\small(BIN)} 
& $5.61_{\pm 1.12}$ & $8.26_{\pm 2.38}$ & $0.90_{\pm 0.05}$

& $7.21_{\pm 1.73}$ & $10.33_{\pm 2.86}$ & $0.84_{\pm 0.07}$ 

& $7.41_{\pm 1.68}$ & $10.10_{\pm 2.27}$ & $0.85_{\pm 0.06}$
\\

CDR {\small(BIN)} 
& $5.99_{\pm 1.03}$ & $8.43_{\pm 1.63}$ & $0.90_{\pm 0.03}$ 

& $7.45_{\pm 1.41}$ & $10.47_{\pm 2.50}$ & $0.83_{\pm 0.07}$

& $7.57_{\pm 1.71}$ & $10.77_{\pm 2.59}$ & $0.83_{\pm 0.07}$ 
\\

Selfie {\small(BIN)} 
& $6.82_{\pm 1.55}$ & $9.92_{\pm 2.83}$ & $0.85_{\pm 0.07}$ 

& $6.49_{\pm 1.06}$ & $9.17_{\pm 2.07}$ & $0.88_{\pm 0.05}$ 

& $8.10_{\pm 1.19}$ & $11.23_{\pm 2.00}$ & $0.81_{\pm 0.07}$
\\

CoTeaching {\small(BIN)} $^\dagger$ 
& $6.08_{\pm 1.25}$ & $8.90_{\pm 2.14}$ & $0.88_{\pm 0.05}$ 

& $7.49_{\pm 1.14}$ & $10.52_{\pm 2.02}$ & $0.84_{\pm 0.05}$ 

& $9.92_{\pm 2.44}$ & $14.79_{\pm 4.51}$ & $0.70_{\pm 0.11}$ \\

CNLCU-S {\small(BIN)} $^\dagger$ 
& $5.46_{\pm 1.14}$ & $8.12_{\pm 2.31}$ & $0.90_{\pm 0.05}$ 

& $7.27_{\pm 1.38}$ & $10.01_{\pm 1.77}$ & $0.85_{\pm 0.04}$ 

& $8.48_{\pm 2.07}$ & $12.52_{\pm 3.77}$ & $0.74_{\pm 0.19}$ \\

CNLCU-H {\small(BIN)} $^\dagger$ 
& $6.19_{\pm 1.08}$ & $8.67_{\pm 1.84}$ & $0.89_{\pm 0.04}$

& $7.35_{\pm 1.60}$ & $10.34_{\pm 2.89}$ & $0.84_{\pm 0.07}$ 

& $7.73_{\pm 1.00}$ & $10.96_{\pm 1.60}$ & $0.83_{\pm 0.04}$ 
\\

CoSelfie {\small(BIN)} $^\dagger$ 
& $6.26_{\pm 1.67}$ & $8.94_{\pm 3.18}$ & $0.88_{\pm 0.07}$

& $8.15_{\pm 1.54}$ & $11.51_{\pm 2.33}$ & $0.80_{\pm 0.07}$

& $9.21_{\pm 1.08}$ & $13.10_{\pm 2.02}$ & $0.75_{\pm 0.07}$ \\

\midrule
\textbf{GRAIN (ours)} 
& $\bm{4.88_{\pm 0.86}}$ & $\bm{6.49_{\pm 0.67}}$ & $\bm{0.93_{\pm 0.03}}$ 

& $\bm{5.57_{\pm 0.89}}$ & $\bm{7.87_{\pm 1.38}}$ & $\bm{0.90_{\pm 0.04}}$

& $\bm{6.00_{\pm 0.88}}$ & $\bm{8.46_{\pm 1.58}}$ & $\bm{0.88_{\pm 0.07}}$ 
\\
\textbf{GRAIN-{\small AUG} (ours)}
& $\bm{4.76_{\pm 0.47}}$ & $\bm{6.59_{\pm 0.12}}$ & $\bm{0.93_{\pm 0.02}}$ 

& $\bm{5.30_{\pm 0.61}}$ & $\bm{7.18_{\pm 0.89}}$ & $\bm{0.92_{\pm 0.03}}$

& $\bm{5.77_{\pm 0.81}}$ & $\bm{8.03_{\pm 1.16}}$ & $\bm{0.90_{\pm 0.03}}$
\\
\bottomrule
\end{tabular}
}
\end{table*}

% ==============================================================================
% ==============================================================================
%                    [GRAIN] Biwi Kinect (Pitch) - Gaussian
% ==============================================================================
% ==============================================================================

\begin{table*}[h]
\centering
\caption{Mean $\pm$ {standard deviation} on Biwi Kinect (Pitch) under simulated Gaussian label noise with corruption levels controlled by $\text{std}_{\max}$. The best baseline per metric is highlighted in {\color{blue}blue}, and the results produced from our proposed GRAIN and GRAIN-{\small AUG} {methods} are presented in \textbf{boldface}.}
\label{tab:biwi_kinect_pitch_gaussian}
\resizebox{\textwidth}{!}{
\begin{tabular}{l|ccc|ccc|ccc}
\toprule
\multirow{2}{*}{\bf Method} & \multicolumn{3}{c|}{$\text{\bf std}_{\text{\bf max}}\text{\bf  = 30}$} & \multicolumn{3}{c|}{$\text{\bf std}_{\text{\bf max}}\text{\bf  = 60}$} & \multicolumn{3}{c}{$\text{\bf std}_{\text{\bf max}}\text{\bf  = 90}$}\\
\cmidrule(lr){2-4}
\cmidrule(lr){5-7}
\cmidrule(lr){8-10}
 & \bf MAE $\downarrow$ & \bf RMSE $\downarrow$ & $\boldsymbol{R}^2$ $\uparrow$ & \bf MAE $\downarrow$ & \bf RMSE $\downarrow$ & $\boldsymbol{R}^2$ $\uparrow$ & \bf MAE $\downarrow$ & \bf RMSE $\downarrow$ & $\boldsymbol{R}^2$ $\uparrow$   \\
\midrule

% ==============================================================================
%                  [GRAIN] Biwi Kinect (Pitch) - Gaussian - MSE
% ==============================================================================

Vanilla 
& $6.18_{\pm 1.78}$ & $8.77_{\pm 3.01}$ & $0.85_{\pm 0.11}$

& $8.01_{\pm 1.97}$ & $11.48_{\pm 2.71}$ & $0.77_{\pm 0.10}$

& $10.59_{\pm 1.83}$ & $14.93_{\pm 2.24}$ & $0.62_{\pm 0.12}$ 
\\

SuperLoss 
& $5.78_{\pm 1.36}$ & $8.07_{\pm 2.18}$ & $0.88_{\pm 0.06}$

% & {\color{blue} $6.68_{\pm 1.57}$} & $9.44_{\pm 2.41}$ & $3.37_{\pm 4.82}$ & {\color{blue} $0.93_{\pm 0.04}$} & $0.84_{\pm 0.08}$ 
& {\color{blue} $6.80_{\pm 1.60}$} & $9.43_{\pm 2.05}$ & $0.85_{\pm 0.06}$ 

% & {\color{blue} $7.63_{\pm 2.00}$} & {\color{blue} $11.37_{\pm 4.05}$} & $2.58_{\pm 2.44}$ & $0.87_{\pm 0.11}$ & $0.75_{\pm 0.19}$ \\
& {\color{blue} $8.12_{\pm 2.01}$} & $11.75_{\pm 3.10}$ & $0.76_{\pm 0.12}$ \\

RDI 
& $6.25_{\pm 1.68}$ & $8.81_{\pm 2.55}$ & $0.86_{\pm 0.07}$

& $8.15_{\pm 1.81}$ & $11.67_{\pm 2.81}$ & $0.76_{\pm 0.12}$ 

& $10.10_{\pm 1.75}$ & $14.36_{\pm 2.65}$ & $0.65_{\pm 0.13}$ \\

AUX 
& $6.17_{\pm 1.76}$ & $8.61_{\pm 2.42}$ & $0.86_{\pm 0.08}$ 

& $8.23_{\pm 1.66}$ & $11.85_{\pm 2.88}$ & $0.75_{\pm 0.12}$ 

& $10.05_{\pm 1.55}$ & $13.77_{\pm 2.07}$ & $0.68_{\pm 0.10}$ \\

C-Mixup 
& $5.74_{\pm 1.71}$ & $8.21_{\pm 2.78}$ & $0.87_{\pm 0.08}$

& $7.36_{\pm 1.25}$ & $10.15_{\pm 1.86}$ & $0.83_{\pm 0.06}$ 

& $9.35_{\pm 2.04}$ & $13.21_{\pm 2.71}$ & $0.69_{\pm 0.13}$ \\

CDR 
& $6.61_{\pm 1.87}$ & $9.17_{\pm 2.55}$ & $0.85_{\pm 0.08}$ 

& $8.36_{\pm 1.61}$ & $12.02_{\pm 2.83}$ & $0.74_{\pm 0.14}$ 

& $10.38_{\pm 1.78}$ & $14.35_{\pm 2.26}$ & $0.65_{\pm 0.11}$ 
\\

Selfie 
& $5.98_{\pm 1.82}$ & $8.68_{\pm 3.11}$ & $0.86_{\pm 0.10}$ 

& $8.38_{\pm 2.20}$ & $12.20_{\pm 2.84}$ & $0.74_{\pm 0.13}$

& $11.06_{\pm 2.95}$ & $15.13_{\pm 3.24}$ & $0.60_{\pm 0.18}$ \\

CoTeaching $^\dagger$ 
& $6.13_{\pm 1.95}$ & $8.53_{\pm 2.95}$ & $0.86_{\pm 0.08}$ 

& $7.59_{\pm 2.66}$ & $11.29_{\pm 3.91}$ & $0.76_{\pm 0.15}$ 

& $9.02_{\pm 3.14}$ & $13.23_{\pm 4.23}$ & $0.68_{\pm 0.18}$ 
\\

CNLCU-S $^\dagger$ 
& $6.06_{\pm 1.87}$ & $8.76_{\pm 2.83}$ & $0.86_{\pm 0.08}$ 

& $7.76_{\pm 2.46}$ & $11.34_{\pm 3.46}$ & $0.77_{\pm 0.13}$ 

& $9.84_{\pm 2.62}$ & $14.63_{\pm 3.52}$ & $0.62_{\pm 0.17}$ 
\\

CNLCU-H $^\dagger$ 
& $6.18_{\pm 1.87}$ & $8.62_{\pm 2.79}$ & $0.86_{\pm 0.08}$ 

& $7.87_{\pm 2.74}$ & $11.50_{\pm 3.59}$ & $0.76_{\pm 0.15}$ 

& $10.16_{\pm 2.67}$ & $15.26_{\pm 3.34}$ & $0.60_{\pm 0.17}$ 
\\

CoSelfie $^\dagger$ 
& $6.07_{\pm 1.82}$ & $8.29_{\pm 2.66}$ & $0.87_{\pm 0.07}$ 

& $8.02_{\pm 2.51}$ & $11.75_{\pm 2.99}$ & $0.76_{\pm 0.13}$

& $10.24_{\pm 2.61}$ & $15.03_{\pm 3.40}$ & $0.60_{\pm 0.21}$ 

\\

\midrule

% ==============================================================================
%                  [GRAIN] Biwi Kinect (Pitch) - Gaussian - CE
% ==============================================================================

Vanilla {\small(BIN)} 
& $6.16_{\pm 1.40}$ & $8.91_{\pm 2.59}$ & $0.85_{\pm 0.09}$  

& $7.21_{\pm 1.46}$ & $9.42_{\pm 1.95}$ & $0.85_{\pm 0.07}$ 

& $8.95_{\pm 2.33}$ & $12.92_{\pm 4.10}$ & $0.69_{\pm 0.19}$
\\

SuperLoss {\small(BIN)} 
& {\color{blue} $5.51_{\pm 1.48}$} & {\color{blue} $7.53_{\pm 2.12}$} & {\color{blue} $0.90_{\pm 0.05}$} 

& $7.02_{\pm 0.62}$ & {\color{blue} $9.21_{\pm 0.82}$} & {\color{blue} $0.86_{\pm 0.03}$} 

& $8.44_{\pm 2.09}$ & $12.52_{\pm 3.36}$ & $0.72_{\pm 0.14}$ \\

RDI {\small(BIN)} 
& $6.25_{\pm 1.68}$ & $8.81_{\pm 2.55}$ & $0.86_{\pm 0.07}$ 

& $7.41_{\pm 1.36}$ & $9.69_{\pm 1.78}$ & $0.84_{\pm 0.05}$ 

& $8.71_{\pm 2.04}$ & $11.98_{\pm 3.14}$ & $0.75_{\pm 0.12}$ \\

AUX {\small(BIN)} 
& $6.04_{\pm 1.22}$ & $8.22_{\pm 1.91}$ & $0.88_{\pm 0.06}$ 

& $7.52_{\pm 1.89}$ & $10.11_{\pm 2.80}$ & $0.81_{\pm 0.11}$ 

& $8.38_{\pm 2.04}$ & {\color{blue} $11.58_{\pm 3.28}$} & {\color{blue} $0.76_{\pm 0.13}$} 
\\

C-Mixup {\small(BIN)} 
& $5.62_{\pm 0.99}$ & $7.75_{\pm 1.68}$ & {\color{blue} $0.90_{\pm 0.04}$} 

& $6.93_{\pm 1.08}$ & $9.57_{\pm 1.63}$ & $0.84_{\pm 0.06}$

& $8.64_{\pm 2.39}$ & $12.22_{\pm 4.35}$ & $0.71_{\pm 0.22}$ \\

CDR {\small(BIN)} 
& $5.66_{\pm 1.52}$ & $7.54_{\pm 1.92}$ & {\color{blue} $0.90_{\pm 0.05}$} 

& $8.36_{\pm 1.61}$ & $12.02_{\pm 2.83}$ & $0.74_{\pm 0.14}$ 

& $10.38_{\pm 1.78}$ & $14.35_{\pm 2.26}$ & $0.65_{\pm 0.11}$ 
\\

Selfie {\small(BIN)} 
& $7.44_{\pm 2.62}$ & $10.43_{\pm 3.60}$ & $0.80_{\pm 0.11}$ 

& $12.81_{\pm 1.97}$ & $17.05_{\pm 2.56}$ & $0.51_{\pm 0.16}$ 

& $14.83_{\pm 2.38}$ & $20.26_{\pm 2.71}$ & $0.32_{\pm 0.15}$
\\

CoTeaching {\small(BIN)} $^\dagger$ 
& $7.91_{\pm 2.24}$ & $11.81_{\pm 3.00}$ & $0.75_{\pm 0.13}$ 

& $10.75_{\pm 2.87}$ & $15.75_{\pm 4.05}$ & $0.58_{\pm 0.18}$ 

& $13.06_{\pm 2.03}$ & $17.71_{\pm 3.02}$ & $0.47_{\pm 0.19}$ 
\\

CNLCU-S {\small(BIN)} $^\dagger$ 
& $6.78_{\pm 2.11}$ & $9.48_{\pm 2.65}$ & $0.84_{\pm 0.08}$ 

& $9.64_{\pm 3.18}$ & $14.05_{\pm 4.52}$ & $0.64_{\pm 0.20}$ 

& $13.35_{\pm 3.06}$ & $17.91_{\pm 3.78}$ & $0.46_{\pm 0.19}$ 
\\

CNLCU-H {\small(BIN)} $^\dagger$ 
& $7.75_{\pm 2.16}$ & $11.35_{\pm 3.47}$ & $0.76_{\pm 0.14}$ 

& $9.67_{\pm 2.28}$ & $14.38_{\pm 3.24}$ & $0.65_{\pm 0.14}$ 

& $13.04_{\pm 2.04}$ & $17.89_{\pm 2.77}$ & $0.46_{\pm 0.17}$ 
\\

CoSelfie {\small(BIN)} $^\dagger$ 
& $7.25_{\pm 2.44}$ & $10.78_{\pm 4.29}$ & $0.78_{\pm 0.15}$ 

& $9.29_{\pm 2.51}$ & $13.75_{\pm 3.54}$ & $0.67_{\pm 0.15}$ 

& $14.08_{\pm 3.40}$ & $19.25_{\pm 4.65}$ & $0.36_{\pm 0.25}$ 
\\

\midrule
\textbf{GRAIN (ours)} 
& $\bm{5.26_{\pm 0.95}}$ & $\bm{7.37_{\pm 1.66}}$ & $\bm{0.90_{\pm 0.05}}$

& $\bm{6.03_{\pm 1.27}}$ & $\bm{8.12_{\pm 2.00}}$ & $\bm{0.89_{\pm 0.05}}$

& $\bm{7.59_{\pm 2.21}}$ & $\bm{11.05_{\pm 3.60}}$ & $\bm{0.78_{\pm 0.12}}$ 
\\
\textbf{GRAIN-{\small AUG} (ours)} 
& $\bm{4.97_{\pm 1.41}}$ & $\bm{6.74_{\pm 1.71}}$ & $\bm{0.92_{\pm 0.04}}$ 

& $\bm{5.83_{\pm 1.75}}$ & $\bm{8.39_{\pm 3.04}}$ & $\bm{0.86_{\pm 0.10}}$ 

& $\bm{6.88_{\pm 2.36}}$ & $\bm{9.42_{\pm 3.40}}$ & $\bm{0.84_{\pm 0.09}}$ 
\\
\bottomrule
\end{tabular}
}
\end{table*}

% ==============================================================================
% ==============================================================================
%                    [GRAIN] Biwi Kinect (Roll) - Gaussian
% ==============================================================================
% ==============================================================================

\begin{table*}[h]
\centering
\caption{Mean $\pm$ {standard deviation} on Biwi Kinect (Roll) under simulated Gaussian label noise with corruption levels controlled by $\text{std}_{\max}$. The best baseline per metric is highlighted in {\color{blue}blue}, and the results produced from our proposed GRAIN and GRAIN-{\small AUG} {methods} are presented in \textbf{boldface}.}
\label{tab:biwi_kinect_roll_gaussian}
\resizebox{\textwidth}{!}{
\begin{tabular}{l|ccc|ccc|ccc}
\toprule
\multirow{2}{*}{\bf Method} & \multicolumn{3}{c|}{$\text{\bf std}_{\text{\bf max}}\text{\bf  = 30}$} & \multicolumn{3}{c|}{$\text{\bf std}_{\text{\bf max}}\text{\bf  = 60}$} & \multicolumn{3}{c}{$\text{\bf std}_{\text{\bf max}}\text{\bf  = 90}$}\\
\cmidrule(lr){2-4}
\cmidrule(lr){5-7}
\cmidrule(lr){8-10}
 & \bf MAE $\downarrow$ & \bf RMSE $\downarrow$ & $\boldsymbol{R}^2$ $\uparrow$ & \bf MAE $\downarrow$ & \bf RMSE $\downarrow$ & $\boldsymbol{R}^2$ $\uparrow$ & \bf MAE $\downarrow$ & \bf RMSE $\downarrow$ & $\boldsymbol{R}^2$ $\uparrow$   \\
\midrule

% ==============================================================================
%                 [GRAIN] Biwi Kinect (Roll) - Gaussian - MSE
% ==============================================================================

Vanilla 
& $5.94_{\pm 2.05}$ & $8.78_{\pm 3.34}$ & $0.54_{\pm 0.07}$ 

& $7.04_{\pm 2.39}$ & $10.11_{\pm 3.59}$ & $0.38_{\pm 0.06}$ 

& $8.13_{\pm 2.92}$ & $11.95_{\pm 4.62}$ & $0.15_{\pm 0.12}$ \\

SuperLoss 
& $5.70_{\pm 2.19}$ & $8.59_{\pm 3.45}$ & $0.57_{\pm 0.07}$ 

& $6.98_{\pm 2.13}$ & $10.51_{\pm 3.67}$ & $0.14_{\pm 0.15}$ 

& $8.17_{\pm 3.07}$ & $12.35_{\pm 4.96}$ & $0.04_{\pm 0.05}$ 
\\

RDI 
& $5.45_{\pm 2.15}$ & $7.97_{\pm 3.21}$ & {\color{blue} $0.63_{\pm 0.06}$} 

& {\color{blue} $6.41_{\pm 2.20}$} & {\color{blue} $9.60_{\pm 3.41}$} & {\color{blue} $0.44_{\pm 0.07}$} 

& $7.94_{\pm 2.76}$ & $11.70_{\pm 4.25}$ & $0.17_{\pm 0.15}$ \\

AUX 
& $5.29_{\pm 2.01}$ & $8.04_{\pm 3.17}$ & $0.61_{\pm 0.05}$ 

& $6.67_{\pm 2.02}$ & $10.02_{\pm 3.34}$ & $0.38_{\pm 0.02}$ 

& {\color{blue} $7.55_{\pm 2.59}$} & {\color{blue} $11.25_{\pm 3.85}$} & {$0.22_{\pm 0.15}$}
\\

C-Mixup 
& $5.29_{\pm 1.88}$ & {\color{blue} $7.74_{\pm 2.77}$} & {\color{blue} $0.63_{\pm 0.04}$} 

& $6.80_{\pm 2.49}$ & $10.10_{\pm 3.88}$ & $0.38_{\pm 0.12}$ 

& $7.85_{\pm 2.93}$ & $11.60_{\pm 4.49}$ & $0.20_{\pm 0.16}$ 
\\

CDR 
& $5.39_{\pm 1.88}$ & $8.04_{\pm 3.07}$ & $0.61_{\pm 0.07}$ 

& $6.68_{\pm 2.54}$ & $9.84_{\pm 3.73}$ & $0.32_{\pm 0.06}$ 

& $7.98_{\pm 3.54}$ & $11.62_{\pm 5.15}$ & {\color{blue} $0.24_{\pm 0.08}$} 
\\

Selfie 
& $6.73_{\pm 3.11}$ & $9.91_{\pm 4.89}$ & $0.43_{\pm 0.22}$ 

& $7.52_{\pm 2.21}$ & $11.27_{\pm 3.54}$ & $0.01_{\pm 0.03}$ 

& $8.40_{\pm 2.87}$ & $12.69_{\pm 4.38}$ & $0.01_{\pm 0.02}$ \\

CoTeaching $^\dagger$ 
& $6.08_{\pm 2.71}$ & $9.19_{\pm 4.26}$ & $0.51_{\pm 0.15}$ 

& $8.20_{\pm 2.90}$ & $12.26_{\pm 4.54}$ & $0.09_{\pm 0.12}$

& $8.32_{\pm 2.82}$ & $12.45_{\pm 4.47}$ & $0.06_{\pm 0.08}$ 
\\

CNLCU-S $^\dagger$ 
& $6.30_{\pm 2.94}$ & $9.62_{\pm 4.67}$ & $0.47_{\pm 0.19}$ 

& $7.91_{\pm 2.99}$ & $11.76_{\pm 4.61}$ & $0.16_{\pm 0.18}$ 

& $8.44_{\pm 2.74}$ & $12.66_{\pm 4.30}$ & $0.01_{\pm 0.04}$ 
\\

CNLCU-H $^\dagger$ 
& $6.15_{\pm 2.71}$ & $9.46_{\pm 4.47}$ & $0.49_{\pm 0.16}$

& $7.82_{\pm 3.04}$ & $11.85_{\pm 4.59}$ & $0.14_{\pm 0.16}$ 

& $8.47_{\pm 2.73}$ & $12.67_{\pm 4.23}$ & $0.01_{\pm 0.02}$ 
\\

CoSelfie $^\dagger$ 
& $6.58_{\pm 3.35}$ & $9.98_{\pm 5.14}$ & $0.43_{\pm 0.24}$ 

& $8.06_{\pm 3.03}$ & $12.22_{\pm 4.63}$ & $0.10_{\pm 0.13}$ 

& $8.48_{\pm 2.78}$ & $12.75_{\pm 4.30}$ & $0.00_{\pm 0.01}$ \\

\midrule
% ==============================================================================
%                 [GRAIN] Biwi Kinect (Roll) - Gaussian - CE
% ==============================================================================

Vanilla {\small(BIN)} 
& {\color{blue} $5.23_{\pm 1.87}$} & $8.26_{\pm 3.42}$ & $0.60_{\pm 0.07}$ 

& $7.77_{\pm 2.87}$ & $11.65_{\pm 4.36}$ & $0.17_{\pm 0.14}$ 

& $8.40_{\pm 2.82}$ & $12.62_{\pm 4.33}$ & $0.02_{\pm 0.04}$ 
\\

SuperLoss {\small(BIN)}
& $7.30_{\pm 2.47}$ & $10.58_{\pm 3.91}$ & $0.31_{\pm 0.11}$ 

& $8.26_{\pm 3.01}$ & $12.34_{\pm 4.63}$ & $0.09_{\pm 0.12}$ 

& $8.37_{\pm 2.82}$ & $12.58_{\pm 4.35}$ & $0.03_{\pm 0.03}$ 
\\

RDI {\small(BIN)} 
& $6.11_{\pm 2.18}$ & $9.04_{\pm 3.66}$ & $0.51_{\pm 0.10}$ 

& $8.24_{\pm 2.65}$ & $12.35_{\pm 4.01}$ & $0.05_{\pm 0.04}$ 

& $8.35_{\pm 2.79}$ & $12.64_{\pm 4.34}$ & $0.02_{\pm 0.02}$ \\

AUX {\small(BIN)} 
& $5.72_{\pm 2.30}$ & $8.77_{\pm 3.90}$ & $0.56_{\pm 0.12}$ 

& $8.15_{\pm 2.44}$ & $12.17_{\pm 3.74}$ & $0.06_{\pm 0.10}$ 

& $8.25_{\pm 2.87}$ & $12.48_{\pm 4.42}$ & $0.05_{\pm 0.05}$ 
\\

C-Mixup {\small(BIN)} 
& $5.87_{\pm 1.95}$ & $8.80_{\pm 3.34}$ & $0.53_{\pm 0.07}$ 

& $7.88_{\pm 2.79}$ & $11.50_{\pm 4.52}$ & $0.21_{\pm 0.11}$ 

& $8.06_{\pm 2.70}$ & $11.93_{\pm 4.07}$ & $0.12_{\pm 0.03}$ \\

CDR {\small(BIN)} 
& $5.92_{\pm 2.20}$ & $8.86_{\pm 3.58}$ & $0.54_{\pm 0.09}$ 

& $8.04_{\pm 2.71}$ & $12.05_{\pm 4.30}$ & $0.11_{\pm 0.10}$ 

& $8.41_{\pm 2.82}$ & $12.64_{\pm 4.29}$ & $0.02_{\pm 0.02}$ 
\\

Selfie {\small(BIN)} 
& $8.38_{\pm 2.84}$ & $12.69_{\pm 4.44}$ & $0.02_{\pm 0.03}$ 

& $8.39_{\pm 2.86}$ & $12.67_{\pm 4.40}$ & $0.02_{\pm 0.03}$  

& $8.33_{\pm 2.84}$ & $12.59_{\pm 4.40}$ & $0.03_{\pm 0.05}$ 
\\

CoTeaching {\small(BIN)} $^\dagger$ 
& $8.20_{\pm 2.89}$ & $12.33_{\pm 4.35}$ & $0.07_{\pm 0.05}$  

& $8.34_{\pm 2.75}$ & $12.52_{\pm 4.23}$ & $0.04_{\pm 0.04}$ 

& $8.40_{\pm 2.83}$ & $12.67_{\pm 4.35}$ & $0.02_{\pm 0.03}$
\\

CNLCU-S {\small(BIN)} $^\dagger$ 
& $8.12_{\pm 2.99}$ & $12.22_{\pm 4.57}$ & $0.10_{\pm 0.12}$ 
& $8.45_{\pm 2.80}$ & $12.65_{\pm 4.33}$ & $0.02_{\pm 0.03}$ 
& $8.40_{\pm 2.83}$ & $12.64_{\pm 4.39}$ & $0.02_{\pm 0.03}$ 
\\

CNLCU-H {\small(BIN)} $^\dagger$ 
& $8.22_{\pm 2.72}$ & $12.34_{\pm 4.13}$ & $0.06_{\pm 0.06}$

& $8.35_{\pm 2.83}$ & $12.65_{\pm 4.40}$ & $0.02_{\pm 0.03}$ 

& $8.37_{\pm 2.82}$ & $12.55_{\pm 4.36}$ & $0.04_{\pm 0.04}$ 
\\

CoSelfie {\small(BIN)} $^\dagger$ 
& $8.35_{\pm 2.89}$ & $12.62_{\pm 4.45}$ & $0.03_{\pm 0.04}$ 

& $8.39_{\pm 2.90}$ & $12.65_{\pm 4.43}$ & $0.02_{\pm 0.04}$ 

& $8.28_{\pm 2.77}$ & $12.42_{\pm 4.26}$ & $0.05_{\pm 0.06}$ 
\\

\midrule
\textbf{GRAIN (ours)} 
& $\bm{4.80_{\pm 1.53}}$ & $\bm{7.37_{\pm 2.79}}$ & $\bm{0.64_{\pm 0.06}}$

& $\bm{6.37_{\pm 1.98}}$ & $\bm{9.20_{\pm 2.80}}$ & $\bm{0.45_{\pm 0.21}}$

& $\bm{7.25_{\pm 2.66}}$ & $\bm{10.68_{\pm 3.67}}$ & $\bm{0.27_{\pm 0.12}}$ 
\\
\textbf{GRAIN-{\small AUG} (ours)} 
& $\bm{5.13_{\pm 1.48}}$ & $\bm{7.70_{\pm 2.84}}$ & $\bm{0.63_{\pm 0.08}}$ 

& $\bm{6.26_{\pm 2.04}}$ & $\bm{9.15_{\pm 3.19}}$ & $\bm{0.45_{\pm 0.07}}$ 

& $\bm{7.30_{\pm 2.30}}$ & $\bm{10.94_{\pm 3.19}}$ & $\bm{0.25_{\pm 0.17}}$ 
\\
\bottomrule
\end{tabular}
}
\end{table*}

\section{Discussion: Limitations, Future Work, and Broader Impact}\label{appdx.limitation}

{
\subsection{Connections to Broader Research and Potential Applications}
\label{app:extended_related_work}

From a statistical perspective, GRAIN relates to measurement-error analysis and likelihood-based inference
\citep{yi2017statistical,grace2021handbook,reid2026likelihood}, with broader discussions of noisy-data learning and multivariate regression in \citep{hu2024statistical,cui2023multivariate}.
Related work addresses response misclassification \citep{guo2026addressing,fang2022regularized},
covariate errors and variable selection \citep{yi2015functional,yi2019parametric,cui2025multivariate,cui2024variable},
and longitudinal or survival observations with missingness, truncation, or censoring
\citep{yi2015variable,chen2021analysis,chen2021variable,chen2020model}.
Extensions to crowdsourced or causation labels \citep{guo2023label,guo2024learning,hu2023addressing}
and missing or interval-censored responses \citep{bian2025empirical,bian2024unified}
would require observation-specific models rather than treating all imperfect supervision as additive label noise.

Beyond a single training distribution, source-free adaptation connects uncertainty control, noisy pseudo-labels, and discretized regression \citep{yi2023source,xu2025unraveling,xu2025revisiting,xu2026discretized}. Combining these ideas with structure-preserving graph learning and graph adaptation \citep{fang2025homophily,fang2026saga}
could extend GRAIN to continuous targets on relational inputs under distribution shift.
Polymer-property transfer learning \citep{fang2026transfer} provides a concrete application for investigating such extensions when property measurements are uncertain.

Potential downstream uses include robust auxiliary outcome or score prediction within physician evaluation and treatment-effect analysis \citep{bian2025physician,yi2023estimation}, longitudinal causal modeling, treatment-rule learning, and behavioral analyses \citep{hu2026inference,bian2025joint}. These applications require task-specific models, with causal identification and confounding addressed separately where appropriate. Evaluation could additionally examine limitations under intentional input perturbations \citep{li2026versatile,li2026priors,cai2026fuse}.
}

\subsection{Limitations and Outlook}
\label{appdx.limitations_outlook}

\paragraph{Theoretical Limitations}
The statistical guarantees in Theorems \ref{thm.robust_main} and \ref{thm.robust_main_dro} are 
established under a linear feature parameterization over fixed 
representations, which serves as a tractable proxy for the regression 
head's behavior. In practice, the feature extractor is a deep network 
jointly optimized with all other components. One principled connection 
between these two settings is provided by the lazy training regime~\citep{chizat2019lazy, jacot2018neural}, 
under which overparameterized networks trained with small learning rates 
remain close to their initialization and behave approximately as linear 
models in the induced reproducing kernel Hilbert space. Under such a 
regime, the guarantees of Theorems \ref{thm.robust_main} and \ref{thm.robust_main_dro} offer a 
principled characterization of the regression head's statistical 
behavior given sufficiently expressive and stable features. We note that the empirical results consistently validate the framework across multiple architectures and noise conditions, providing practical support beyond the linear 
setting. Bridging this connection rigorously for the joint optimization setting remains 
an open problem. Additionally, the contrastive 
regularization in Section~\ref{sec.feature_structuring} currently lacks formal 
guarantees; a rigorous analysis of its 
order-preserving effect on feature geometry under general noise 
distributions and jointly optimized representations constitutes 
an important direction for future theoretical development.

\paragraph{Future Work.}
We identify several directions for future investigation. 
First, 
extending the theoretical guarantees to weakly dependent 
settings, such as time series or spatially correlated 
observations, and to more general label noise generation models 
would broaden the statistical applicability of GRAIN beyond the 
current i.i.d.\ assumption; one natural approach is to replace the 
concentration inequalities used in the current analysis with 
martingale-based counterparts, such as Azuma--Hoeffding or 
Bernstein-type inequalities for dependent sequences. 
Second, 
rigorously bridging the gap between the linear theoretical setting 
and the deep network setting used in practice, for instance by 
leveraging lazy training arguments beyond the regression head to 
cover the joint optimization of all components, constitutes an 
important open problem. 
Third, developing a formal analysis of the 
contrastive regularization and its order-preserving effect on feature 
geometry under general noise distributions and jointly optimized 
representations remains a key theoretical direction; drawing on 
emerging theory for contrastive learning~\citep{wang2020understanding, 
saunshi2019theoretical} to characterize how the order-preserving 
objective shapes the geometry of learned representations under noisy 
supervision provides a promising starting point.

\paragraph{Broader Impact}
This work addresses a fundamental challenge in supervised learning: 
reliable regression under imperfect supervision. The methods developed 
here are general-purpose and applicable across domains where continuous 
labels are subject to measurement error or annotation uncertainty, 
including medical imaging, remote sensing, and scientific instrumentation. 
In such high-stakes settings, improved robustness to label noise may 
contribute to more reliable predictive systems and reduce the risk of 
performance degradation due to data quality issues.

%%%%%%%%%%%%%%%%%%%%%%%%%%%%%%%%%%%%%%%%%%%%%%%%%%%%%%%%%%%%

% \newpage
% \input{checklist.tex}

\end{document}